\documentclass[letterpaper, 10 pt, conference]{ieeeconf}
\usepackage[caption=false,font=footnotesize]{subfig}
\usepackage{amssymb}
\IEEEoverridecommandlockouts 
\usepackage{amsmath,amsfonts}
\usepackage{graphicx}
\usepackage{float}
\usepackage{physics} % for \atan2 and other math functions
\usepackage{cancel} % for \cancel command
\usepackage{cite}
\usepackage{tikz} % for tikzpicture environment
\usepackage[linesnumbered,ruled,vlined]{algorithm2e} % for algorithms and \tcp command
\newtheorem{lemma}{Lemma}

\usepackage{algorithm}

\title{Optimal Control and Structurally-Informed Gradient Optimization of a Custom 4-DOF Rigid-Body Manipulator}
\author{Brock Marcinczyk,  
        Logan E. Beaver
\thanks{Department of Mechanical and Aerospace Engineering, 
        Old Dominion University, Norfolk, VA, USA. 
        E-mails:, {\tt\small ... @odu.edu}}}

\usetikzlibrary{patterns}
\begin{document}
\setlength{\textfloatsep}{4pt}
\setlength{\floatsep}{4pt}
\setlength{\intextsep}{4pt}

\AtBeginDocument{
  \setlength{\abovedisplayskip}{2pt}
  \setlength{\belowdisplayskip}{2pt}
  \setlength{\abovedisplayshortskip}{0pt}
  \setlength{\belowdisplayshortskip}{0pt}
}
\maketitle
\thispagestyle{empty}
\pagestyle{empty}

\begin{abstract}
This work develops a control-centric framework for a custom 4-DOF rigid-body manipulator by coupling a reduced-order Pontryagin’s Maximum Principle (PMP) controller with a physics-informed Gradient Descent stage. The reduced PMP model provides a closed-form optimal control law for the joint accelerations, while the Gradient Descent module determines the corresponding time horizons by minimizing a cost functional built directly from the full Rigid-Body Dynamics. Structural-mechanics reaction analysis is used only to initialize feasible joint velocities—most critically the azimuthal component—ensuring that the optimizer begins in a physically admissible region. The resulting kinematic trajectories and dynamically consistent time horizons are then supplied to the symbolic Euler–Lagrange model to yield closed-form inverse-dynamics inputs. This pipeline preserves a strict control-theoretic structure while embedding the physical constraints and loading behavior of the manipulator in a computationally efficient way.
\end{abstract}
\section{Introduction}
In this project, we develop an unconstrained optimal controller based on the closed-form double-integrator solution obtained from Pontryagin’s Maximum Principle. The terminal positions and boundary conditions are not prescribed a priori; instead, they are determined naturally through a Gradient–Descent time-horizon estimator that incorporates both structural mechanics and the results of the Euler–Lagrange (EL) dynamics. This coupling allows the solver to initialize the joint velocities in a physically meaningful way—most notably the azimuthal velocity $\phi$, which is estimated using a Shigley-style reaction analysis~\cite{marcinczyk2026appendix,shigley2015}. The result is a controller that is both robust and structurally feasible, while remaining computationally efficient due to the symbolic, closed-form nature of Pontryagin’s Maximum Principle.

Although structural mechanics is not typically emphasized in optimal 
control formulations, its inclusion here is unavoidable. The 
Shigley–style reaction analysis directly determines the manipulator’s 
allowable torque envelope, which in turn defines the feasible region 
for both the Gradient–Descent horizon estimator and the reduced 
optimal control law. In practice, the azimuthal reaction torque places the 
tightest limit on admissible trajectories; exceeding this limit leads 
to motor stall, loss of controllability, and divergence in the Gradient Descent 
updates.

\subsection{Related Works}
\quad Reduced–order optimal control has been explored in several control–theoretic contexts. Cohen \emph{et al.}~\cite{Cohen2020Approximate}, for example, introduced a reduced Hamilton–Jacobi–Bellman (HJB) framework that lowers computational demands by simplifying the structure of the value function. Although their formulation does not target robotic manipulators specifically, it provides an important precedent: by reducing the order of the optimal control model, one can obtain solutions that remain tractable without sacrificing performance.

\indent The present work adopts this same philosophy but implements it through Pontryagin’s Maximum Principle (PMP) rather than the HJB framework. Whereas Cohen employs a nonlinear functional for the control input, the cost functional here is quadratic in $\ddot{q}_{i}$, enabling the reduced PMP equations to yield a fully symbolic, closed-form solution. This produces an analytic double-integrator control law that can be computed entirely offline, bypassing the iterative machinery inherent in dynamic programming and offering a computationally lightweight framework for robotic manipulation.
 \newline
\indent A recurring difficulty in the literature is the use of gradient-based controllers for nonlinear systems. Steep, ill-conditioned gradients—particularly those arising from coupled rotational kinematics—can severely compromise stability. As demonstrated by Canzoneri and Giarré~\cite{Canzoneri2020GDGP}, such nonlinearities routinely \textit{plague} gradient-descent-based methods, yielding incorrect update directions, divergence, or undesirable coupling between degrees of freedom. To avoid these pitfalls, Gradient Descent in this work is restricted solely to the time-horizon estimation step, not the control law itself.
\newline
\indent Many optimal control methods for robotic manipulators rely on numerically intensive solvers or repeated optimization cycles. By contrast, the present framework remains lightweight: the Euler–Lagrange equations supply the full dynamics, the inverse dynamics remain symbolic, and the only numerical components are a forward-Euler integrator and the Gradient-Descent time estimator. Unconstrained PMP dictates the optimal inputs, while Gradient Descent determines \textit{when} the optimal trajectory terminates. This separation allows high-resolution trajectory generation without exposing the controller to the nonlinear instabilities that \textit{plague} gradient-based schemes. Using reduced models are computationally efficient, and the use of Euler-Lagrange and Optimal Control are computationally feasible tools.\cite{beaver2023flatness}\\
\indent Operational-space formulations have also influenced modern manipulator control. Khatib’s pseudo-inertia matrix~\cite{khatib1995inertial}; where 
$\Lambda = (J_{v} M^{-1} J_{v}^{\top})^{-1}$, 
characterizes how easily the end-effector can accelerate in each Cartesian direction. Ill-conditioned entries of $\Lambda$ correspond to kinematic or dynamic singularities. Recent work~\cite{Almarkhi2019Singularity} formalizes this connection and shows that singular behavior can be detected directly from the structure of $\Lambda$. This motivates the directional weighting strategy used here: by penalizing directions associated with low operational inertia, the Gradient-Descent horizon estimator avoids drifting toward near-singular configurations and maintains numerical stability.\\
\textbf{Paper Overview: }\\
The remainder of this manuscript proceeds in four stages.
First, the full Euler–Lagrange formulation of the 4-DOF manipulator is developed alongside the structural-mechanics framework used to evaluate joint reactions, bending loads, azimuthal torque limits, and the corresponding payload capacity. Next, the Gradient–Descent horizon estimator is introduced, together with its coupling to the Euler–Lagrange dynamics and the operational-space weighting strategy that prevents the system from drifting toward near-singular configurations. This is followed by the reduced-order optimal control derivation based on Pontryagin’s Maximum Principle, which yields a closed-form double-integrator expression for the optimal joint accelerations. Finally, numerical results are presented, including convergence behavior, optimized trajectories, actuator responses, and the structural feasibility of the resulting motions.
A concluding discussion and all supplementary derivations are provided in the appendix. We now proceed to the development of the control law, together with the supporting derivations associated with the system dynamics and control formulation.
\section{Rigid Body Dynamics and Structural Mechanics}
\begin{figure}[H] \centering \includegraphics[width=0.75\linewidth]{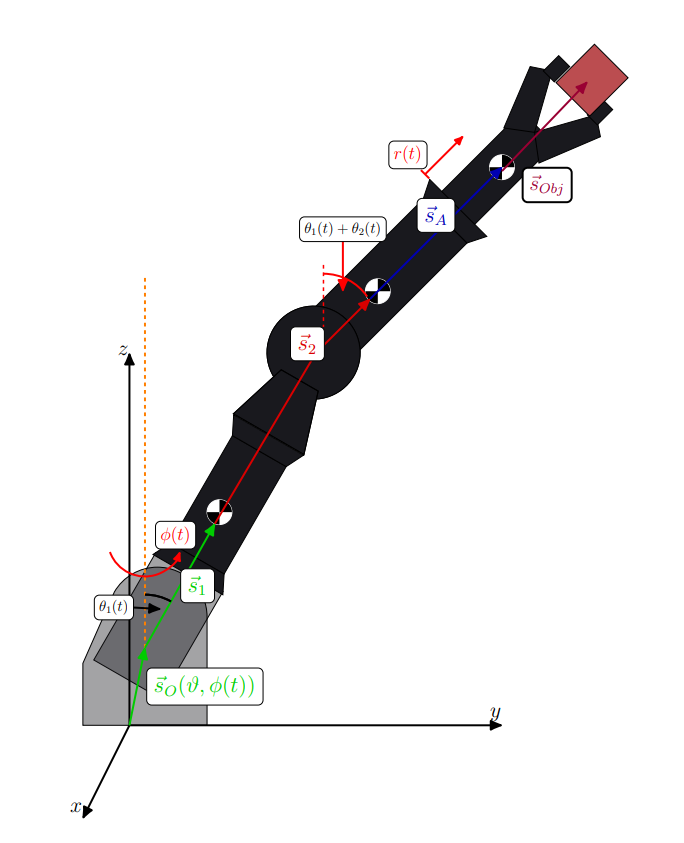}2 \caption{Free-body diagram of the 4-DOF robotic apparatus.} \label{fig:trashbot_fbd} \end{figure} To model the motion of the manipulator, all kinematics and inertial quantities are expressed in the spherical basis shown in Fig.~\ref{fig:trashbot_fbd}. This coordinate choice keeps the representation compact and allows each link’s inertia to be treated directly in its own rotating frame. At this stage of the project, the link inertias are taken directly from the CAD model. A more accurate conversion—using experimentally measured masses and a radius-of-gyration formulation—will be incorporated in future work once the physical prototype is available. \\ To model the Robotic Model, the position vectors are defined in terms of the unit vectors in the radial basis: Where: \begin{align*} e_{r_{0}} = \langle \cos\left(\phi\right)\sin\left(\vartheta\right), \sin\left(\phi\right)\sin\left(\vartheta\right), \cos\left(\vartheta\right)\rangle^{\top}\\ e_{r_{1}} = \langle \cos\left(\phi\right)\sin\left(\theta_{1}\right), \sin\left(\phi\right)\sin\left(\theta_{1}\right), \cos\left(\theta_{1}\right)\rangle^{\top} \\ e_{r_{2}} = \langle \cos\left(\phi\right)\sin\left(\theta_{12}\right), \sin\left(\phi\right)\sin\left(\theta_{12}\right), \cos\left(\theta_{12}\right)\rangle^{\top}\\ \end{align*} \
The corresponding position vectors are given through:  
\begin{align*}
\vec{s}_{0}      &= \ell_{0}\, e_{r_{0}} \\
\vec{s}_{1}      &= \vec{s}_{0} + \bar{\ell}_{1}\, e_{r1} \\
\vec{s}_{2}      &= \vec{s}_{0} + \ell_{1}\, e_{r1} + \bar{\ell}_{2}\, e_{r2} \\
\vec{s}_{A}      &= \vec{s}_{0} + \ell_{1}\, e_{r1} + \left(\ell_{2} + r(t)\right)e_{r2} \\
\vec{s}_{Obj}    &= \vec{s}_{0} + \ell_{1}\, e_{r1} + \left(\ell_{2} + r(t) + r' + \delta_{r}\right)e_{r2}
\end{align*}
 Note, $\vartheta$ is the vertical angle between the frame of reference, and the polar mounting of the Robotic Apparatus; and does not change so it is left as a constant. The centers of mass, velocities, and corresponding kinetic and potential energies are assembled directly from these spherical basis vectors. Basic Rayleigh damping is added for each coordinate, and the full Euler–Lagrange equations follow from
\begin{align}
\frac{d}{dt}\!\left(\frac{\partial \mathcal{L}}{\partial \dot{q}_i}\right)
-
\frac{\partial \mathcal{L}}{\partial q_i}
+
\frac{\partial \mathcal{R}}{\partial \dot{q}_i}
&= U_i,
\label{EL}
\end{align}
\noindent
All symbolic derivations,
intermediate expressions, and coordinate transformations are summarized in the
appendix.
\noindent\textbf{Virtual Spring--Damper Note.}
Each degree of freedom incorporates a virtual spring to model actuator
elasticity, using the manufacturer's rated force or torque divided by the
corresponding stroke. The associated damping coefficient is chosen from the
critically damped approximation. The stiffness and damping values are thus
\begin{align*}
    \kappa_{q_i} &= \frac{u_{\mathrm{rated}}}{q_{\mathrm{stroke}}},\qquad
    b_{q_i} = 2\sqrt{\frac{M_{q_i}\,u_{\mathrm{rated}}}{q_{\mathrm{stroke}}}}.
\end{align*}
This formulation is deliberate. By defining both the virtual stiffness and damping directly from the rated actuator limits, the springs enforce motion that never demands more torque than the drivetrain can safely supply—even if the joint is operating through a large reduction ratio. More importantly, the convention remains valid under hardware changes: should the gearing be modified, the rated torque does not change, so the virtual spring–damper model preserves its physical consistency without requiring retuning. This keeps the GDTH layer grounded in actuator feasibility while still allowing the optimal controller to explore aggressive trajectories.
Note however, that for Direct-Current motors, we can exploit this principle and in future models; PID tuning will be appplied in future models for its inverse dynamics. 
These virtual elements provide a simple but physically meaningful approximation
of actuator compliance within the Euler--Lagrange framework. In future work,
the virtual Rayleigh model will be substituted with a full Stribeck friction model
using experimentally measured damping and friction parameters; this refinement
is not included here, as it lies beyond the scope of the present study.\\
\textbf{Structural Mechanics Note.} While the Euler–Lagrange model above governs the manipulator’s motion, the structural mechanics are analyzed in the same frame of reference. 
This separation is intentional: the EL model describes how the joints accelerate and interact dynamically, whereas the structural model determines how those joint torques manifest as bending moments, shear forces, and gear loads within the hardware. Reaction forces at the shoulder and elbow, as well as the transmitted torques through the shaft, are evaluated using standard methods from Shigley’s \emph{Mechanical Engineering Design}. 
In future works, these reactions are later used to define stress-based constraints and gear-loading limits for the controller. A full breakdown of the reaction analysis, including free-body expansions and gearshaft equilibrium equations, is provided in the appendix.
These results from the reaction analysis provide a direct estimate of the maximum payload that the system can safely recover. Based on the torque surfaces and the CAD-derived inertial parameters used in this study, the critical payload capacity is approximately 15 lb (6.80389 kg); previous CAD-iterations have had weight capacities set as 10 lbs; and the increased gear ratio as a result, transmits more torque and thus a higher torque limit can be obtained. Beyond this mass, the required azimuthal reaction torque exceeds the motor’s stall limit.
\begin{figure}[H]
    \centering
    \includegraphics[width=0.75\linewidth]{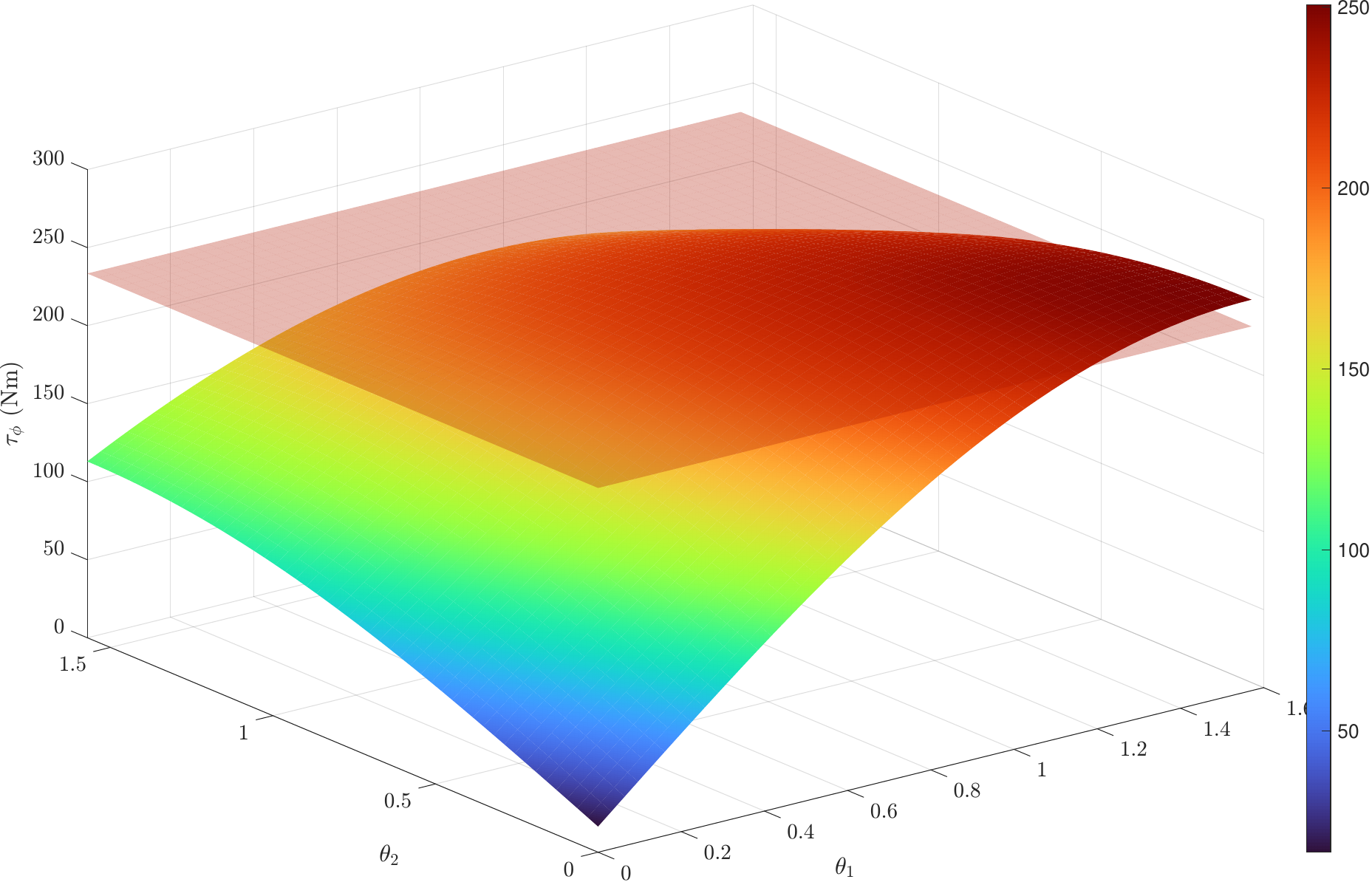}
    \caption{
        \textbf{Reaction torque surfaces for varying $\theta_1,\theta_2$ values.}
        Case shown: $m_{\text{obj}} = 7.11~\mathrm{kg}$.
        The light-red plane denotes the motor-equivalent stall torque
        after gearing. Rigid-body reactions are shown beneath this limit.
    }
    \label{fig:torque_surface_7kg}
\end{figure}
\begin{figure}[H]
    \centering
    \includegraphics[width=0.75\linewidth]{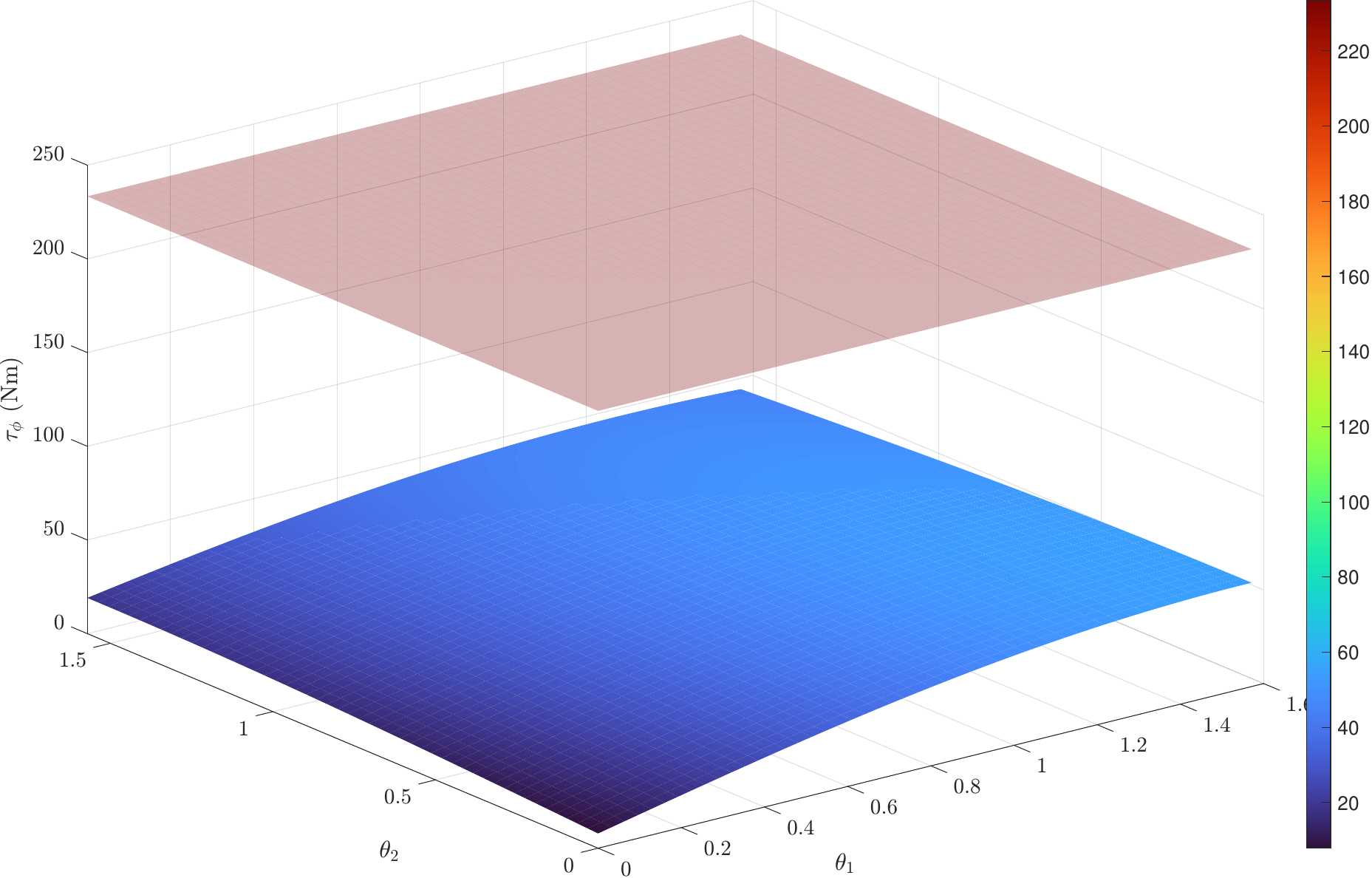}%
    % from file: :contentReference[oaicite:0]{index=0}
    \caption{
        \textbf{Reaction torque surfaces for varying $\theta_1,\theta_2$ values.}
        Case shown: $m_{\text{obj}} = 0~\mathrm{kg}$ (unloaded arm).
        The stall-torque plane is identical to the previous figure, but the
        reaction torques are substantially reduced across the full workspace,
        demonstrating the mass-dependent scaling of azimuthal loading.
    }
    \label{fig:torque_surface_0kg}
\end{figure}

The semi-transparent z-plane in each plot denotes the motor’s stall torque limit, and the surface shows the required azimuthal reaction torque as a function of the joint angles 
$\theta_{1}$
 and 
$\theta_{2}$. Intersections where the surface exceeds this plane correspond to configurations where the system would be unable to lift or rotate the payload without exceeding actuator limits.
\section{Optimal Control}
With the time horizons provided by Gradient Descent, we now construct the optimal
kinematics using a reduced-order representation of the manipulator. The reduced
model retains only the components of the state that admit a closed-form
solution under Pontryagin’s Maximum Principle (PMP), allowing the optimal
inputs to be expressed symbolically without integrating the full nonlinear
dynamics.

Although the nonlinear coupling terms of the Euler–Lagrange system are omitted,
the reduced model remains physically valid. Because the time horizons are
computed from the fully coupled rigid-body dynamics, the resulting trajectories
respect the same inertial structure, feasibility limits, and physical
constraints. The reduced PMP system therefore acts as an analytically tractable
surrogate whose solutions remain consistent with rigid-body mechanics.

\subsection{Optimality Conditions}

We briefly summarize the fundamental conditions for optimality under an
unconstrained PMP formulation.

\begin{lemma}[Unconstrained PMP Optimality]
Consider a system with reduced state $x$ and control input $u$, governed by  
$\dot{x} = f(x,u)$ with running cost $L(x,u)$.  
For an optimal control $u^\star(t)$ over horizon $[0,t_f]$, the following
conditions must hold:
\begin{enumerate}
    \item \textbf{Hamiltonian definition:}
    \[
        \mathcal{H}(x,\lambda,u)
        = L(x,u) + \lambda^\top f(x,u).
    \]

    \item \textbf{Costate dynamics:}
    \[
        \dot{\lambda}(t) = -\frac{\partial \mathcal{H}}{\partial x}.
    \]

    \item \textbf{Stationarity condition:}
    \[
        \frac{\partial \mathcal{H}}{\partial u} = 0,
    \]

    \item \textbf{State evolution:}
    \[
        \dot{x}(t) = \frac{\partial \mathcal{H}}{\partial \lambda}
        = f(x(t),u^\star(t)).
    \]
\end{enumerate}
\end{lemma}

These conditions characterize the optimal kinematics used in the reduced PMP
framework and enable a closed-form, double-integrator control law consistent
with the rigid-body dynamics certified by Gradient Descent. 

\begin{align*}
    J^{\star} = \min_{u}\int_{t_{0}}^{t_{f_{q_{i}}}} \frac{u^{\top} u}{2} dt \\
    \text{Subject to: } \\ 
    \quad u = \ddot{q}\\
    f(x,u) = \langle \dot{q}, \ddot{q}_{i}\rangle^{\top}
\end{align*}
The Hamiltonian, not the Rigid Body Dynamics Hamiltonian is given by: 
\begin{align*}
    \mathcal{H}^{\star} = \lambda^{\top}f(x,u) +  \frac{u^{\top} u}{2}\\
     \quad = \lambda_{q} \dot{q}_{i} + \lambda_{\dot{q}_{i}} u + \frac{u^{2}}{2}
\end{align*}
Applying the Optimality Conditions yields the optimal system accelerations as follows: 
\begin{align*}
\frac{\partial \mathcal{\mathcal{H}^{\star} }}{\partial u} = u + \dot{\lambda}_{q} u = 0 \Rightarrow u^{\star} = -\dot{\lambda}_{q}.
\end{align*}
Substituting this into the Hamiltonian will then yield the following form: 
\begin{align*}
    \mathcal{H}^{\star} \Bigg|_{u = u^{\star}} = \lambda_{q} \dot{q} - \frac{\dot{\lambda}_{q}^{2}}{2}
\end{align*}
From the optimality conditions, it is necessary to define the Hamiltonian in terms of the costate derivative. 
Taking the gradient of the substituted Hamiltonian in terms of the state vector: 
\begin{align*}
    \frac{\partial \mathcal{H}^{\star}}{\partial \lambda} 
    =  \begin{bmatrix} \dot{q} \\ \dot{\lambda}_{q}\end{bmatrix}\\ \frac{\partial \mathcal{H}^{\star}}{\partial x}  = \begin{bmatrix} 0 \\ -\lambda_q\end{bmatrix}
\end{align*}
Performing the Integrations will yield: 
\begin{align*}
    \lambda_{q} = c_{1} \\
    \dot{\lambda_{q}} = \ddot{q}^{\star} = -\left(c_{1}t + c_{2}\right)
\end{align*}
Noting that the acceleration is equal to the costate associated with the velocity, the optimized kinematics take the form
\begin{align*}
    \dot{q}^{\star} &= \frac{c_{1} t^{2}}{2} + c_{2} t + c_{3},\\ 
    q^{\star} &= \frac{c_{1} t^{3}}{6} + \frac{c_{2} t^{2}}{2} + c_{3} t + c_{4}.
\end{align*}
Since these expressions form a Hermite polynomial structure, which can become lengthy when written explicitly, the solution is expressed compactly in matrix form as
\begin{align*}
 \begin{bmatrix}
    \dfrac{t_{f_{q_i}}^{2}}{2} & t_{f_{q_i}} & 1 & 0\\
    \dfrac{t_{f_{q_i}}^{3}}{6} & \dfrac{t_{f_{q_i}}^{2}}{2} & t_{f_{q_i}} & 1 \\
     0 & 0 & 1 & 0  \\
     0 & 0 & 0 & 1  \\
 \end{bmatrix}
 \begin{bmatrix}
    c_{1} \\ c_{2} \\ c_{3} \\ c_{4}
 \end{bmatrix}
 =
 \begin{bmatrix}
    \dot{q}_{F}\\ q_{F}\\ \dot{q}_{0}\\ q_{0}
 \end{bmatrix}.
\end{align*}

\begin{figure}[H]
    \centering

    % --- q(t) ---
    \subfloat[State trajectory $q(t)$]{%
        \includegraphics[width=0.45\linewidth]{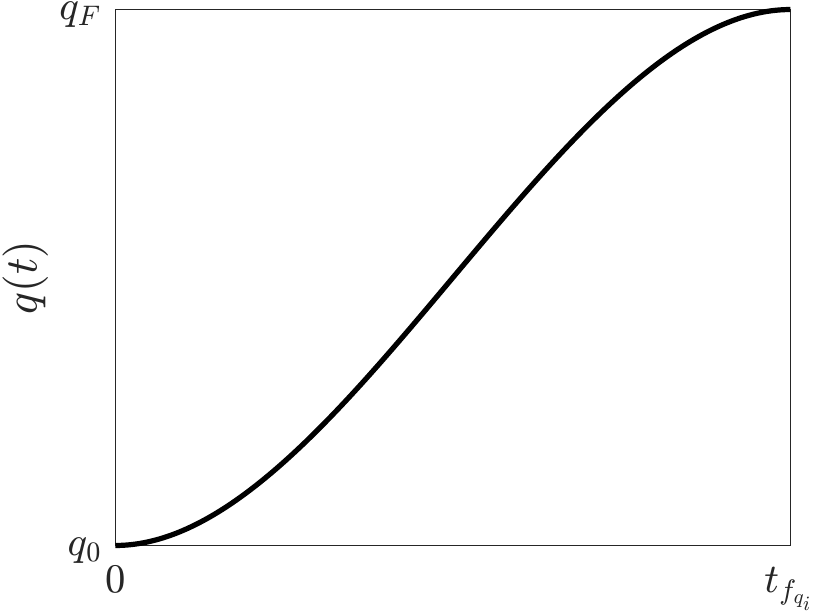}%
        \label{fig:q_t}
    }\hfill
    % 
    % --- qdot(t) ---
    \subfloat[Velocity profile $\dot{q}(t)$]{%
        \includegraphics[width=0.45\linewidth]{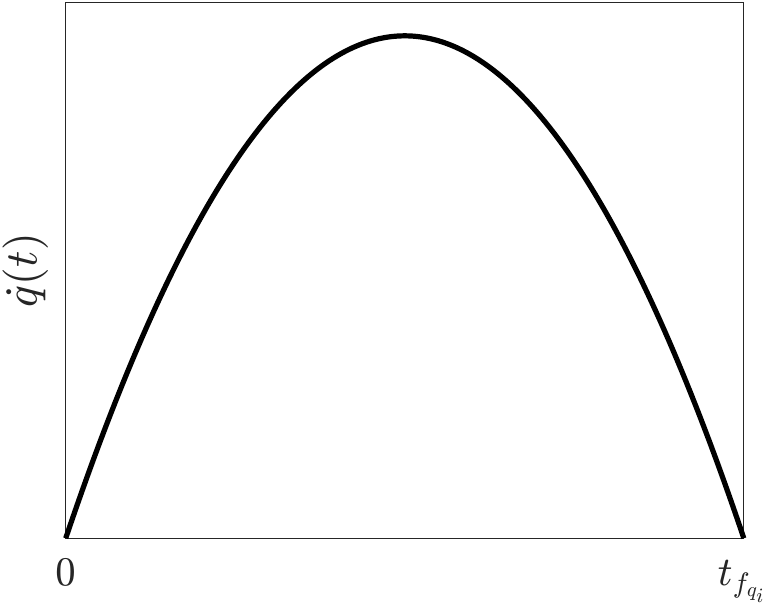}%
        \label{fig:qdot_t}
    }

    \caption{Pontryagin-optimal rest-to-rest PMP trajectory components.  
    (a) State evolution $q(t)$ and (b) velocity evolution $\dot{q}(t)$ for 
    boundary conditions $(q_0, \dot{q}_0) \rightarrow (q_F, \dot{q}_F)$ 
    with $\dot{q}_0 = \dot{q}_F = 0$.}
    \label{fig:PMP_rest_to_rest}
\end{figure}

Because the optimal control solution enforces the boundary conditions by construction, the resulting trajectories remain within the admissible manifold—perturbations that would violate these constraints are infeasible, as illustrated in Lemmon’s lecture notes (Fig.~1)~\cite{lemmon_ee565_module3_2015}. Consequently, the velocity profile follows a smooth quadratic form for rest-to-rest motion, and the corresponding position trajectory assumes the familiar half-bell shape, as shown in Fig.~\ref{fig:PMP_rest_to_rest}.

To make the Euler--Lagrange dynamics both controllable and computationally efficient, the solutions from \ref{EL} are written in the standard manipulator form with applied dissipation:
\begin{align}
M(q^{\star})\,\ddot{q}^{\star}
+ C(q^{\star},\dot{q}^{\star})\,\dot{q}^{\star}
+ \nabla V(q)^{\star}
+ B\,\dot{q}^{\star}
= U.
\end{align}
Thus, the optimized dynamics arise directly from the Euler--Lagrange formulation with dissipation and applied generalized inputs, and the substitution of the optimal control solution yields a torque-level controller that is both robust and inherently stable.

To determine the optimal time horizon associated with these Euler--Lagrange trajectories, a Gradient--Descent procedure is applied to the kinetic-energy profile. This reduces unsteady, velocity-dominated motion and produces a time allocation that is dynamically consistent with the optimal kinematics.
 Asides from the Dynamics in this case; the optimized kinematics produced from the algorithm produces smooth motion in the joint coordinates, and similarly with the cases in velocity as well. 

\subsection{Gradient Descent--Based Horizon Estimation}
The Gradient--Descent Horizon Estimator determines how long each joint
should evolve under the reduced PMP control law before reaching its desired
configuration. Rather than prescribing terminal times, the estimator computes
them automatically by minimizing a cost functional derived from the Rigid--Body
Dynamics Hamiltonian:
\begin{align}
\mathcal{H}
    = \tfrac{1}{2}\dot{q}_{i}^{\top}M\dot{q}_{i}
    + g\sum_{i=1}^{4} m_{i}z_{i}
    + \sum_{i=1}^{4}\tfrac{\kappa_{i}}{2}\big(q_{i}-q_{o,i}\big)^{2},
\label{Hamiltonian}
\end{align}
where $M$ is the mass matrix, the second term represents gravitational
potential energy, and the final term models elastic contributions introduced by
virtual springs.
To regulate the motion toward a desired spatial target, Gradient Descent dynamically weights
the kinematic error using the inverse of the Pseudo--Operational Inertia
Matrix~\cite{khatib1995inertial}. Because the inertia distribution of the robot
varies strongly with posture, the vertical error of the first link is normalized
with respect to its center of mass:
\begin{align}
e_{z,1}(t)
=  z_{1}- \frac{l_0\cos\vartheta + \bar{l}_{1}}{l_0\cos\vartheta + l_{1}}z_{\mathrm{Des}} .
\label{e_z1}
\end{align}
This scaling ensures consistency between the error metric and the inertial
representation embedded in $M$.
The end--effector error is obtained directly from forward kinematics. Because
the full end--effector kinematics already appear in the mass matrix, no
additional normalization is required:
\begin{align}
e_{EE}(t) &=
\begin{bmatrix}
\vec{s}_{0}\hat{x} + \ell_{1}c_{\phi}s_{\theta_{1}}
    + \ell_{\mathrm{eff}}s_{\theta_{12}}c_{\phi}
    - x_{\mathrm{Des,EE}}
\\[0.7em]
\vec{s}_{0}\hat{y} +\ell_{1}s_{\phi}s_{\theta_{1}}
    + \ell_{\mathrm{eff}}s_{\theta_{12}}s_{\phi}
    - y_{\mathrm{Des,EE}}
\\[0.7em]
\vec{s}_{0}\hat{z} +\ell_{1}c_{\theta_{1}}
    + \ell_{\mathrm{eff}}c_{\theta_{12}}
    - z_{\mathrm{Des,EE}}
\end{bmatrix},
\\
c_{\alpha}&=\cos\alpha,\; s_{\alpha}=\sin\alpha. 
\end{align}
To prevent the algorithm from drifting into low--inertia directions or approaching
kinematic singularities, each Cartesian component is weighted by the inverse of
the corresponding entry in the operational-space inertia matrix:
\begin{align*}
\beta_{x}(t)=\frac{1}{\|\Lambda_{11}\|+\varepsilon}, \quad
\beta_{y}(t)=\frac{1}{\|\Lambda_{22}\|+\varepsilon}, \quad
\beta_{z}(t)=\frac{1}{\|\Lambda_{33}\|+\varepsilon}.
\end{align*}
Here, $\varepsilon$ is a slight perturbation; which prevents division by zero but massively drives the cost up and thus the Cost. 
Combining these elements yields the Gradient Descent cost functional
\begin{align}
J = \min_{\dot{q}_{i}}
\left(
    \mathcal{H}
    + \tfrac{1}{2} e_{EE}^{\top} R_{EE} e_{EE}
    + \tfrac{1}{2} e_{1}^{\top}  R_{1} e_{1}
\right),
\label{eq:J_cost} 
\end{align}
which is minimized iteratively. Here, $R_{EE}$, and $R_{1}$ are the diagonal weight matrices from the operational space inertia matrix. After computing the gradient $\nabla_{\mathcal{X}}J$,
a momentum update is applied to produce smoother descent directions:
\begin{align}
    n_{\mathcal{X}_{j},i+1}
        &= \beta_{q_i} n_{q_i,i}
           + (1-\beta_{q_i})\,\nabla_{\mathcal{X}}J, 
\label{MU}
\\
    \mathcal{X}_{i+1}
        &= \mathcal{X}_i
        - \alpha^{k}_{\mathcal{X}_{j},i}\,n_{\mathcal{X}_{j},i+1},
\label{X_new}
\end{align}
where $\mathcal{X}$ denotes a generic joint coordinate. Because the radial
actuator moves more slowly than the rotational joints, each coordinate uses its
own learning rate $\alpha$, which gradually decays over time:
\begin{align}
   \alpha^{k}_{X_{j},i}
   = \frac{\alpha_{0}}{1+\eta t_{i}}.
\end{align}
Before Gradient Descent begins its updates, the initial torque estimate is computed by
combining the gravity-gradient contribution from the Euler--Lagrange dynamics
with the azimuthal reaction torque predicted by the structural model:
\begin{align*}
  \tau_{0}
    &=
    \frac{\partial}{\partial q_i}
    \!\left( g\sum_{i=1}^{4} m_i z_i \right)
    \begin{bmatrix}
        \hat{r} \\
        \hat{\theta}_{1} \\
        \hat{\theta}_{2} \\
        0
    \end{bmatrix}
    + u_{\phi_{0}}\,\hat{\phi}.
\end{align*}
The corresponding initial velocity used for Gradient Descent is then scaled according to the
available actuator torque:
\begin{align*}
v_{0}
    = v_{\mathrm{N.L}}
      \,\mathrm{sgn}(q_{i,f}-q_{i,0})\!
      \left(1-\frac{\|\tau_{0}\|}{\tau_{\mathrm{Stall}}}\right).
\end{align*}
Note that for $\dot{r}$, the term is weight according to the mass-ratio $\mu$, which is given by:
\begin{align*}
    \mu = \frac{m_{Claw}}{m_{Claw} + m_{Obj}}.
\end{align*}
Overall,the Gradient Descent controller serves as the bridge between the reduced-order Optimal Controller and
the full Euler--Lagrange dynamics. Instead of performing trajectory optimization
in joint space, Gradient Descent determines \emph{when} each joint should stop evolving,
yielding a physically meaningful time horizon that respects both inertia and
structural loading constraints.  The summary of the algorithm is presented in algorithm 1. 
\begin{algorithm}
\centering
\scalebox{0.875}{
\begin{minipage}{0.97\linewidth}

\caption{Gradient--Descent Time--Horizon Estimator (GDTH)}
\label{alg:GDTH} 
\DontPrintSemicolon

\textbf{Inputs:}
EE target $(x^d,y^d,z^d)$; initial state $q(0),\dot q(0)$;  
parameters $(m_{\rm Obj},\left(t_{i+1}-t_{i}\right))$.

\textbf{Outputs:}
$q^\star$ and $(t_r,t_{\theta_1},t_{\theta_2},t_\phi)$.

\BlankLine

Initialize freeze flags; set $i=1$.

\While{$\|e_q(i)\|>\varepsilon$ \textbf{and} $i<\mathrm{Iter}$}{

\tcp{1. Spatial weighting}
$e_q(i)\leftarrow |e_q(i)|$;  
$(\beta_x,\beta_y,\beta_z,\beta_{x_1},\beta_{y_1},\beta_{z_1})=\texttt{weightCalculation}(q(i),m_{\rm Obj})$.

\tcp{2. Euler prediction}
$X^{-}(i\!+\!1)=q(i)+\dot q(i)\left(t_{i+1}-t_{i}\right)$.

\tcp{3. Cost gradient}
$\nabla J(i)=\texttt{gradientDescent}(\beta,q(i),\dot q(i),x^d,y^d,z_1^d,z^d)$.

\tcp{4. Momentum descent}
$m(i\!+\!1)=\beta_m m(i)+(1{-}\beta_m)\nabla J(i)$;  
$X^{+}(i\!+\!1)=X^{-}(i\!+\!1)-\alpha(i)\,m(i\!+\!1)$.

\tcp{5. Velocity projection}
$\dot q(i\!+\!1)=\texttt{ClampControls}(\dot q(i\!+\!1))$.

\tcp{6. FK and cost}
$(x_{EE},y_{EE},z_{EE},e_q(i\!+\!1),e_{z_1}(i\!+\!1),J(i\!+\!1))=\texttt{costFunction}(\cdots)$.

\tcp{7. Joint--space error}
$e_{\rm gen}(i\!+\!1)=q^d-q(i\!+\!1)$.

\tcp{8. Freeze checks}
If $|e_{{\rm gen},j}(i\!+\!1)|\le\varepsilon_j$:  
$q_j(i\!+\!1)\!\gets\! q_j(i)$, $\dot q_j(i\!+\!1)\!=\!0$, record $t_j$.

\tcp{9. Overshoot rollback}
If $|e_{{\rm gen},j}(i\!+\!1)|>|e_{{\rm gen},j}(i)|$:  
$q_j(i\!+\!1)\!\gets\!q_j(i)$, $\dot q_j(i\!+\!1)\!\gets\!\dot q_j(i)$,  
$\alpha_j(i\!+\!1)=\alpha_j(i)/(1+\eta t(i))$.

\tcp{10. Exit if fully frozen}
If all DOFs frozen: break.

$i\leftarrow i+1$;
}

\KwRet{$q^\star=q(i)$ and $(t_r,t_{\theta_1},t_{\theta_2},t_\phi)$}.

\end{minipage}
}
\end{algorithm}
\section{Results}
The Gradient Descent based algorithm has a high convergence rate; with some noticeably low tolerances $\epsilon_{r} = 0.25''$, and for joint angles, the tolerance is $\epsilon_{\Theta} = 2^{\circ}$. 
\begin{figure}
    \centering
    \includegraphics[width=0.75\linewidth]{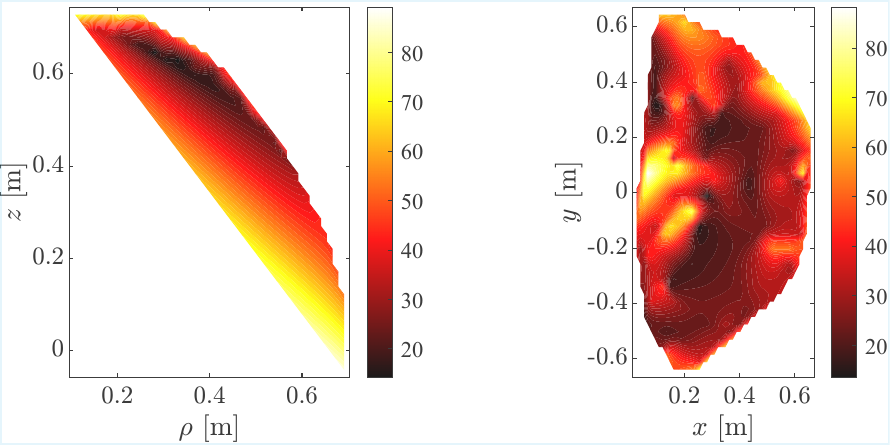}
    \caption{A total of 100 randomized target points were sampled across the robot’s reachable workspace. End–effector error is reported in millimeters. Across nearly all trials, the final Gradient–Descent solution converged to within roughly an inch of the desired target, and this behavior remained consistent across multiple CAD model revisions of the manipulator.}
    \label{fig:GDTH_TortureTests}
\end{figure}
\subsection{Simulation of Optimally Controlled Robot Arm}
Let us discuss a multi-stage case; where the robot is to pick up a  4 kg mass, and drop it into a bin. Consider the diagram as shown below. 
\begin{figure}
    \centering
    \includegraphics[width=1\linewidth]{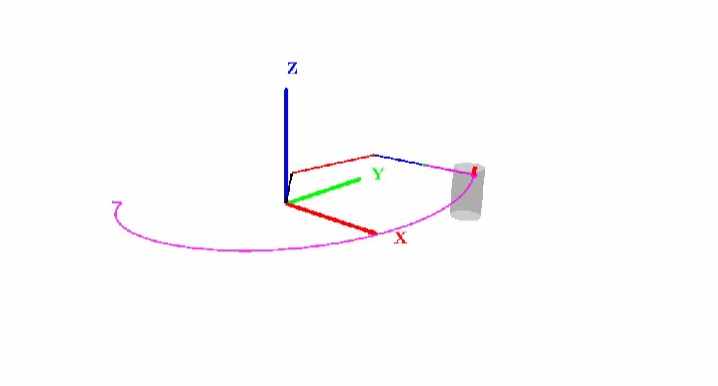}
    \caption{Optimal Control Simulation}
    \label{fig:placeholder}
\end{figure}
The system parameters is provided through this table 1. Note due to the manufacturing complications with the hardware, i.e. its low speed, the learning rate had to be asymmetrically modified from the rest of my other coordinates, which has an outstandingly lower learning rate relative to the other models. 
\begin{table}
\centering
\footnotesize
\renewcommand{\arraystretch}{0.5}
\setlength{\tabcolsep}{2pt}
\caption{Simulation Constants Used in PMP--GDTH Trajectory Optimization}
\label{tab:sim_constants}
\begin{tabular}{p{0.47\linewidth} p{0.40\linewidth}}
\hline
\textbf{Parameter} & \textbf{Value} \\
\hline
Rated radial force $u_{r,\mathrm{rated}}$ & $200~\mathrm{N}$ \\
Rated elbow torque $u_{\theta_{1},\mathrm{rated}}$ & $53~\mathrm{Nm}$ \\
Rated shoulder torque $u_{\theta_{2},\mathrm{rated}}$ & $53~\mathrm{Nm}$ \\
Rated azimuth torque $u_{\phi,\mathrm{rated}}$ & $53~\mathrm{Nm}$\\

Steel gear efficiency $\eta_{\theta_1}$ & $0.95$ \\
CF gear efficiency $\eta_{\theta_2},\eta_{\phi}$ & $0.85$ \\
Motor stall torque $\tau_{\mathrm{stall}}$ & $68.64655~\mathrm{Nm}$ \\
Eff.\ stall (shoulder) & $130.4~\mathrm{Nm}$ \\
Eff.\ stall (elbow) & $116.7~\mathrm{Nm}$ \\
Eff.\ stall (azimuth) & $233.4~\mathrm{Nm}$ \\
No-load linear velocity $V_{\mathrm{NL}}$ & $0.004\mu$ \\
No-load angular speed $\Omega_{\mathrm{NL}}$ & $6.3~\mathrm{rad/s}$ \\
\hline
\multicolumn{2}{c}{\textbf{Virtual Springs}} \\
\hline
Radial stroke $\Delta r$      & $0.1016~\mathrm{m}$ \\
Shoulder stroke $\Delta\theta_1$ & $\pi/2$ \\
Elbow stroke $\Delta\theta_2$    & $\pi/6$ \\
Azimuth stroke $\Delta\phi$      & $\pi/2$ \\
\hline
\multicolumn{2}{c}{\textbf{GDTH Parameters}} \\
\hline
Alpha decay $\alpha(t)$ & $\alpha_0/(1+\eta t)$, $\eta=0.01$ \\
Initial $\alpha_0$ & $[3\!\times10^{-8},\,10^{-3}]$ \\ $(\beta_r,\beta_{\theta_1},\beta_{\theta_2},\beta_\phi)$ & $(.025,.025,.025,.025)$ \\
Damped inertia $\Lambda_1$ & $\Lambda=10^{-3}$ \\
Damped inertia $\Lambda_{EE}$ & $\Lambda=10^{-3}$ \\
\hline
\multicolumn{2}{c}{\textbf{Waypoint Geometry}} \\
\hline
Waypoint 1 $(x_{EE},y_{EE},z_1,e_{EE})$ & $(  -0.4500 ,   0.4500,    0.1947,   0.0100)$ \\
Waypoint 2 $(x_{EE},y_{EE},z_1,e_{EE})$ & $( 0.5000    0.5000    0.2280    0.2000)$ \\
Waypoint 3 $(x_{EE},y_{EE},z_1,e_{EE})$ & $(0.3756    0.3756    0.4164    0.5080)$ \\
\hline
\multicolumn{2}{c}{\textbf{Waypoint Mass}} \\
\hline
$m_{Obj}$ & $(0,4\, kg,0)$ \\
\hline
\end{tabular}
\end{table}
It comes as no surprise that for case 2; the initial velocity guess is lowered. Significantly for the radial coordinate with the introduction of the mass ratio $\mu$, and obviously for the joint torques; the presence of $m_{Obj} = 2kg$ is going to obviously lower the required torque limits.
\begin{figure}[H]
    \centering
    \includegraphics[width=1\linewidth]{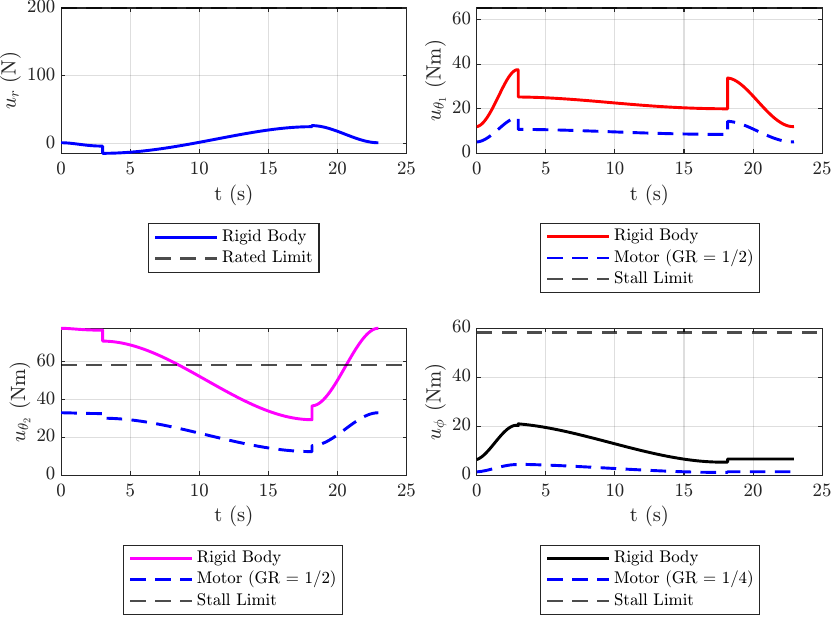}
    \caption{Force/ Torque  Characteristics. Note here however, that the stall limits in this graph were taken at the motor stalls; and not the rigid body.}
    \label{fig:placeholder}
\end{figure}
\begin{figure}
    \centering
    \includegraphics[width=\linewidth]{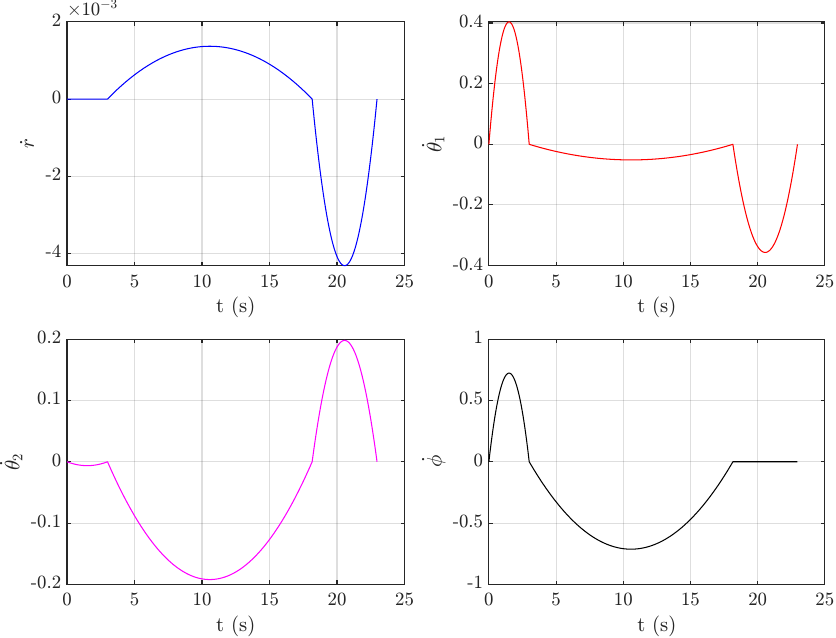}
    \caption{Velocity Trajectories}
    \label{fig:placeholder}
\end{figure}
Additionally, we find that the forces in the force plot and the velocities have a correlation; and that is that as its magnitude increase, its force decrease (which is obvious in an electromechanical setting.) 
\section{Future Work}
\subsection{Future Dynamics}
Since the robotic apparatus is not yet fully built, friction parameters cannot be measured directly. In future work, the virtual Rayleigh damping used in this project will be replaced with a more realistic friction model. The idea is to match the Rayleigh dissipation gradient to a full Stribeck friction expression so the dynamics include Coulomb friction, static friction, and velocity–dependent decay:
\begin{align*}
    \frac{\partial \mathcal{R}}{\partial \dot{q}_{i}}
    = b_{q_i}\dot{q}_i
    \;\Rightarrow\;
    \mu N\,\mathrm{sign}(\dot{q}_i)
    + \left(F_{s} - \mu N\,\mathrm{sign}(\dot{q}_i)\right)
      e^{-\frac{\|\dot{q}_i\|}{\dot{q}_s}}\\   + b_{q_i}\dot{q}_i .
\end{align*}
Once real friction data is collected from hardware, this model will replace the current virtual damping terms.

\subsection{Future Controls}
On the control side, the next stage of development will focus on making the system more adaptive and safer to operate. Following the philosophy of Gholampour et al.~\cite{gholampour2025massadaptive}, who developed a \textbf{linear} mass-adaptive controller, this work will extend the concept to a \textbf{nonlinear} dynamics framework. The robot will estimate the payload mass $m_{\mathrm{Obj}}$ directly from interaction forces so the controller can automatically adjust when grasping objects with different sizes, orientations, or inertial properties.\\
Safety constraints will be enforced using High-Order Control Barrier Functions (HOCBFs) in the spirit of Cohen’s Approximate Optimal Control framework~\cite{Cohen2020Approximate}. These constraints will help ensure feasible motion even as the estimated mass changes. Virtual servoing will also be implemented so the DC motors can follow the continuous-time control inputs produced by the optimizer. In addition to this, a PID controller will be designed around the inverse dynamics generated by the PMP–EL coupling, allowing the physical hardware to accurately track the optimal trajectories during real-world operation.\\
Mechanical limits will also be incorporated. For example, the critical mass will be used as a hardline inequality constraint. 
\begin{align*}
    m_{Obj} \leq m_{Crit} -  \delta_{m}
\end{align*}
Where $m_{Crit}$ is the critical weight (10 lbs), and $\delta_{m}$ is the slack variable for mass. 
Finally, the obstacle-avoidance constraint will also become mass-aware. 
\section{Conclusion}
In this work, a robust and computationally efficient trajectory planner has been developed for a high-resolution rigid-body model of the manipulator. The framework integrates a reduced-order Pontryagin optimal control solution, a physics-informed gradient-descent time-horizon estimator, and closed-form Euler–Lagrange inverse dynamics. Despite repeated changes to the CAD geometry and inertial parameters throughout the mechanical design process, the controller consistently produced highly accurate trajectories and demonstrated strong numerical stability across all tested configurations. For the unconstrained case, the proposed formulation serves as a certified and reliable candidate for trajectory generation, achieving smooth PMP extremals and physically consistent rigid-body motion.

The scope of the present work is limited to unconstrained optimal control, wherein the system operates without explicit joint limits, workspace restrictions, or collision-avoidance barriers. This choice enables clean closed-form derivations and highlights the intrinsic behavior of the PMP–EL architecture. Future work will incorporate joint-limit enforcement, workspace projection, and obstacle avoidance through Control Barrier Functions (CBFs) or projection-based constraint layers, extending the controller to realistic operational environments while preserving the underlying optimal structure.

\bibliography{ref.bib}

@inproceedings{Cohen2020Approximate,
  author    = {Michael H. Cohen and  et al.},
  title     = {Approximate Optimal Control for Safety-Critical Systems},
  booktitle = {Proceedings of the 60th IEEE Conference on Decision and Control (CDC)},
  year      = {2020},
  pages     = {-},
  doi       = {10.1109/CDC42340.2020.9303896},
  month     = {December},
}

@article{khatib1995inertial,
  author    = {Oussama Khatib},
  title     = {Inertial Properties in Robotic Manipulation: An Object‐Level Framework},
  journal   = {The International Journal of Robotics Research},
  volume    = {14},
  number    = {1},
  pages     = {19--36},
  year      = {1995},
  doi       = {10.1177/027836499501400103}
}

@inproceedings{Almarkhi2019Singularity,
  author    = {Ahmad A. Almarkhi and Anthony A. Maciejewski},
  title     = {Singularity Analysis for Redundant Manipulators of Arbitrary Kinematic Structure},
  booktitle = {Proceedings of the 16th International Conference on Informatics in Control, Automation and Robotics (ICINCO 2019)},
  pages     = {42--49},
  year      = {2019},
  publisher = {SCITEPRESS},
  address   = {Prague, Czech Republic},
  doi       = {},
}

@article{beaver2023flatness,
  title={Optimal Control of Differentially Flat Systems is Surprisingly Easy},
  author={Beaver, Logan E. and Malikopoulos, Andreas A.},
  journal={Automatica},
  year={2023},
  note={Preprint},
}

@book{shigley2015,
  title        = {Shigley's Mechanical Engineering Design},
  author       = {Budynas, Richard G. and Nisbett, J. Keith},
  edition      = {10},
  year         = {2015},
  publisher    = {McGraw-Hill Education},
  address      = {New York, NY},
  note         = {See Example 13--9 for a detailed example of reaction torque analysis.}
}

@techreport{marcinczyk2026appendix,
  author      = {Marcinczyk, Brock},
  title       = {Optimal Control of a 4-DOF Rigid Body Manipulator -- Appendix},
  institution = {Old Dominion University},
  year        = {2026},
  type        = {Technical Appendix},
  note        = {Supplementary material providing full Euler--Lagrange derivations and Structural Mechanics},
}

@article{gholampour2025massadaptive,
  title        = {Mass-Adaptive Admittance Control for Robotic Manipulators},
  author       = {Gholampour, Hossein and Slightam, Jonathon E. and Beaver, Logan E.},
  journal      = {arXiv preprint arXiv:2504.16224},
  year         = {2025},
  url          = {https://www.arxiv.org/abs/2504.16224}
}

@article{Canzoneri2020GDGP,
  author    = {Canzoneri, Luca and Giarr{\'e}, Laura},
  title     = {Gradient Descent-Based Task-Orientation Robot Control Enhanced With Gaussian Process Predictions},
  journal   = {IEEE Transactions on Robotics},
  year      = {2020},
  volume    = {36},
  number    = {4},
  pages     = {1120--1136},
  doi       = {10.1109/TRO.2020.2988252}
}

@misc{lemmon_ee565_module3_2015,
  author       = {Lemmon, Jonathan W.},
  title        = {Optimal Control Theory -- Module 3 -- Maximum Principle},
  howpublished = {Lecture notes, EE565, University of Notre Dame},
  year         = {2015},
  note         = {See Fig.~1, ``Trajectory $y^*$ and its perturbation.''},
}
%\printbibliography
\newpage

\onecolumn
\maketitle

\section{Optimal Control and Structurally-Informed Gradient Optimization of a Custom 4-DOF Rigid-Body Manipulator - Appendix}
\quad  This appendix provides the complete derivation of the dynamic model and supporting mechanical formulations used throughout the control architecture. The purpose of this section is not to emphasize structural mechanics for their own sake, but rather to document how the physical parameters of the manipulator define the control-affine model and the admissible input set. The appendix is divided into three parts:\\
\quad \textbf{Section A - Rigid Body Dynamics (Pg. 2-19) }
This section describes the Euler-Lagrange-based derivation using Rigid-Body Dynamics with a 4-DOF Manipulator in explicit form with no shorthand notation, using spherical coordinates. 
\[
    \dfrac{d}{dt}\left(\frac{\partial \mathcal{L}}{\partial \dot{q}_{i}}\right) - \dfrac{\partial \mathcal{L}}{\partial q_{i}} + \dfrac{\partial \mathcal{R}}{\partial \dot{q}_{i}} = Q_{j}\]
    
\textbf{Section B - Structural Mechanics and Reaction Analysis}
Although the Euler-Lagrange Solutions provide sufficient solutions for $\dot{r}_{0}, \dot{\theta}_{1,0}, \dot{\theta}_{2,0}$, the azimuthal gravity component does not exist since the azimuthal component is confined to the xy plane and thus no form of gravity is present in this plane of motion. Thus, we turn to structural mechanics using the guidance of Shigley-Style Gear Analysis for the Torque-Based Reaction in $\hat{\phi}$. \\
\textbf{Section C — Inertia Calculations}

The final component needed to complete the dynamic model is the consistent
evaluation of all rigid–body inertia tensors in the physical coordinate frame
used throughout the Euler–Lagrange formulation. Although Autodesk Inventor
provides mass properties, these quantities are reported in the CAD reference
frame and therefore must be transformed into the manipulator’s dynamic frame
before they can be used in the mass matrix $M(q)$, Coriolis matrix $C(q,\dot q)$,
or generalized forces $Q_j$.\\
With all these in mind, let us delve into the Rigid-Body Dynamics based derivation for the model.

\section*{Rigid Body Dynamics}
\begin{figure}[H]
    \centering
    \includegraphics[width=0.5\linewidth]{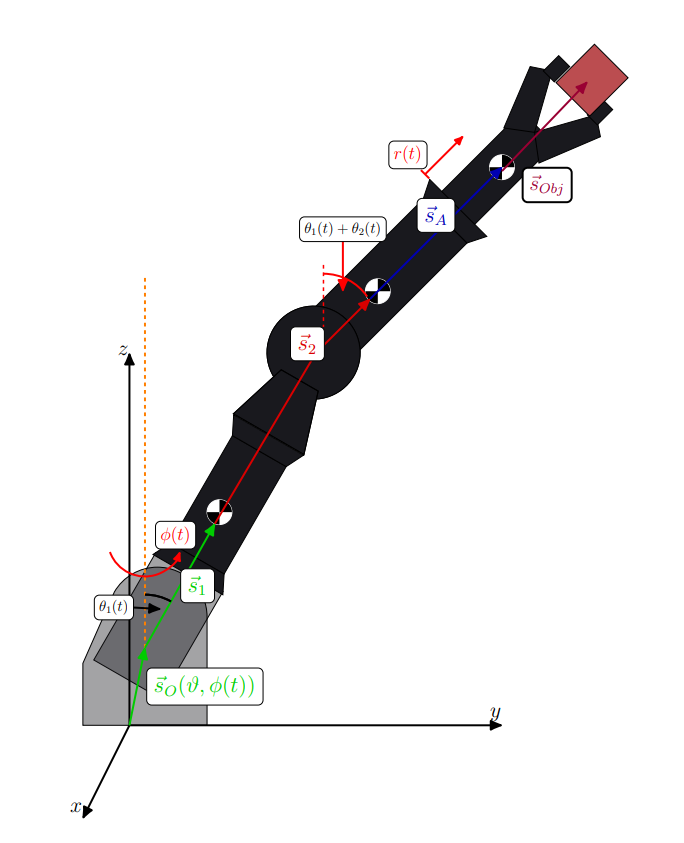}
    \caption{Free Body Diagram for Robotic Apparatus, note that the diagram holds for both the Euler-Lagrange, and the Structural Mechanics.}
    \label{fig:placeholder}
\end{figure}

Using unit vector notation, these terms are given in basis form, and for the radial coordinates: 
\begin{align*}
    e_{r_{0}} = \langle \cos\left(\phi\right)\sin\left(\vartheta\right), \sin\left(\phi\right)\sin\left(\vartheta\right), \cos\left(\vartheta\right)\rangle^{\top}\\
        e_{r_{1}} = \langle \cos\left(\phi\right)\sin\left(\theta_{1}\right), \sin\left(\phi\right)\sin\left(\theta_{1}\right), \cos\left(\theta_{1}\right)\rangle^{\top} \\ 
        e_{r_{2}} = \langle \cos\left(\phi\right)\sin\left(\theta_{12}\right), \sin\left(\phi\right)\sin\left(\theta_{12}\right), \cos\left(\theta_{12}\right)\rangle^{\top}\\
\end{align*}\
Where $e_{r_{0}}$ is the unit vector for the baseframe, $e_{r_{1}}$ is the unit vector for the first robotic link, and $e_{r_{2}}$ is for the second link. For the polar coordinates, we have the following; and note that with $\vartheta$ as a constant, we are \textbf{NOT} identifying its unit vector as it is constant and cannot be differentiated in its Rotational Kinetic Energy form without going to zero. It does have an azimuthal component, which is considered. 
\begin{align*}
        e_{\theta_{1}} = \langle \cos\left(\theta_{1}\right) \cos\left(\phi\right), \cos\left(\theta_{1}\right)\sin\left(\phi\right), -\sin\left(\theta_{1}\right) \rangle^{\top}\\
          e_{\theta_{12}} = \langle \cos\left(\theta_{12}\right) \cos\left(\phi\right), \cos\left(\theta_{12}\right)\sin\left(\phi\right), -\sin\left(\theta_{12}\right) \rangle^{\top}\\
\end{align*}
Note here, that the accumulative notation $\theta_{1,2}$ is indicative of the accumulative angle $\theta_{12} = \theta_{1} + \theta_{2}$, and finally for the azimuthal  coordinate we present the following: 
\begin{align*}
              e_{\phi} = \langle-\sin\left(\phi\right), \cos\left(\phi\right), 0 \rangle^{\top}
\end{align*}
Thus, let us define the lengths of the terms as follows: 
\begin{table}[H]
\centering
\caption{Geometric, Mass, and Inertial Parameters of the Manipulator}
\begin{tabular}{c l}
\hline
\textbf{Symbol} & \textbf{Description} \\
\hline
$\ell_{0}$          & Base-frame length measured from the mounting of the first link \\[4pt]
$\bar{\ell}_{i}$    & Center-of-mass location of link $i$ \\[4pt]
$\ell_{i}$          & Length of link $i$ \\[4pt]
$\delta_{r}$        & Distance from actuator end to object mass \\[4pt]
$R$        & Radius of the Actuator Rod \\[4pt]
$\bar{r}$           & Center of mass of the actuator rod \\[4pt]
$r'$                & COM offset term such that $(r(t)\big|_{r=\bar{r}} + r') = r_{\text{Ext}}$ \\[4pt]
$r_{\text{Ext}}$    & Total actuator stroke length \\[4pt]
$\theta_{i,\text{Ext}}$ & Virtual stroke length in the radial coordinate for joint $i$ \\[4pt]
$\phi_{\text{Ext}}$     & Virtual stroke length in the azimuthal coordinate \\[4pt]
$m_{BF}$             & Base-Frame Mass \\[4pt]
$m_{0}$             & Claw  Mass \\[4pt]
$m_{i}$             & Mass of link $i$ \\[4pt]
$I_{jk,i}$          & Diagonalized inertia tensor element of link $i$ (axes $j,k$) \\[4pt]
$m_{\text{ACT}}$    & Actuator mass \\[4pt]
$m_{\text{Obj}}$    & Object (payload) mass \\[4pt]
$h_{w} $  & Height of Gear Contact relative to base frame \\[4pt]
$H_{w} $ &  Height of Reaction at $\phi$ motor shaft relative to base frame \\[4pt]
$l_{G,i}$ & Relative azimuthal distances of the gears relative to the base frame \\[4pt]
$G_{p}$  & Applied Gear ratio of the torque (as seen with the motor.)  \\[4pt]
$\vartheta$ & Angle of the frame of reference relative to the $\theta_{1}$ mounting.    \\[4pt]
\hline
\end{tabular}
\end{table}
Thus, the position vectors for the Robotic Link, will yield the following solutions: 
\begin{align*}
    \vec{s}_{0} = \begin{bmatrix} x_{0} \\ y_{0} \\ z_{0} \end{bmatrix} = {\ell}_{0} \cdot e_{r_{0}} = {\ell}_{0} \begin{bmatrix} \cos\left(\phi\right) \sin\left(\vartheta\right) \\ \sin\left(\vartheta\right) \sin\left(\phi\right) \\  \cos\left(\vartheta\right)
    \end{bmatrix}
\end{align*}
The position vector $\vec{s}_{0}$ is measured at the baseframe, explicitly where the gearshaft has been planted within the actual module. We express the base frame position vector as the distance between the gear mounting, and the mount of the first link. Since the base-frame will maintain a uniform inertia distribution in its dynamics by itself; a parallel axis theory is then applied to relate the inertia at the end of the first position vector $\vec{s}_{0}$. The rest of the inertial values, are taken at the centers of mass. 

The first position vector is defined as follows
\begin{align*}
    & \vec{s}_1 = 
    &  \\
    & \left(\begin{array}{c} \cos\left(\phi \right)\,\left(\bar{\ell}_{1}\,\sin\left(\theta _{1}\right)+{\ell}_{0}\,\sin\left(\mathrm{\vartheta}\right)\right)\\ \sin\left(\phi \right)\,\left(\bar{\ell}_{1}\,\sin\left(\theta _{1}\right)+{\ell}_{0}\,\sin\left(\mathrm{\vartheta}\right)\right)\\ \bar{\ell}_{1}\,\cos\left(\theta _{1}\right)+{\ell}_{0}\,\cos\left(\mathrm{\vartheta}\right) \end{array}\right) \\
\end{align*}
Similary, for the positions of the link 2 center of mass, the actuator position vector, and the end-effector; are given by: 
\begin{align*}
    & \vec{s}_2 = 
    &  \\
    & \left(\begin{array}{c} \cos\left(\phi \right)\,\left(\bar{\ell}_{2}\,\sin\left(\theta _{2}+\theta _{1}\right)+{\ell}_{1}\,\sin\left(\theta _{1}\right)+{\ell}_{0}\,\sin\left(\mathrm{\vartheta}\right)\right)\\ \sin\left(\phi \right)\,\left(\bar{\ell}_{2}\,\sin\left(\theta _{2}+\theta _{1}\right)+{\ell}_{1}\,\sin\left(\theta _{1}\right)+{\ell}_{0}\,\sin\left(\mathrm{\vartheta}\right)\right)\\ \bar{\ell}_{2}\,\cos\left(\theta _{2}+\theta _{1}\right)+{\ell}_{1}\,\cos\left(\theta _{1}\right)+{\ell}_{0}\,\cos\left(\mathrm{\vartheta}\right) \end{array}\right) \\
\end{align*}

\begin{align*}
    & \vec{s}_{ACT} = 
    &  \\
    & \left(\begin{array}{c} \sin\left(\theta _{2}+\theta _{1}\right)\,\cos\left(\phi \right)\,\left(r+{\ell}_{2}\right)+{\ell}_{1}\,\cos\left(\phi \right)\,\sin\left(\theta _{1}\right)+{\ell}_{0}\,\cos\left(\phi \right)\,\sin\left(\mathrm{\vartheta}\right)\\ \sin\left(\theta _{2}+\theta _{1}\right)\,\sin\left(\phi \right)\,\left(r+{\ell}_{2}\right)+{\ell}_{1}\,\sin\left(\phi \right)\,\sin\left(\theta _{1}\right)+{\ell}_{0}\,\sin\left(\phi \right)\,\sin\left(\mathrm{\vartheta}\right)\\ \cos\left(\theta _{2}+\theta _{1}\right)\,\left(r+{\ell}_{2}\right)+{\ell}_{1}\,\cos\left(\theta _{1}\right)+{\ell}_{0}\,\cos\left(\mathrm{\vartheta}\right) \end{array}\right) \\
\end{align*}

\begin{align*}
    & \vec{s}_{Obj} = 
    &  \\
    & \left(\begin{array}{c} \sin\left(\theta _{2}+\theta _{1}\right)\,\cos\left(\phi \right)\,\left(r^{\prime }+r+{\ell}_{2}+\delta _{r}\right)+{\ell}_{1}\,\cos\left(\phi \right)\,\sin\left(\theta _{1}\right)+{\ell}_{0}\,\cos\left(\phi \right)\,\sin\left(\mathrm{\vartheta}\right)\\ \sin\left(\theta _{2}+\theta _{1}\right)\,\sin\left(\phi \right)\,\left(r^{\prime }+r+{\ell}_{2}+\delta _{r}\right)+{\ell}_{1}\,\sin\left(\phi \right)\,\sin\left(\theta _{1}\right)+{\ell}_{0}\,\sin\left(\phi \right)\,\sin\left(\mathrm{\vartheta}\right)\\ \cos\left(\theta _{2}+\theta _{1}\right)\,\left(r^{\prime }+r+{\ell}_{2}+\delta _{r}\right)+{\ell}_{1}\,\cos\left(\theta _{1}\right)+{\ell}_{0}\,\cos\left(\mathrm{\vartheta}\right) \end{array}\right) \\
\end{align*}
Taking the velocity based derivatives, notice at the base frame, it only depends on the azimuthal coordinate $\phi$ since $\vartheta$ is constant.  
\begin{align*}
    \vec{v}_{0} &= \left(\dfrac{d\vec{s}_{1}}{d\phi} \right)\dot{\phi} \\ &= \ell_{0} \dot{\phi} \cdot \begin{bmatrix} -\sin\left(\phi\right) \sin\left(\vartheta\right) \\ \cos\left(\vartheta\right) \sin\left(\phi\right) \\  0
    \end{bmatrix}
\end{align*}
The velocity vector of the first link, is given through the partial derivatives of the position vector with respect to $\phi$, and the polar coordinate $\theta_{1}$.
\begin{align*}
v_1 &= 
\frac{\partial \vec{s}_{1}}{\partial \phi}\,\dot{\phi}
\;+\;
\frac{\partial \vec{s}_{1}}{\partial \theta_{1}}\,\dot{\theta}_{1}
\\[6pt]
&=
\begin{bmatrix}
    \bar{\ell}_{1}\dot{\theta}_{1}\cos\phi\,\cos\theta_{1}
    - {\ell}_{0}\dot{\phi}\sin\phi\,\sin\vartheta
    - \bar{\ell}_{1}\dot{\phi}\sin\phi\,\sin\theta_{1}
    \\[6pt]
    \bar{\ell}_{1}\dot{\phi}\cos\phi\,\sin\theta_{1}
    + {\ell}_{0}\dot{\phi}\cos\phi\,\sin\vartheta
    + \bar{\ell}_{1}\dot{\theta}_{1}\cos\theta_{1}\,\sin\phi
    \\[6pt]
    -\bar{\ell}_{1}\,\dot{\theta}_{1}\,\sin\theta_{1}
\end{bmatrix}
\end{align*}
The velocity vector of the second link, is given through the partial derivatives of the position vector with respect to $\phi$, and the polar coordinates $\theta_{1}$ and $\theta_{2}$. 
\begin{align*}
v_2 &= 
\frac{\partial \vec{s}_{2}}{\partial \phi}\,\dot{\phi}
\;+\;
\frac{\partial \vec{s}_{2}}{\partial \theta_{1}}\,\dot{\theta}_{1}
\;+\;
\frac{\partial \vec{s}_{2}}{\partial \theta_{2}}\,\dot{\theta}_{2}
\\[6pt]
&=
\begin{bmatrix}
    \bar{\ell}_{2}\cos(\theta_{2}+\theta_{1})\cos\phi\,(\dot\theta_{2}+\dot\theta_{1})
    - {\ell}_{0}\dot{\phi}\sin\phi\,\sin\vartheta
    - {\ell}_{1}\dot{\phi}\sin\phi\,\sin\theta_{1}
    - \bar{\ell}_{2}\dot{\phi}\sin(\theta_{2}+\theta_{1})\sin\phi
    + {\ell}_{1}\dot{\theta}_{1}\cos\phi\,\cos\theta_{1}
    \\[6pt]
    \bar{\ell}_{2}\cos(\theta_{2}+\theta_{1})\sin\phi\,(\dot\theta_{2}+\dot\theta_{1})
    + \bar{\ell}_{2}\dot{\phi}\sin(\theta_{2}+\theta_{1})\cos\phi
    + {\ell}_{1}\dot{\phi}\cos\phi\,\sin\theta_{1}
    + {\ell}_{0}\dot{\phi}\cos\phi\,\sin\vartheta
    + {\ell}_{1}\dot{\theta}_{1}\cos\theta_{1}\sin\phi
    \\[6pt]
    -{\ell}_{1}\dot{\theta}_{1}\sin\theta_{1}
    - \bar{\ell}_{2}\sin(\theta_{2}+\theta_{1})(\dot\theta_{2}+\dot\theta_{1})
\end{bmatrix}
\end{align*}

\begin{align*}
\vec{v}_{\mathrm{ACT}}
&=
\frac{\partial \vec{s}_{\mathrm{ACT}}}{\partial \phi}\,\dot{\phi}
+
\frac{\partial \vec{s}_{\mathrm{ACT}}}{\partial \theta_1}\,\dot{\theta}_1
+
\frac{\partial \vec{s}_{\mathrm{ACT}}}{\partial \theta_2}\,\dot{\theta}_2
+
\frac{\partial \vec{s}_{\mathrm{ACT}}}{\partial r}\,\dot{r}.\\ =&
\begin{bmatrix}
    \dot{r}\,\sin(\theta_{2}+\theta_{1})\cos\phi
    - {\ell}_{1}\dot{\phi}\sin\phi\,\sin\theta_{1}
    - {\ell}_{0}\dot{\phi}\sin\phi\,\sin\vartheta
    + \cos(\theta_{2}+\theta_{1})\cos\phi\,(r+{\ell}_{2})(\dot\theta_{2}+\dot\theta_{1})
    \\[-2pt]
    \qquad
    - \dot{\phi}\sin(\theta_{2}+\theta_{1})\sin\phi\,(r+{\ell}_{2})
    + {\ell}_{1}\dot{\theta}_{1}\cos\phi\,\cos\theta_{1}
    \\[8pt]
    \dot{r}\,\sin(\theta_{2}+\theta_{1})\sin\phi
    + \cos(\theta_{2}+\theta_{1})\sin\phi\,(r+{\ell}_{2})(\dot\theta_{2}+\dot\theta_{1})
    + \dot{\phi}\sin(\theta_{2}+\theta_{1})\cos\phi\,(r+{\ell}_{2})
    \\[-2pt]
    \qquad
    + {\ell}_{1}\dot{\phi}\cos\phi\,\sin\theta_{1}
    + {\ell}_{0}\dot{\phi}\cos\phi\,\sin\vartheta
    + {\ell}_{1}\dot{\theta}_{1}\cos\theta_{1}\sin\phi
    \\[8pt]
    \dot{r}\,\cos(\theta_{2}+\theta_{1})
    - {\ell}_{1}\dot{\theta}_{1}\sin\theta_{1}
    - \sin(\theta_{2}+\theta_{1})(r+{\ell}_{2})(\dot\theta_{2}+\dot\theta_{1})
\end{bmatrix}
\end{align*}
\begin{align*}
\vec{v}_{\mathrm{Obj}}
&= \frac{\partial \vec{s}_{\mathrm{Obj}}}{\partial \phi}\,\dot{\phi}
+
\frac{\partial \vec{s}_{\mathrm{Obj}}}{\partial \theta_1}\,\dot{\theta}_1
+
\frac{\partial \vec{s}_{\mathrm{Obj}}}{\partial \theta_2}\,\dot{\theta}_2
+
\frac{\partial \vec{s}_{\mathrm{Obj}}}{\partial r}\,\dot r \\=&
\begin{bmatrix}
    \dot{r}\sin(\theta_{2}+\theta_{1})\cos\phi
    - \ell_{1}\dot{\phi}\sin\phi\,\sin\theta_{1}
    - \ell_{0}\dot{\phi}\sin\phi\,\sin\vartheta
    \\[4pt]
    \qquad
    + \cos(\theta_{2}+\theta_{1})\cos\phi\,(\dot{\theta}_2+\dot{\theta}_1)(r' + r + \ell_2 + \delta_r)
    \\[-2pt]
    \qquad
    - \dot{\phi}\sin(\theta_{2}+\theta_{1})\sin\phi\,(r' + r + \ell_2 + \delta_r)
    + \ell_{1}\dot{\theta}_{1}\cos\phi\cos\theta_{1}
    \\[10pt]
    \dot{r}\sin(\theta_{2}+\theta_{1})\sin\phi
    + \cos(\theta_{2}+\theta_{1})\sin\phi\,(\dot{\theta}_2+\dot{\theta}_1)(r' + r + \ell_2 + \delta_r)
    \\[-2pt]
    \qquad
    + \dot{\phi}\sin(\theta_{2}+\theta_{1})\cos\phi\,(r' + r + \ell_2 + \delta_r)
    + \ell_{1}\dot{\phi}\cos\phi\sin\theta_{1}
    \\[-2pt]
    \qquad
    + \ell_{0}\dot{\phi}\cos\phi\sin\vartheta
    + \ell_{1}\dot{\theta}_{1}\cos\theta_{1}\sin\phi
    \\[10pt]
    \dot{r}\cos(\theta_{2}+\theta_{1})
    - \ell_{1}\dot{\theta}_{1}\sin\theta_{1}
    \\[-2pt]
    \qquad
    - \sin(\theta_{2}+\theta_{1})(r' + r + \ell_2 + \delta_r)(\dot{\theta}_2+\dot{\theta}_1)
\end{bmatrix}
\end{align*}
Let us now define the Moments of Inertia in terms of the Joint Coordinates $I_{\theta_{1}}, I_{\theta_{2}}, J_{\phi}$, to do this; we use the polar and azimuth basis vectors to couple the moments of inertia to the states. 
% ============================================================
%  State–Coupled Rotational Inertias
% ============================================================

\subsection*{State–Coupled Polar Inertias}

The rotational inertia of each link about its generalized coordinate 
is obtained through the quadratic form
\begin{align*}
    I(q) = e_{q}^{\top} I \, e_{q},
\end{align*}
where $e_{q}$ is the unit vector associated with the generalized rotational
coordinate and $I$ is the link's diagonal inertia tensor.
\subsubsection*{Base Frame Azimuthal Inertia}
\begin{align*}
    J_{0} = e_{\phi}^{\top} \, I_{0} \, e_{\phi} = I_{xx,0}\,\sin^{2}\phi + I_{yy,0}\cos^{2}\phi
\end{align*}
% ------------------ I1 ---------------------
\subsubsection*{Link--1 Polar Inertia}
\begin{align*}
    I_{1} &= e_{\theta_{1}}^{\top} \, I_{1} \, e_{\theta_{1}}
    \\[4pt]
    &=
    \sin^{2}\!\left(\theta_{1}\right)\, I_{\mathrm{zz},1}
    + \cos^{2}\!\left(\theta_{1}\right)\sin^{2}\!\left(\phi\right)\, I_{\mathrm{yy},1}
    + \cos^{2}\!\left(\theta_{1}\right)\cos^{2}\!\left(\phi\right)\, I_{\mathrm{xx},1}.
\end{align*}

% ------------------ I2 ---------------------
\subsubsection*{Link--2 Polar Inertia}
\begin{align*}
    I_{2} &= e_{\theta_{12}}^{\top} \, I_{2} \, e_{\theta_{12}}
    \\[4pt]
    &=
    \sin^{2}\!\left(\theta_{1}+\theta_{2}\right)\, I_{\mathrm{zz},2}
    + \cos^{2}\!\left(\theta_{1}+\theta_{2}\right)\sin^{2}\!\left(\phi\right)\, I_{\mathrm{yy},2}
    + \cos^{2}\!\left(\theta_{1}+\theta_{2}\right)\cos^{2}\!\left(\phi\right)\, I_{\mathrm{xx},2}.
\end{align*}

% ============================================================
%  Actuator Inertia Tensor from Triple Integral
% ============================================================

\subsubsection*{Actuator Inertia Tensor}

The actuator inertia tensor results from evaluating the cylindrical
triple–integrals
\[
I_{ii}^{\mathrm{ACT}}
=
\varrho_{ACT}\int_{-r/2}^{r/2}
\int_{0}^{2\pi}
\int_{0}^{R}
\rho
\begin{bmatrix}
Y^{2}+Z^{2} & 0 & 0 \\
0 & X^{2}+Z^{2} & 0 \\
0 & 0 & X^{2}+Y^{2}
\end{bmatrix}
\, d\rho\, d\alpha\, dZ,
\]
which yield
\begin{align*}
I_{ii}^{\mathrm{ACT}}
&=
\begin{bmatrix}
\displaystyle 
\frac{\pi}{12} R^{2} r\,\varrho_{\mathrm{ACT}} \left(3R^{2} + r^{2}\right)
& 0 & 0
\\[6pt]
0 &
\displaystyle 
\frac{\pi}{12} R^{2} r\,\varrho_{\mathrm{ACT}} \left(3R^{2} + r^{2}\right)
& 0
\\[6pt]
0 & 0 &
\displaystyle 
\frac{\pi}{2} R^{4} r\,\varrho_{\mathrm{ACT}}
\end{bmatrix}.
\end{align*}

% ------------------ I_ACT (theta) ---------------------
\subsection{Actuator Polar Inertia}
Applying the $e_{\theta_{12}}$ projection yields
\begin{align*}
    & I_{ACT} = 
     0.2618\,R^2\,r\,\varrho_{\mathrm{ACT}}\,\left(3\,R^2\,{\sin\left(\theta _{2}+\theta _{1}\right)}^2-r^2\,{\sin\left(\theta _{2}+\theta _{1}\right)}^2+3\,R^2+r^2\right) \\
\end{align*}
\subsection*{Azimuthal Inertia}
The momentum of inertia about the azimuthal coordinate is computed via
\begin{align*}
J_{\mathrm{eq}}
= e_{\phi}^{\top}
\left( I_{0} + I_{1} + I_{2} + I_{\mathrm{ACT}} \right)
e_{\phi},
\end{align*}
with $e_{\phi} = [-\sin\phi,\; \cos\phi,\; 0]^{\top}$.
Note that since there is only one azimuthal coordinate $\phi$, Koning's Decomposition Theory does not apply since there exists only one azimuthal coordinate $\phi$. 
Using the diagonal form of each tensor,
\[
e_{\phi}^{\top}
\operatorname{diag}(I_{xx,i}, I_{yy,i}, I_{zz,i})
e_{\phi}
=
I_{xx,i}\cos^{2}\phi + I_{yy,i}\sin^{2}\phi,
\]
the full equivalent inertia is
\begin{align*}
J_{\mathrm{eq}}
&=
\left(I_{\mathrm{xx},0}\cos^{2}\phi + I_{\mathrm{yy},0}\sin^{2}\phi\right)
+
\left(I_{\mathrm{xx},1}\cos^{2}\phi + I_{\mathrm{yy},1}\sin^{2}\phi\right)
+
\left(I_{\mathrm{xx},2}\cos^{2}\phi + I_{\mathrm{yy},2}\sin^{2}\phi\right)
\\[6pt]
&\quad
+ \frac{\pi}{12}\,R^{2} r^{3}\varrho_{\mathrm{ACT}}
+ \frac{\pi}{4}\,R^{4} r\,\varrho_{\mathrm{ACT}}.
\end{align*}
\subsection{Kinetic Energy}
Now that the velocities are defined, in addition to the inertial distributions; we can define its kinetic energy as follows:
\begin{align*}
T &= \tfrac{1}{2} m_1 \mathbf{v}_1^\top \mathbf{v}_1
   + \tfrac{1}{2} m_2 \mathbf{v}_2^\top \mathbf{v}_2\notag \\
  & + \tfrac{1}{2} m_{\mathrm{Rod}} \mathbf{v}_A^\top \mathbf{v}_A
   + \tfrac{1}{2} m_{\mathrm{obj}} \mathbf{v}_{\mathrm{obj}}^\top \mathbf{v}_{\mathrm{obj}} \notag \\
  &\quad + \tfrac{1}{2} I_{\theta_1} \dot{\theta}_1^2
   + \tfrac{1}{2} J_1 \dot{\phi}^2
   + \tfrac{1}{2} I_{\theta_2} (\dot{\theta}_1+\dot{\theta}_2)^2
   + \tfrac{1}{2} J_2 \dot{\phi}^2 \\
&=
\Bigg[
    \tfrac{1}{2} I_{\mathrm{yy},0}
    + \tfrac{1}{2} I_{\mathrm{yy},1}
    + \tfrac{1}{2} I_{\mathrm{yy},2}
    + \tfrac{1}{2} I_{\mathrm{xx},0} \sin^{2}\!\phi
    + \tfrac{1}{2} I_{\mathrm{xx},1} \sin^{2}\!\phi
    + \tfrac{1}{2} I_{\mathrm{xx},2} \sin^{2}\!\phi
\\
&\qquad 
    - \tfrac{1}{2} I_{\mathrm{yy},0} \sin^{2}\!\phi
    - \tfrac{1}{2} I_{\mathrm{yy},1} \sin^{2}\!\phi
    - \tfrac{1}{2} I_{\mathrm{yy},2} \sin^{2}\!\phi
    + 0.1309\,R^{2} r^{3}\varrho_{\mathrm{ACT}}
    + 0.3927\,R^{4} r\,\varrho_{\mathrm{ACT}}
\Bigg]\dot{\phi}^{2}
\\[10pt]
&\quad
+ \frac{1}{2} m_{1}
\left(
    \bar \ell_{1}\dot{\phi}\cos\phi\sin\theta_{1}
    + \ell_{0}\dot{\phi}\cos\phi\sin\vartheta
    + \bar \ell_{1}\dot{\theta}_{1}\cos\theta_{1}\sin\phi
\right)^{2}
\\[6pt]
&\quad
+ \dot{\theta}_{1}^{2}
\left(
    \tfrac{1}{2}\sin^{2}\!\theta_{1} \, I_{\mathrm{zz},1}
    + \tfrac{1}{2}\cos^{2}\!\theta_{1}\sin^{2}\!\phi \, I_{\mathrm{yy},1}
    + \tfrac{1}{2}\cos^{2}\!\theta_{1}\cos^{2}\!\phi \, I_{\mathrm{xx},1}
\right)
\\[6pt]
&\quad
+ \frac{1}{2} m_{1}
\left(
    \bar \ell_{1}\dot{\phi}\sin\phi\sin\theta_{1}
    + \ell_{0}\dot{\phi}\sin\phi\sin\vartheta
    - \bar \ell_{1}\dot{\theta}_{1}\cos\phi\cos\theta_{1}
\right)^{2}
\\[6pt]
&\quad
+ \frac{1}{2} m_{2}
\left(
    \ell_{1}\dot{\theta}_{1}\sin\theta_{1}
    + \bar \ell_{2}\dot{\theta}_{1}\sin(\theta_{1}+\theta_{2})
    + \bar \ell_{2}\dot{\theta}_{2}\sin(\theta_{1}+\theta_{2})
\right)^{2}
\\[6pt]
&\quad
+ \frac{1}{2}(m_{\mathrm{Obj}} + m_{0})
\left(
    \ell_{1}\dot{\theta}_{1}\sin\theta_{1}
    - \dot{r}\cos(\theta_{1}+\theta_{2})
    + \sin(\theta_{1}+\theta_{2})(\dot{\theta}_{1}+\dot{\theta}_{2})(r' + r + \ell_{2} + \delta_{r})
\right)^{2}
\\[6pt]
&\quad
+ \frac{1}{2}(m_{\mathrm{Obj}} + m_{0})
\left(
    \dot{r}\sin(\theta_{1}+\theta_{2})\sin\phi
    + \cos(\theta_{1}+\theta_{2})\sin\phi(\dot{\theta}_{1}+\dot{\theta}_{2})(r' + r + \ell_{2} + \delta_{r})
\right.
\\
&\qquad\qquad\qquad
\left.
    + \dot{\phi}\sin(\theta_{1}+\theta_{2})\cos\phi(r' + r + \ell_{2} + \delta_{r})
    + \ell_{1}\dot{\phi}\cos\phi\sin\theta_{1}
    + \ell_{0}\dot{\phi}\cos\phi\sin\vartheta
    + \ell_{1}\dot{\theta}_{1}\cos\theta_{1}\sin\phi
\right)^{2}
\\[6pt]
&\quad
+ \frac{1}{2}(m_{\mathrm{Obj}} + m_{0})
\left(
    \dot{r}\sin(\theta_{1}+\theta_{2})\cos\phi
    - \ell_{1}\dot{\phi}\sin\phi\sin\theta_{1}
\right.
\\
&\qquad\qquad\qquad
\left.
    - \ell_{0}\dot{\phi}\sin\phi\sin\vartheta
    + \cos(\theta_{1}+\theta_{2})\cos\phi(\dot{\theta}_{1}+\dot{\theta}_{2})(r' + r + \ell_{2} + \delta_{r})
\right.
\\
&\qquad\qquad\qquad
\left.
    - \dot{\phi}\sin(\theta_{1}+\theta_{2})\sin\phi(r' + r + \ell_{2} + \delta_{r})
    + \ell_{1}\dot{\theta}_{1}\cos\phi\cos\theta_{1}
\right)^{2}
\\[6pt]
&\quad
+ (\dot{\theta}_{1}+\dot{\theta}_{2})^{2}
\left(
    \tfrac{1}{2}\sin^{2}(\theta_{1}+\theta_{2}) I_{\mathrm{zz},2}
    + \tfrac{1}{2}\cos^{2}(\theta_{1}+\theta_{2})\sin^{2}\phi \, I_{\mathrm{yy},2}
    + \tfrac{1}{2}\cos^{2}(\theta_{1}+\theta_{2})\cos^{2}\phi \, I_{\mathrm{xx},2}
\right)
\\[6pt]
&\quad
+ \frac{1}{2} m_{\mathrm{ACT}}
\left(
    \ell_{1}\dot{\theta}_{1}\sin\theta_{1}
    - \dot{r}\cos(\theta_{1}+\theta_{2})
    + \sin(\theta_{1}+\theta_{2})(r + \ell_{2})(\dot{\theta}_{1}+\dot{\theta}_{2})
\right)^{2}
\\[6pt]
&\quad
+ \frac{1}{2} m_{2}
\left(
    \bar \ell_{2}\cos(\theta_{1}+\theta_{2})\sin\phi(\dot{\theta}_{1}+\dot{\theta}_{2})
    + \bar \ell_{2}\dot{\phi}\sin(\theta_{1}+\theta_{2})\cos\phi
\right.
\\
&\qquad\qquad\qquad
\left.
    + \ell_{1}\dot{\phi}\cos\phi\sin\theta `_{1}
    + \ell_{0}\dot{\phi}\cos\phi\sin\vartheta
    + \ell_{1}\dot{\theta}_{1}\cos\theta_{1}\sin\phi
\right)^{2}
\\[6pt]
&\quad
+ \frac{1}{2} m_{2}
\left(
    \ell_{1}\dot{\phi}\sin\phi\sin\theta_{1}
    + \ell_{0}\dot{\phi}\sin\phi\sin\vartheta
    - \bar \ell_{2}\cos(\theta_{1}+\theta_{2})\cos\phi(\dot{\theta}_{1}+\dot{\theta}_{2})
\right.
\\
&\qquad\qquad\qquad
\left.
    + \bar \ell_{2}\dot{\phi}\sin(\theta_{1}+\theta_{2})\sin\phi
    - \ell_{1}\dot{\theta}_{1}\cos\phi\cos\theta_{1}
\right)^{2}
\end{align*}

\newpage
\begin{align*}
&\quad
\frac{1}{2} m_{\mathrm{ACT}}
\Big(
    \dot r\,\sin(\theta_{1}+\theta_{2})\sin\phi
    + \cos(\theta_{1}+\theta_{2})\sin\phi\,(r+\ell_{2})(\dot\theta_{1}+\dot\theta_{2})
\\
&\qquad\qquad
    + \dot\phi\,\sin(\theta_{1}+\theta_{2})\cos\phi\,(r+\ell_{2})
    + \ell_{1}\dot\phi\cos\phi\sin\theta_{1}
    + \ell_{0}\dot\phi\cos\phi\sin\vartheta
    + \ell_{1}\dot\theta_{1}\cos\theta_{1}\sin\phi
\Big)^{2}
\\[8pt]
&\quad
+ \frac{1}{2} m_{\mathrm{ACT}}
\Big(
    \dot r\,\sin(\theta_{1}+\theta_{2})\cos\phi
    - \ell_{1}\dot\phi\sin\phi\sin\theta_{1}
    - \ell_{0}\dot\phi\sin\phi\sin\vartheta
\\
&\qquad\qquad
    + \cos(\theta_{1}+\theta_{2})\cos\phi\,(r+\ell_{2})(\dot\theta_{1}+\dot\theta_{2})
    - \dot\phi\sin(\theta_{1}+\theta_{2})\sin\phi\,(r+\ell_{2})
    + \ell_{1}\dot\theta_{1}\cos\phi\cos\theta_{1}
\Big)^{2}
\\[8pt]
&\quad
+ \frac{1}{2}\,\bar \ell_{1}^{2} \, m_{1}\, \dot{\theta}_{1}^{2}\, \sin^{2}\theta_{1}
\\[8pt]
&\quad
+ 0.1309\,R^{2} r\, \varrho_{\mathrm{ACT}} 
(\dot\theta_{1}+\dot\theta_{2})^{2}
\Big(
    3R^{2}\sin^{2}(\theta_{1}+\theta_{2})
    - r^{2}\sin^{2}(\theta_{1}+\theta_{2})
    + 3R^{2}
    + r^{2}
\Big)
\\[8pt]
&\quad
+ \frac{1}{2} \, \ell_{0}^{2} m_{\mathrm{BF}} \dot\phi^{2}\cos^{2}\phi\,\sin^{2}\vartheta
+ \frac{1}{2} \, \ell_{0}^{2} m_{\mathrm{BF}} \dot\phi^{2}\sin^{2}\phi\,\sin^{2}\vartheta .
\end{align*}
\subsection{Potential Energy}
The Potential Energy, is given through the Gravitational Potential Energy, and the Elastic Potential Energy; which the springs are defined as their rated store-brand force/ torque, over its stroke length/ angle.

\begin{align*}
V &= 
\frac{1}{2}\kappa_{\phi}(\phi-\phi_0)^2
+ \frac{1}{2}\kappa_r(r-r_0)^2
+ \frac{1}{2}\kappa_{\theta_1}(\theta_1-\theta_{1,0})^2
+ \frac{1}{2}\kappa_{\theta_2}(\theta_2-\theta_{2,0})^2
\notag\\[6pt]
&\quad
=g\,m_{\mathrm{ACT}}
   \bigl(\cos(\theta_1+\theta_2)(r+\ell_2)
        + \ell_1\cos\theta_1 + \ell_0\cos\vartheta \bigr)
\notag\\
&\quad
+ g\,m_2
   \bigl(\bar \ell_2\cos(\theta_1+\theta_2)
        + \ell_1\cos\theta_1 + \ell_0\cos\vartheta \bigr)
\notag\\
&\quad
+ g\,(m_{\mathrm{obj}}+m_0)
   \bigl(\cos(\theta_1+\theta_2)(r' + r + \ell_2 + \delta_r)
        + \ell_1\cos\theta_1 + \ell_0\cos\vartheta\bigr)
\notag\\
&\quad
+ g\,m_1
   \bigl(\bar \ell_1\cos\theta_1 + \ell_0\cos\vartheta \bigr)
\notag\\
+ \cancel{g\,\ell_0\,m_{\mathrm{BF}}\cos\vartheta}^{ = 0}.
\end{align*}
Note here, that the gravitational potential energy in $g\,\ell_0\,m_{\mathrm{BF}}\cos\vartheta$, its gradient with respect to the manipulator’s generalized coordinates is identically zero.
Consequently, it plays no role in the Euler–Lagrange equations.
This reflects physical reality: a rigidly mounted base does not “participate” in the robot’s motion in the form of gravity—it simply holds the rest of the system up, as any respectable structural component should.
All dynamically relevant gravitational effects arise solely from the moving masses, whose contributions appear through the Jacobians of their position vectors, encoding the familiar parallel-axis behavior.
\subsection{Rayeligh Dissipation}
For the Rayleigh Damping, we use virtual damping provided by the virtual springs (which are assumed as critical), which yields: 
\begin{align*}
    \kappa_{q_{i}} = \frac{u_{Rated}}{q_{Stroke}} \Rightarrow b_{q_{i}} = 2\sqrt{\frac{M_{q_{i}} u_{rated}}{q_{Stroke}}} 
\end{align*}
\indent In future work, damping and friction will be extended beyond the virtual–spring model, with a full Stribeck friction law incorporated directly into the Rayleigh dissipation gradient of the Euler–Lagrange equations, as it constitutes a physically dissipative term. The virtual–spring approximation behaves well for the rotational joints once gear ratios are applied. The radial coordinate, however, is inherently slower (the linear actuator operates at 
4mm/s under no loading), which allows the virtual spring to capture the admissible stroke range but makes unrestricted velocity regulation more difficult. Nevertheless, when the Gradient–Descent/optimal–control kinematics are injected into the Euler–Lagrange dynamics, the resulting forces remain within feasible limits due to the stabilizing influence of the virtual stiffness.
\begin{align*}
    & \mathcal{R} = 
     \frac{1}{2}\,b_{r}\,{\dot{r}}^2+\frac{1}{2}\,b_{\phi }\,{\dot{\phi }}^2+\frac{1}{2}\,b_{\theta ,2}\,{\dot{\theta }_{2}}^2+\,b_{\theta ,1}\,{\dot{\theta }_{1}}^2 \\
\end{align*}
In the solutions, since the friction model will be updated; the solution in explicit form in the Euler Lagrange solutions, for the dissipative component; will be left as follows: $\dfrac{\partial \mathcal{R}}{\partial \dot{q}_{i}}$, since this term will evolve as experimentation goes onwards. 

\subsection{Euler-Lagrange}
Now that the Kinetic and Potential energy is defined, in addition to the Rayleigh Damping, let us now take the Euler-Lagrange Equations of Motion to find the closed-form dynamic torques; applying:
\[\dfrac{d}{dt}\left(\frac{\partial \mathcal{L}}{\partial \dot{q}_{i}}\right) - \dfrac{\partial \mathcal{L}}{\partial q_{i}} + \dfrac{\partial \mathcal{R}}{\partial \dot{q}_{i}} = Q_{j}\]. 
The Euler-Lagrange solutions are given on pages 8 through  17. For clarity and brevity, our discussion focuses on the final equations of motion rather than the intermediate algebra for the Euler-Lagrange equations.
\subsubsection*{Radial Coordinate}
\begin{align*}
    Q_r = u_r =& \frac{d}{dt}\left(\frac{\partial \mathcal{L}}{\partial \dot{r}}\right) - \dfrac{\partial \mathcal{L}}{\partial r} + \dfrac{\partial \mathcal{R}}{\partial \dot{r}} \\ 
    u_{r} =   & \frac{\partial \mathcal{R}}{\partial \dot{r}}+\kappa _{r}\,r-\kappa _{r}\,r_{0}+m_{0}\,\ddot{r}+m_{\mathrm{ACT}}\,\ddot{r}+m_{\mathrm{Obj}}\,\ddot{r}-0.5000\,\delta _{r}\,m_{0}\,{\dot{\phi }}^2-0.5000\,\delta _{r}\,m_{\mathrm{Obj}}\,{\dot{\phi }}^2-\delta _{r}\,m_{0}\,{\dot{\theta }_{1}}^2- \\
    & \delta _{r}\,m_{0}\,{\dot{\theta }_{2}}^2-\delta _{r}\,m_{\mathrm{Obj}}\,{\dot{\theta }_{1}}^2-\delta _{r}\,m_{\mathrm{Obj}}\,{\dot{\theta }_{2}}^2-0.5000\,\ell_{2}\,m_{0}\,{\dot{\phi }}^2-0.5000\,\ell_{2}\,m_{\mathrm{ACT}}\,{\dot{\phi }}^2- \\
    & 0.5000\,\ell_{2}\,m_{\mathrm{Obj}}\,{\dot{\phi }}^2-\ell_{2}\,m_{0}\,{\dot{\theta }_{1}}^2-\ell_{2}\,m_{0}\,{\dot{\theta }_{2}}^2-\ell_{2}\,m_{\mathrm{ACT}}\,{\dot{\theta }_{1}}^2-\ell_{2}\,m_{\mathrm{ACT}}\,{\dot{\theta }_{2}}^2-\ell_{2}\,m_{\mathrm{Obj}}\,{\dot{\theta }_{1}}^2- \\
    & \ell_{2}\,m_{\mathrm{Obj}}\,{\dot{\theta }_{2}}^2-0.5000\,m_{0}\,{\dot{\phi }}^2\,r-0.5000\,m_{\mathrm{ACT}}\,{\dot{\phi }}^2\,r-0.5000\,m_{\mathrm{Obj}}\,{\dot{\phi }}^2\,r-0.5000\,m_{0}\,{\dot{\phi }}^2\,r^{\prime }-0.5000\,m_{\mathrm{Obj}}\,{\dot{\phi }}^2\,r^{\prime }- \\
    & m_{0}\,r\,{\dot{\theta }_{1}}^2-m_{0}\,r\,{\dot{\theta }_{2}}^2-m_{\mathrm{ACT}}\,r\,{\dot{\theta }_{1}}^2-m_{\mathrm{ACT}}\,r\,{\dot{\theta }_{2}}^2-m_{\mathrm{Obj}}\,r\,{\dot{\theta }_{1}}^2-m_{\mathrm{Obj}}\,r\,{\dot{\theta }_{2}}^2- \\
    & m_{0}\,r^{\prime }\,{\dot{\theta }_{1}}^2-m_{0}\,r^{\prime }\,{\dot{\theta }_{2}}^2-m_{\mathrm{Obj}}\,r^{\prime }\,{\dot{\theta }_{1}}^2-m_{\mathrm{Obj}}\,r^{\prime }\,{\dot{\theta }_{2}}^2-0.3927\,R^4\,{\dot{\phi }}^2\,\varrho _{\mathrm{ACT}}- \\
    & 0.5890\,R^4\,\varrho _{\mathrm{ACT}}\,{\dot{\theta }_{1}}^2-0.5890\,R^4\,\varrho _{\mathrm{ACT}}\,{\dot{\theta }_{2}}^2+g\,m_{0}\,\cos\left(\theta _{2}+\theta _{1}\right)+g\,m_{\mathrm{ACT}}\,\cos\left(\theta _{2}+\theta _{1}\right)+ \\
    & g\,m_{\mathrm{Obj}}\,\cos\left(\theta _{2}+\theta _{1}\right)+0.5000\,\ell_{1}\,m_{0}\,{\dot{\phi }}^2\,\cos\left(\theta _{2}+2\,\theta _{1}\right)+0.5000\,\ell_{1}\,m_{\mathrm{ACT}}\,{\dot{\phi }}^2\,\cos\left(\theta _{2}+2\,\theta _{1}\right)+ \\
    & 0.5000\,\ell_{1}\,m_{\mathrm{Obj}}\,{\dot{\phi }}^2\,\cos\left(\theta _{2}+2\,\theta _{1}\right)-2\,\delta _{r}\,m_{0}\,\dot{\theta }_{1}\,\dot{\theta }_{2}-2\,\delta _{r}\,m_{\mathrm{Obj}}\,\dot{\theta }_{1}\,\dot{\theta }_{2}- \\
    & 2\,\ell_{2}\,m_{0}\,\dot{\theta }_{1}\,\dot{\theta }_{2}-2\,\ell_{2}\,m_{\mathrm{ACT}}\,\dot{\theta }_{1}\,\dot{\theta }_{2}-2\,\ell_{2}\,m_{\mathrm{Obj}}\,\dot{\theta }_{1}\,\dot{\theta }_{2}-0.3927\,R^2\,{\dot{\phi }}^2\,r^2\,\varrho _{\mathrm{ACT}}- \\
    & 2\,m_{0}\,r\,\dot{\theta }_{1}\,\dot{\theta }_{2}-2\,m_{\mathrm{ACT}}\,r\,\dot{\theta }_{1}\,\dot{\theta }_{2}-2\,m_{\mathrm{Obj}}\,r\,\dot{\theta }_{1}\,\dot{\theta }_{2}-2\,m_{0}\,r^{\prime }\,\dot{\theta }_{1}\,\dot{\theta }_{2}- \\
    & 2\,m_{\mathrm{Obj}}\,r^{\prime }\,\dot{\theta }_{1}\,\dot{\theta }_{2}-0.1963\,R^2\,r^2\,\varrho _{\mathrm{ACT}}\,{\dot{\theta }_{1}}^2-0.1963\,R^2\,r^2\,\varrho _{\mathrm{ACT}}\,{\dot{\theta }_{2}}^2+ \\
    & 0.5000\,\ell_{0}\,m_{0}\,{\dot{\phi }}^2\,\cos\left(\mathrm{\vartheta}+\theta _{2}+\theta _{1}\right)+0.5000\,\ell_{0}\,m_{\mathrm{ACT}}\,{\dot{\phi }}^2\,\cos\left(\mathrm{\vartheta}+\theta _{2}+\theta _{1}\right)+ \\
    & 0.5000\,\ell_{0}\,m_{\mathrm{Obj}}\,{\dot{\phi }}^2\,\cos\left(\mathrm{\vartheta}+\theta _{2}+\theta _{1}\right)+0.5000\,\delta _{r}\,m_{0}\,{\dot{\phi }}^2\,\cos\left(2\,\theta _{2}+2\,\theta _{1}\right)+ \\
    & 0.5000\,\delta _{r}\,m_{\mathrm{Obj}}\,{\dot{\phi }}^2\,\cos\left(2\,\theta _{2}+2\,\theta _{1}\right)-1.1781\,R^4\,\varrho _{\mathrm{ACT}}\,\dot{\theta }_{1}\,\dot{\theta }_{2}+0.5000\,\ell_{2}\,m_{0}\,{\dot{\phi }}^2\,\cos\left(2\,\theta _{2}+2\,\theta _{1}\right)+ \\
    & 0.5000\,\ell_{2}\,m_{\mathrm{ACT}}\,{\dot{\phi }}^2\,\cos\left(2\,\theta _{2}+2\,\theta _{1}\right)+0.5000\,\ell_{2}\,m_{\mathrm{Obj}}\,{\dot{\phi }}^2\,\cos\left(2\,\theta _{2}+2\,\theta _{1}\right)+ \\
    & 0.5000\,m_{0}\,{\dot{\phi }}^2\,r\,\cos\left(2\,\theta _{2}+2\,\theta _{1}\right)+0.5000\,m_{\mathrm{ACT}}\,{\dot{\phi }}^2\,r\,\cos\left(2\,\theta _{2}+2\,\theta _{1}\right)+0.5000\,m_{\mathrm{Obj}}\,{\dot{\phi }}^2\,r\,\cos\left(2\,\theta _{2}+2\,\theta _{1}\right)+ \\
    & 0.5000\,m_{0}\,{\dot{\phi }}^2\,r^{\prime }\,\cos\left(2\,\theta _{2}+2\,\theta _{1}\right)+0.5000\,m_{\mathrm{Obj}}\,{\dot{\phi }}^2\,r^{\prime }\,\cos\left(2\,\theta _{2}+2\,\theta _{1}\right)- \\
    & 0.5000\,\ell_{0}\,m_{0}\,{\dot{\phi }}^2\,\cos\left(-\mathrm{\vartheta}+\theta _{2}+\theta _{1}\right)-0.5000\,\ell_{0}\,m_{\mathrm{ACT}}\,{\dot{\phi }}^2\,\cos\left(-\mathrm{\vartheta}+\theta _{2}+\theta _{1}\right)- \\
    & 0.5000\,\ell_{0}\,m_{\mathrm{Obj}}\,{\dot{\phi }}^2\,\cos\left(-\mathrm{\vartheta}+\theta _{2}+\theta _{1}\right)+\ell_{1}\,m_{0}\,\ddot{\theta }_{1}\,\sin\left(\theta _{2}\right)+\ell_{1}\,m_{\mathrm{ACT}}\,\ddot{\theta }_{1}\,\sin\left(\theta _{2}\right)+ \\
    & \ell_{1}\,m_{\mathrm{Obj}}\,\ddot{\theta }_{1}\,\sin\left(\theta _{2}\right)+0.1963\,R^4\,\varrho _{\mathrm{ACT}}\,{\dot{\theta }_{1}}^2\,\cos\left(2\,\theta _{2}+2\,\theta _{1}\right)+ \\
    & 0.1963\,R^4\,\varrho _{\mathrm{ACT}}\,{\dot{\theta }_{2}}^2\,\cos\left(2\,\theta _{2}+2\,\theta _{1}\right)-0.5000\,\ell_{1}\,m_{0}\,{\dot{\phi }}^2\,\cos\left(\theta _{2}\right)-0.5000\,\ell_{1}\,m_{\mathrm{ACT}}\,{\dot{\phi }}^2\,\cos\left(\theta _{2}\right)- \\
    & 0.5000\,\ell_{1}\,m_{\mathrm{Obj}}\,{\dot{\phi }}^2\,\cos\left(\theta _{2}\right)-\ell_{1}\,m_{0}\,{\dot{\theta }_{1}}^2\,\cos\left(\theta _{2}\right)-\ell_{1}\,m_{\mathrm{ACT}}\,{\dot{\theta }_{1}}^2\,\cos\left(\theta _{2}\right)- \\
    & \ell_{1}\,m_{\mathrm{Obj}}\,{\dot{\theta }_{1}}^2\,\cos\left(\theta _{2}\right)-0.3927\,R^2\,r^2\,\varrho _{\mathrm{ACT}}\,\dot{\theta }_{1}\,\dot{\theta }_{2}-0.1963\,R^2\,r^2\,\varrho _{\mathrm{ACT}}\,{\dot{\theta }_{1}}^2\,\cos\left(2\,\theta _{2}+2\,\theta _{1}\right)- \\
    & 0.1963\,R^2\,r^2\,\varrho _{\mathrm{ACT}}\,{\dot{\theta }_{2}}^2\,\cos\left(2\,\theta _{2}+2\,\theta _{1}\right)+0.3927\,R^4\,\varrho _{\mathrm{ACT}}\,\dot{\theta }_{1}\,\dot{\theta }_{2}\,\cos\left(2\,\theta _{2}+2\,\theta _{1}\right)- \\
    & 0.3927\,R^2\,r^2\,\varrho _{\mathrm{ACT}}\,\dot{\theta }_{1}\,\dot{\theta }_{2}\,\cos\left(2\,\theta _{2}+2\,\theta _{1}\right) \\
    & -0.5000\,\ell_{0}\,m_{\mathrm{Obj}}\,{\dot{\phi }}^2\,\cos\left(-\vartheta+\theta _{2}+\theta _{1}\right)+\ell_{1}\,m_{0}\,\ddot{\theta }_{1}\,\sin\left(\theta _{2}\right)+\ell_{1}\,m_{\mathrm{ACT}}\,\ddot{\theta }_{1}\,\sin\left(\theta _{2}\right)+ \\
   & \ell_{1}\,m_{\mathrm{Obj}}\,\ddot{\theta }_{1}\,\sin\left(\theta _{2}\right)+0.1963\,R^4\,\varrho_{ACT}\,{\dot{\theta }_{1}}^2\,\cos\left(2\,\theta _{2}+2\,\theta _{1}\right)+ \\
   & 0.1963\,R^4\,\varrho_{ACT}\,{\dot{\theta }_{2}}^2\,\cos\left(2\,\theta _{2}+2\,\theta _{1}\right)-0.5000\,\ell_{1}\,m_{0}\,{\dot{\phi }}^2\,\cos\left(\theta _{2}\right)-0.5000\,\ell_{1}\,m_{\mathrm{ACT}}\,{\dot{\phi }}^2\,\cos\left(\theta _{2}\right)- \\
   & 0.5000\,\ell_{1}\,m_{\mathrm{Obj}}\,{\dot{\phi }}^2\,\cos\left(\theta _{2}\right)-\ell_{1}\,m_{0}\,{\dot{\theta }_{1}}^2\,\cos\left(\theta _{2}\right)-\ell_{1}\,m_{\mathrm{ACT}}\,{\dot{\theta }_{1}}^2\,\cos\left(\theta _{2}\right)- \\
   & \ell_{1}\,m_{\mathrm{Obj}}\,{\dot{\theta }_{1}}^2\,\cos\left(\theta _{2}\right)-0.3927\,R^2\,r^2\,\varrho_{ACT}\,\dot{\theta }_{1}\,\dot{\theta }_{2}-0.1963\,R^2\,r^2\,\varrho_{ACT}\,{\dot{\theta }_{1}}^2\,\cos\left(2\,\theta _{2}+2\,\theta _{1}\right)- \\
   & 0.1963\,R^2\,r^2\,\varrho_{ACT}\,{\dot{\theta }_{2}}^2\,\cos\left(2\,\theta _{2}+2\,\theta _{1}\right)+0.3927\,R^4\,\varrho_{ACT}\,\dot{\theta }_{1}\,\dot{\theta }_{2}\,\cos\left(2\,\theta _{2}+2\,\theta _{1}\right)- \\
   & 0.3927\,R^2\,r^2\,\varrho_{ACT}\,\dot{\theta }_{1}\,\dot{\theta }_{2}\,\cos\left(2\,\theta _{2}+2\,\theta _{1}\right) \\
\end{align*}

\subsubsection{Polar Coordinate \#1}
For the Polar Coordinate, $\theta_{1}: $
\begin{align*}
     Q_{\theta_1} = u_{\theta_1}
&= \frac{d}{dt}\!\left(\frac{\partial \mathcal{L}}{\partial \dot{\theta}_1}\right)
 - \frac{\partial \mathcal{L}}{\partial \theta_1}
 + \frac{\partial \mathcal{R}}{\partial \dot{\theta}_1}  \\= 
    & 0.2500\,I_{\mathrm{xx},1}\,\ddot{\theta }_{1}+0.2500\,I_{\mathrm{xx},2}\,\ddot{\theta }_{1}+0.2500\,I_{\mathrm{xx},2}\,\ddot{\theta }_{2}+0.2500\,I_{\mathrm{yy},1}\,\ddot{\theta }_{1}+0.2500\,I_{\mathrm{yy},2}\,\ddot{\theta }_{1}+ \\
    & 0.2500\,I_{\mathrm{yy},2}\,\ddot{\theta }_{2}+0.5000\,I_{\mathrm{zz},1}\,\ddot{\theta }_{1}+0.5000\,I_{\mathrm{zz},2}\,\ddot{\theta }_{1}+0.5000\,I_{\mathrm{zz},2}\,\ddot{\theta }_{2}+\frac{\partial \mathcal{R}}{\partial \dot{\theta}_{1}}+\kappa _{\theta ,1}\,\theta _{1}- \\
    & \kappa _{\theta ,1}\,\theta _{1,0}+{\delta _{r}}^2\,m_{0}\,\ddot{\theta }_{1}+{\delta _{r}}^2\,m_{0}\,\ddot{\theta }_{2}+{\delta _{r}}^2\,m_{\mathrm{Obj}}\,\ddot{\theta }_{1}+{\delta _{r}}^2\,m_{\mathrm{Obj}}\,\ddot{\theta }_{2}+{\ell_{1}}^2\,m_{0}\,\ddot{\theta }_{1}+ \\
    & {\ell_{2}}^2\,m_{0}\,\ddot{\theta }_{1}+{\ell_{1}}^2\,m_{2}\,\ddot{\theta }_{1}+{\ell_{2}}^2\,m_{0}\,\ddot{\theta }_{2}+{\ell_{1}}^2\,m_{\mathrm{ACT}}\,\ddot{\theta }_{1}+{\ell_{2}}^2\,m_{\mathrm{ACT}}\,\ddot{\theta }_{1}+{\ell_{2}}^2\,m_{\mathrm{ACT}}\,\ddot{\theta }_{2}+ \\
    & {\ell_{1}}^2\,m_{\mathrm{Obj}}\,\ddot{\theta }_{1}+{\ell_{2}}^2\,m_{\mathrm{Obj}}\,\ddot{\theta }_{1}+{\ell_{2}}^2\,m_{\mathrm{Obj}}\,\ddot{\theta }_{2}+{\bar{\ell}_{1}}^2\,m_{1}\,\ddot{\theta }_{1}+{\bar{\ell}_{2}}^2\,m_{2}\,\ddot{\theta }_{1}+{\bar{\ell}_{2}}^2\,m_{2}\,\ddot{\theta }_{2}+ \\
    & m_{0}\,r^2\,\ddot{\theta }_{1}+m_{0}\,r^2\,\ddot{\theta }_{2}+m_{\mathrm{ACT}}\,r^2\,\ddot{\theta }_{1}+m_{\mathrm{ACT}}\,r^2\,\ddot{\theta }_{2}+m_{\mathrm{Obj}}\,r^2\,\ddot{\theta }_{1}+m_{\mathrm{Obj}}\,r^2\,\ddot{\theta }_{2}+m_{0}\,{r^{\prime }}^2\,\ddot{\theta }_{1}+ \\
    & m_{0}\,{r^{\prime }}^2\,\ddot{\theta }_{2}+m_{\mathrm{Obj}}\,{r^{\prime }}^2\,\ddot{\theta }_{1}+m_{\mathrm{Obj}}\,{r^{\prime }}^2\,\ddot{\theta }_{2}+0.2500\,I_{\mathrm{xx},1}\,\ddot{\theta }_{1}\,\cos\left(2\,\phi \right)+ \\
    & 0.2500\,I_{\mathrm{xx},2}\,\ddot{\theta }_{1}\,\cos\left(2\,\phi \right)+0.2500\,I_{\mathrm{xx},2}\,\ddot{\theta }_{2}\,\cos\left(2\,\phi \right)-0.2500\,I_{\mathrm{yy},1}\,\ddot{\theta }_{1}\,\cos\left(2\,\phi \right)- \\
    & 0.2500\,I_{\mathrm{yy},2}\,\ddot{\theta }_{1}\,\cos\left(2\,\phi \right)-0.2500\,I_{\mathrm{yy},2}\,\ddot{\theta }_{2}\,\cos\left(2\,\phi \right)+0.2500\,I_{\mathrm{xx},1}\,\ddot{\theta }_{1}\,\cos\left(2\,\theta _{1}\right)+ \\
    & 0.2500\,I_{\mathrm{yy},1}\,\ddot{\theta }_{1}\,\cos\left(2\,\theta _{1}\right)-0.5000\,I_{\mathrm{zz},1}\,\ddot{\theta }_{1}\,\cos\left(2\,\theta _{1}\right)+0.1250\,I_{\mathrm{xx},2}\,\ddot{\theta }_{1}\,\cos\left(-2\,\phi +2\,\theta _{2}+2\,\theta _{1}\right)+ \\
    & 0.1250\,I_{\mathrm{xx},2}\,\ddot{\theta }_{1}\,\cos\left(2\,\phi +2\,\theta _{2}+2\,\theta _{1}\right)+0.1250\,I_{\mathrm{xx},2}\,\ddot{\theta }_{2}\,\cos\left(-2\,\phi +2\,\theta _{2}+2\,\theta _{1}\right)+ \\
    & 0.1250\,I_{\mathrm{xx},2}\,\ddot{\theta }_{2}\,\cos\left(2\,\phi +2\,\theta _{2}+2\,\theta _{1}\right)-0.1250\,I_{\mathrm{yy},2}\,\ddot{\theta }_{1}\,\cos\left(-2\,\phi +2\,\theta _{2}+2\,\theta _{1}\right)- \\
    & 0.1250\,I_{\mathrm{yy},2}\,\ddot{\theta }_{1}\,\cos\left(2\,\phi +2\,\theta _{2}+2\,\theta _{1}\right)-0.1250\,I_{\mathrm{yy},2}\,\ddot{\theta }_{2}\,\cos\left(-2\,\phi +2\,\theta _{2}+2\,\theta _{1}\right)- \\
    & 0.1250\,I_{\mathrm{yy},2}\,\ddot{\theta }_{2}\,\cos\left(2\,\phi +2\,\theta _{2}+2\,\theta _{1}\right)+0.1250\,I_{\mathrm{xx},1}\,\ddot{\theta }_{1}\,\cos\left(2\,\phi -2\,\theta _{1}\right)+ \\
    & 0.1250\,I_{\mathrm{xx},1}\,\ddot{\theta }_{1}\,\cos\left(2\,\phi +2\,\theta _{1}\right)-0.1250\,I_{\mathrm{yy},1}\,\ddot{\theta }_{1}\,\cos\left(2\,\phi -2\,\theta _{1}\right)-0.1250\,I_{\mathrm{yy},1}\,\ddot{\theta }_{1}\,\cos\left(2\,\phi +2\,\theta _{1}\right)+ \\
    & 0.2500\,I_{\mathrm{xx},2}\,\ddot{\theta }_{1}\,\cos\left(2\,\theta _{2}+2\,\theta _{1}\right)+0.2500\,I_{\mathrm{xx},2}\,\ddot{\theta }_{2}\,\cos\left(2\,\theta _{2}+2\,\theta _{1}\right)+ \\
    & 0.2500\,I_{\mathrm{yy},2}\,\ddot{\theta }_{1}\,\cos\left(2\,\theta _{2}+2\,\theta _{1}\right)+0.2500\,I_{\mathrm{yy},2}\,\ddot{\theta }_{2}\,\cos\left(2\,\theta _{2}+2\,\theta _{1}\right)- \\
    & 0.5000\,I_{\mathrm{zz},2}\,\ddot{\theta }_{1}\,\cos\left(2\,\theta _{2}+2\,\theta _{1}\right)-0.5000\,I_{\mathrm{zz},2}\,\ddot{\theta }_{2}\,\cos\left(2\,\theta _{2}+2\,\theta _{1}\right)-0.2500\,I_{\mathrm{xx},1}\,{\dot{\theta }_{1}}^2\,\sin\left(2\,\theta _{1}\right)- \\
    & 0.2500\,I_{\mathrm{yy},1}\,{\dot{\theta }_{1}}^2\,\sin\left(2\,\theta _{1}\right)+0.5000\,I_{\mathrm{zz},1}\,{\dot{\theta }_{1}}^2\,\sin\left(2\,\theta _{1}\right)-0.1250\,I_{\mathrm{xx},2}\,{\dot{\theta }_{1}}^2\,\sin\left(-2\,\phi +2\,\theta _{2}+2\,\theta _{1}\right)- \\
\end{align*}
\newpage
\begin{align*}
    & 2\,m_{0}\,\dot{r}\,r^{\prime }\,\dot{\theta }_{2}+2\,m_{\mathrm{Obj}}\,\dot{r}\,r^{\prime }\,\dot{\theta }_{1}+2\,m_{\mathrm{Obj}}\,\dot{r}\,r^{\prime }\,\dot{\theta }_{2}-\delta _{r}\,g\,m_{0}\,\sin\left(\theta _{2}+\theta _{1}\right)- \\
    & \delta _{r}\,g\,m_{\mathrm{Obj}}\,\sin\left(\theta _{2}+\theta _{1}\right)-g\,\ell_{2}\,m_{0}\,\sin\left(\theta _{2}+\theta _{1}\right)-g\,\ell_{2}\,m_{\mathrm{ACT}}\,\sin\left(\theta _{2}+\theta _{1}\right)-g\,\ell_{2}\,m_{\mathrm{Obj}}\,\sin\left(\theta _{2}+\theta _{1}\right)- \\
    & g\,\bar{\ell}_{2}\,m_{2}\,\sin\left(\theta _{2}+\theta _{1}\right)-g\,m_{0}\,r\,\sin\left(\theta _{2}+\theta _{1}\right)-g\,m_{\mathrm{ACT}}\,r\,\sin\left(\theta _{2}+\theta _{1}\right)-g\,m_{\mathrm{Obj}}\,r\,\sin\left(\theta _{2}+\theta _{1}\right)- \\
    & g\,m_{0}\,r^{\prime }\,\sin\left(\theta _{2}+\theta _{1}\right)-g\,m_{\mathrm{Obj}}\,r^{\prime }\,\sin\left(\theta _{2}+\theta _{1}\right)+1.1781\,R^4\,r\,\varrho_{\mathrm{ACT}}\,\ddot{\theta }_{1}+1.1781\,R^4\,r\,\varrho_{\mathrm{ACT}}\,\ddot{\theta }_{2}+ \\
    & 1.1781\,R^4\,\dot{r}\,\varrho_{\mathrm{ACT}}\,\dot{\theta }_{1}+1.1781\,R^4\,\dot{r}\,\varrho_{\mathrm{ACT}}\,\dot{\theta }_{2}-g\,\ell_{1}\,m_{0}\,\sin\left(\theta _{1}\right)-g\,\ell_{1}\,m_{2}\,\sin\left(\theta _{1}\right)-g\,\ell_{1}\,m_{\mathrm{ACT}}\,\sin\left(\theta _{1}\right)- \\
    & g\,\ell_{1}\,m_{\mathrm{Obj}}\,\sin\left(\theta _{1}\right)-g\,\bar{\ell}_{1}\,m_{1}\,\sin\left(\theta _{1}\right)+\ell_{1}\,m_{0}\,\ddot{r}\,\sin\left(\theta _{2}\right)+\ell_{1}\,m_{\mathrm{ACT}}\,\ddot{r}\,\sin\left(\theta _{2}\right)+ \\
    & \ell_{1}\,m_{\mathrm{Obj}}\,\ddot{r}\,\sin\left(\theta _{2}\right)-0.5000\,{\ell_{1}}^2\,m_{0}\,{\dot{\phi }}^2\,\sin\left(2\,\theta _{1}\right)-0.5000\,{\ell_{1}}^2\,m_{2}\,{\dot{\phi }}^2\,\sin\left(2\,\theta _{1}\right)- \\
    & 0.5000\,{\ell_{1}}^2\,m_{\mathrm{ACT}}\,{\dot{\phi }}^2\,\sin\left(2\,\theta _{1}\right)-0.5000\,{\ell_{1}}^2\,m_{\mathrm{Obj}}\,{\dot{\phi }}^2\,\sin\left(2\,\theta _{1}\right)-0.5000\,{\bar{\ell}_{1}}^2\,m_{1}\,{\dot{\phi }}^2\,\sin\left(2\,\theta _{1}\right)+ \\
    & 0.1309\,R^2\,r^3\,\varrho_{\mathrm{ACT}}\,\ddot{\theta }_{1}+0.1309\,R^2\,r^3\,\varrho_{\mathrm{ACT}}\,\ddot{\theta }_{2}-0.5000\,I_{\mathrm{xx},1}\,\dot{\phi }\,\dot{\theta }_{1}\,\sin\left(2\,\phi \right)- \\
    & 0.5000\,I_{\mathrm{xx},2}\,\dot{\phi }\,\dot{\theta }_{1}\,\sin\left(2\,\phi \right)-0.5000\,I_{\mathrm{xx},2}\,\dot{\phi }\,\dot{\theta }_{2}\,\sin\left(2\,\phi \right)+0.5000\,I_{\mathrm{yy},1}\,\dot{\phi }\,\dot{\theta }_{1}\,\sin\left(2\,\phi \right)+ \\
    & 0.5000\,I_{\mathrm{yy},2}\,\dot{\phi }\,\dot{\theta }_{1}\,\sin\left(2\,\phi \right)+0.5000\,I_{\mathrm{yy},2}\,\dot{\phi }\,\dot{\theta }_{2}\,\sin\left(2\,\phi \right)+ \\
    & 0.2500\,I_{\mathrm{xx},2}\,\dot{\phi }\,\dot{\theta }_{1}\,\sin\left(-2\,\phi +2\,\theta _{2}+2\,\theta _{1}\right)-0.2500\,I_{\mathrm{xx},2}\,\dot{\phi }\,\dot{\theta }_{1}\,\sin\left(2\,\phi +2\,\theta _{2}+2\,\theta _{1}\right)+ \\
    & 0.2500\,I_{\mathrm{xx},2}\,\dot{\phi }\,\dot{\theta }_{2}\,\sin\left(-2\,\phi +2\,\theta _{2}+2\,\theta _{1}\right)-0.2500\,I_{\mathrm{xx},2}\,\dot{\phi }\,\dot{\theta }_{2}\,\sin\left(2\,\phi +2\,\theta _{2}+2\,\theta _{1}\right)- \\
    & 0.2500\,I_{\mathrm{yy},2}\,\dot{\phi }\,\dot{\theta }_{1}\,\sin\left(-2\,\phi +2\,\theta _{2}+2\,\theta _{1}\right)+0.2500\,I_{\mathrm{yy},2}\,\dot{\phi }\,\dot{\theta }_{1}\,\sin\left(2\,\phi +2\,\theta _{2}+2\,\theta _{1}\right)- \\
    & 0.2500\,I_{\mathrm{yy},2}\,\dot{\phi }\,\dot{\theta }_{2}\,\sin\left(-2\,\phi +2\,\theta _{2}+2\,\theta _{1}\right)+0.2500\,I_{\mathrm{yy},2}\,\dot{\phi }\,\dot{\theta }_{2}\,\sin\left(2\,\phi +2\,\theta _{2}+2\,\theta _{1}\right)- \\
    & 0.5000\,{\delta _{r}}^2\,m_{0}\,{\dot{\phi }}^2\,\sin\left(2\,\theta _{2}+2\,\theta _{1}\right)-0.5000\,{\delta _{r}}^2\,m_{\mathrm{Obj}}\,{\dot{\phi }}^2\,\sin\left(2\,\theta _{2}+2\,\theta _{1}\right)- \\
    & 0.2500\,I_{\mathrm{xx},2}\,\dot{\theta }_{1}\,\dot{\theta }_{2}\,\sin\left(-2\,\phi +2\,\theta _{2}+2\,\theta _{1}\right)-0.2500\,I_{\mathrm{xx},2}\,\dot{\theta }_{1}\,\dot{\theta }_{2}\,\sin\left(2\,\phi +2\,\theta _{2}+2\,\theta _{1}\right)+ \\
    & 0.2500\,I_{\mathrm{yy},2}\,\dot{\theta }_{1}\,\dot{\theta }_{2}\,\sin\left(-2\,\phi +2\,\theta _{2}+2\,\theta _{1}\right)+0.2500\,I_{\mathrm{yy},2}\,\dot{\theta }_{1}\,\dot{\theta }_{2}\,\sin\left(2\,\phi +2\,\theta _{2}+2\,\theta _{1}\right)- \\
    & 0.5000\,{\ell_{2}}^2\,m_{0}\,{\dot{\phi }}^2\,\sin\left(2\,\theta _{2}+2\,\theta _{1}\right)-0.5000\,{\ell_{2}}^2\,m_{\mathrm{ACT}}\,{\dot{\phi }}^2\,\sin\left(2\,\theta _{2}+2\,\theta _{1}\right)- \\
    & 0.5000\,{\ell_{2}}^2\,m_{\mathrm{Obj}}\,{\dot{\phi }}^2\,\sin\left(2\,\theta _{2}+2\,\theta _{1}\right)-0.5000\,{\bar{\ell}_{2}}^2\,m_{2}\,{\dot{\phi }}^2\,\sin\left(2\,\theta _{2}+2\,\theta _{1}\right)- \\
\end{align*}
\newpage
\begin{align*}
    & 0.5000\,m_{0}\,{\dot{\phi }}^2\,r^2\,\sin\left(2\,\theta _{2}+2\,\theta _{1}\right)-0.5000\,m_{\mathrm{ACT}}\,{\dot{\phi }}^2\,r^2\,\sin\left(2\,\theta _{2}+2\,\theta _{1}\right)-0.5000\,m_{\mathrm{Obj}}\,{\dot{\phi }}^2\,r^2\,\sin\left(2\,\theta _{2}+2\,\theta _{1}\right)- \\
    & 0.5000\,m_{0}\,{\dot{\phi }}^2\,{r^{\prime }}^2\,\sin\left(2\,\theta _{2}+2\,\theta _{1}\right)-0.5000\,m_{\mathrm{Obj}}\,{\dot{\phi }}^2\,{r^{\prime }}^2\,\sin\left(2\,\theta _{2}+2\,\theta _{1}\right)+ \\
    & 0.5000\,\ell_{0}\,\ell_{2}\,m_{0}\,{\dot{\phi }}^2\,\sin\left(-\vartheta+\theta _{2}+\theta _{1}\right)+0.5000\,\ell_{0}\,\ell_{2}\,m_{\mathrm{ACT}}\,{\dot{\phi }}^2\,\sin\left(-\vartheta+\theta _{2}+\theta _{1}\right)+ \\
    & 0.5000\,\ell_{0}\,\ell_{2}\,m_{\mathrm{Obj}}\,{\dot{\phi }}^2\,\sin\left(-\vartheta+\theta _{2}+\theta _{1}\right)+0.5000\,\ell_{0}\,\bar{\ell}_{2}\,m_{2}\,{\dot{\phi }}^2\,\sin\left(-\vartheta+\theta _{2}+\theta _{1}\right)+ \\
    & 0.5000\,\ell_{0}\,m_{0}\,{\dot{\phi }}^2\,r\,\sin\left(-\vartheta+\theta _{2}+\theta _{1}\right)+0.5000\,\ell_{0}\,m_{\mathrm{ACT}}\,{\dot{\phi }}^2\,r\,\sin\left(-\vartheta+\theta _{2}+\theta _{1}\right)+ \\
    & 0.5000\,\ell_{0}\,m_{\mathrm{Obj}}\,{\dot{\phi }}^2\,r\,\sin\left(-\vartheta+\theta _{2}+\theta _{1}\right)+0.5000\,\ell_{0}\,m_{0}\,{\dot{\phi }}^2\,r^{\prime }\,\sin\left(-\vartheta+\theta _{2}+\theta _{1}\right)+ \\
    & 0.5000\,\ell_{0}\,m_{\mathrm{Obj}}\,{\dot{\phi }}^2\,r^{\prime }\,\sin\left(-\vartheta+\theta _{2}+\theta _{1}\right)-0.5000\,\ell_{0}\,\ell_{1}\,m_{0}\,{\dot{\phi }}^2\,\sin\left(\vartheta+\theta _{1}\right)- \\
    & 0.5000\,\ell_{0}\,\ell_{1}\,m_{2}\,{\dot{\phi }}^2\,\sin\left(\vartheta+\theta _{1}\right)-0.5000\,\ell_{0}\,\ell_{1}\,m_{\mathrm{ACT}}\,{\dot{\phi }}^2\,\sin\left(\vartheta+\theta _{1}\right)- \\
    & 0.5000\,\ell_{0}\,\ell_{1}\,m_{\mathrm{Obj}}\,{\dot{\phi }}^2\,\sin\left(\vartheta+\theta _{1}\right)-0.5000\,\ell_{0}\,\bar{\ell}_{1}\,m_{1}\,{\dot{\phi }}^2\,\sin\left(\vartheta+\theta _{1}\right)+ \\
    & 0.1309\,R^2\,r^3\,\varrho_{\mathrm{ACT}}\,\ddot{\theta }_{1}\,\cos\left(2\,\theta _{2}+2\,\theta _{1}\right)+0.1309\,R^2\,r^3\,\varrho_{\mathrm{ACT}}\,\ddot{\theta }_{2}\,\cos\left(2\,\theta _{2}+2\,\theta _{1}\right)+ \\
    & 0.3927\,R^2\,r^2\,\dot{r}\,\varrho_{\mathrm{ACT}}\,\dot{\theta }_{1}+0.3927\,R^2\,r^2\,\dot{r}\,\varrho_{\mathrm{ACT}}\,\dot{\theta }_{2}+0.3927\,R^4\,r\,\varrho_{\mathrm{ACT}}\,{\dot{\theta }_{1}}^2\,\sin\left(2\,\theta _{2}+2\,\theta _{1}\right)+ \\
    & 0.3927\,R^4\,r\,\varrho_{\mathrm{ACT}}\,{\dot{\theta }_{2}}^2\,\sin\left(2\,\theta _{2}+2\,\theta _{1}\right)-\delta _{r}\,\ell_{1}\,m_{0}\,{\dot{\theta }_{2}}^2\,\sin\left(\theta _{2}\right)- \\
    & \delta _{r}\,\ell_{1}\,m_{\mathrm{Obj}}\,{\dot{\theta }_{2}}^2\,\sin\left(\theta _{2}\right)-\ell_{1}\,\ell_{2}\,m_{0}\,{\dot{\theta }_{2}}^2\,\sin\left(\theta _{2}\right)-\ell_{1}\,\ell_{2}\,m_{\mathrm{ACT}}\,{\dot{\theta }_{2}}^2\,\sin\left(\theta _{2}\right)- \\
    & \ell_{1}\,\ell_{2}\,m_{\mathrm{Obj}}\,{\dot{\theta }_{2}}^2\,\sin\left(\theta _{2}\right)-\ell_{1}\,\bar{\ell}_{2}\,m_{2}\,{\dot{\theta }_{2}}^2\,\sin\left(\theta _{2}\right)-\ell_{1}\,m_{0}\,r\,{\dot{\theta }_{2}}^2\,\sin\left(\theta _{2}\right)- \\
    & \ell_{1}\,m_{\mathrm{ACT}}\,r\,{\dot{\theta }_{2}}^2\,\sin\left(\theta _{2}\right)-\ell_{1}\,m_{\mathrm{Obj}}\,r\,{\dot{\theta }_{2}}^2\,\sin\left(\theta _{2}\right)-\ell_{1}\,m_{0}\,r^{\prime }\,{\dot{\theta }_{2}}^2\,\sin\left(\theta _{2}\right)- \\
    & \ell_{1}\,m_{\mathrm{Obj}}\,r^{\prime }\,{\dot{\theta }_{2}}^2\,\sin\left(\theta _{2}\right)-\delta _{r}\,\ell_{1}\,m_{0}\,{\dot{\phi }}^2\,\sin\left(\theta _{2}+2\,\theta _{1}\right)- \\
    & \delta _{r}\,\ell_{1}\,m_{\mathrm{Obj}}\,{\dot{\phi }}^2\,\sin\left(\theta _{2}+2\,\theta _{1}\right)-\ell_{1}\,\ell_{2}\,m_{0}\,{\dot{\phi }}^2\,\sin\left(\theta _{2}+2\,\theta _{1}\right)-\ell_{1}\,\ell_{2}\,m_{\mathrm{ACT}}\,{\dot{\phi }}^2\,\sin\left(\theta _{2}+2\,\theta _{1}\right)- \\
    & \ell_{1}\,\ell_{2}\,m_{\mathrm{Obj}}\,{\dot{\phi }}^2\,\sin\left(\theta _{2}+2\,\theta _{1}\right)-\ell_{1}\,\bar{\ell}_{2}\,m_{2}\,{\dot{\phi }}^2\,\sin\left(\theta _{2}+2\,\theta _{1}\right)+ \\
    & 0.5000\,\ell_{0}\,\ell_{1}\,m_{0}\,{\dot{\phi }}^2\,\sin\left(-\vartheta+\theta _{1}\right)+0.5000\,\ell_{0}\,\ell_{1}\,m_{2}\,{\dot{\phi }}^2\,\sin\left(-\vartheta+\theta _{1}\right)+ \\
    & 0.5000\,\ell_{0}\,\ell_{1}\,m_{\mathrm{ACT}}\,{\dot{\phi }}^2\,\sin\left(-\vartheta+\theta _{1}\right)+0.5000\,\ell_{0}\,\ell_{1}\,m_{\mathrm{Obj}}\,{\dot{\phi }}^2\,\sin\left(-\vartheta+\theta _{1}\right)+ \\
\end{align*}
\newpage
\begin{align*}
    & 0.5000\,\ell_{0}\,\bar{\ell}_{1}\,m_{1}\,{\dot{\phi }}^2\,\sin\left(-\vartheta+\theta _{1}\right)-\ell_{1}\,m_{0}\,{\dot{\phi }}^2\,r\,\sin\left(\theta _{2}+2\,\theta _{1}\right)-\ell_{1}\,m_{\mathrm{ACT}}\,{\dot{\phi }}^2\,r\,\sin\left(\theta _{2}+2\,\theta _{1}\right)- \\
    & \ell_{1}\,m_{\mathrm{Obj}}\,{\dot{\phi }}^2\,r\,\sin\left(\theta _{2}+2\,\theta _{1}\right)-\ell_{1}\,m_{0}\,{\dot{\phi }}^2\,r^{\prime }\,\sin\left(\theta _{2}+2\,\theta _{1}\right)- \\
    & \ell_{1}\,m_{\mathrm{Obj}}\,{\dot{\phi }}^2\,r^{\prime }\,\sin\left(\theta _{2}+2\,\theta _{1}\right)-0.5000\,\delta _{r}\,\ell_{0}\,m_{0}\,{\dot{\phi }}^2\,\sin\left(\vartheta+\theta _{2}+\theta _{1}\right)- \\
    & 0.5000\,\delta _{r}\,\ell_{0}\,m_{\mathrm{Obj}}\,{\dot{\phi }}^2\,\sin\left(\vartheta+\theta _{2}+\theta _{1}\right)-0.1309\,R^2\,r^3\,\varrho_{\mathrm{ACT}}\,{\dot{\theta }_{1}}^2\,\sin\left(2\,\theta _{2}+2\,\theta _{1}\right)- \\
    & 0.1309\,R^2\,r^3\,\varrho_{\mathrm{ACT}}\,{\dot{\theta }_{2}}^2\,\sin\left(2\,\theta _{2}+2\,\theta _{1}\right)-0.5000\,\ell_{0}\,\ell_{2}\,m_{0}\,{\dot{\phi }}^2\,\sin\left(\vartheta+\theta _{2}+\theta _{1}\right)- \\
    & 0.5000\,\ell_{0}\,\ell_{2}\,m_{\mathrm{ACT}}\,{\dot{\phi }}^2\,\sin\left(\vartheta+\theta _{2}+\theta _{1}\right)-0.5000\,\ell_{0}\,\ell_{2}\,m_{\mathrm{Obj}}\,{\dot{\phi }}^2\,\sin\left(\vartheta+\theta _{2}+\theta _{1}\right)- \\
    & 0.5000\,\ell_{0}\,\bar{\ell}_{2}\,m_{2}\,{\dot{\phi }}^2\,\sin\left(\vartheta+\theta _{2}+\theta _{1}\right)-0.5000\,\ell_{0}\,m_{0}\,{\dot{\phi }}^2\,r\,\sin\left(\vartheta+\theta _{2}+\theta _{1}\right)- \\
    & 0.5000\,\ell_{0}\,m_{\mathrm{ACT}}\,{\dot{\phi }}^2\,r\,\sin\left(\vartheta+\theta _{2}+\theta _{1}\right)-0.5000\,\ell_{0}\,m_{\mathrm{Obj}}\,{\dot{\phi }}^2\,r\,\sin\left(\vartheta+\theta _{2}+\theta _{1}\right)- \\
    & 0.5000\,\ell_{0}\,m_{0}\,{\dot{\phi }}^2\,r^{\prime }\,\sin\left(\vartheta+\theta _{2}+\theta _{1}\right)-0.5000\,\ell_{0}\,m_{\mathrm{Obj}}\,{\dot{\phi }}^2\,r^{\prime }\,\sin\left(\vartheta+\theta _{2}+\theta _{1}\right)- \\
    & 0.3927\,R^4\,r\,\varrho_{\mathrm{ACT}}\,\ddot{\theta }_{1}\,\cos\left(2\,\theta _{2}+2\,\theta _{1}\right)-0.3927\,R^4\,r\,\varrho_{\mathrm{ACT}}\,\ddot{\theta }_{2}\,\cos\left(2\,\theta _{2}+2\,\theta _{1}\right)- \\
    & 0.3927\,R^4\,\dot{r}\,\varrho_{\mathrm{ACT}}\,\dot{\theta }_{1}\,\cos\left(2\,\theta _{2}+2\,\theta _{1}\right)-0.3927\,R^4\,\dot{r}\,\varrho_{\mathrm{ACT}}\,\dot{\theta }_{2}\,\cos\left(2\,\theta _{2}+2\,\theta _{1}\right)- \\
    & \delta _{r}\,\ell_{2}\,m_{0}\,{\dot{\phi }}^2\,\sin\left(2\,\theta _{2}+2\,\theta _{1}\right)-\delta _{r}\,\ell_{2}\,m_{\mathrm{Obj}}\,{\dot{\phi }}^2\,\sin\left(2\,\theta _{2}+2\,\theta _{1}\right)+2\,\delta _{r}\,\ell_{1}\,m_{0}\,\ddot{\theta }_{1}\,\cos\left(\theta _{2}\right)+ \\
    & \delta _{r}\,\ell_{1}\,m_{0}\,\ddot{\theta }_{2}\,\cos\left(\theta _{2}\right)+2\,\delta _{r}\,\ell_{1}\,m_{\mathrm{Obj}}\,\ddot{\theta }_{1}\,\cos\left(\theta _{2}\right)+\delta _{r}\,\ell_{1}\,m_{\mathrm{Obj}}\,\ddot{\theta }_{2}\,\cos\left(\theta _{2}\right)- \\
    & \delta _{r}\,m_{0}\,{\dot{\phi }}^2\,r\,\sin\left(2\,\theta _{2}+2\,\theta _{1}\right)-\delta _{r}\,m_{\mathrm{Obj}}\,{\dot{\phi }}^2\,r\,\sin\left(2\,\theta _{2}+2\,\theta _{1}\right)- \\
    & \delta _{r}\,m_{0}\,{\dot{\phi }}^2\,r^{\prime }\,\sin\left(2\,\theta _{2}+2\,\theta _{1}\right)-\delta _{r}\,m_{\mathrm{Obj}}\,{\dot{\phi }}^2\,r^{\prime }\,\sin\left(2\,\theta _{2}+2\,\theta _{1}\right)+ \\
    & 2\,\ell_{1}\,\ell_{2}\,m_{0}\,\ddot{\theta }_{1}\,\cos\left(\theta _{2}\right)+\ell_{1}\,\ell_{2}\,m_{0}\,\ddot{\theta }_{2}\,\cos\left(\theta _{2}\right)+2\,\ell_{1}\,\ell_{2}\,m_{\mathrm{ACT}}\,\ddot{\theta }_{1}\,\cos\left(\theta _{2}\right)+ \\
    & \ell_{1}\,\ell_{2}\,m_{\mathrm{ACT}}\,\ddot{\theta }_{2}\,\cos\left(\theta _{2}\right)+2\,\ell_{1}\,\ell_{2}\,m_{\mathrm{Obj}}\,\ddot{\theta }_{1}\,\cos\left(\theta _{2}\right)+\ell_{1}\,\ell_{2}\,m_{\mathrm{Obj}}\,\ddot{\theta }_{2}\,\cos\left(\theta _{2}\right)+ \\
    & 2\,\ell_{1}\,\bar{\ell}_{2}\,m_{2}\,\ddot{\theta }_{1}\,\cos\left(\theta _{2}\right)+\ell_{1}\,\bar{\ell}_{2}\,m_{2}\,\ddot{\theta }_{2}\,\cos\left(\theta _{2}\right)-\ell_{2}\,m_{0}\,{\dot{\phi }}^2\,r\,\sin\left(2\,\theta _{2}+2\,\theta _{1}\right)- \\
    & \ell_{2}\,m_{\mathrm{ACT}}\,{\dot{\phi }}^2\,r\,\sin\left(2\,\theta _{2}+2\,\theta _{1}\right)-\ell_{2}\,m_{\mathrm{Obj}}\,{\dot{\phi }}^2\,r\,\sin\left(2\,\theta _{2}+2\,\theta _{1}\right)-\ell_{2}\,m_{0}\,{\dot{\phi }}^2\,r^{\prime }\,\sin\left(2\,\theta _{2}+2\,\theta _{1}\right) \\
    & -\ell_{2}\,m_{\mathrm{Obj}}\,{\dot{\phi }}^2\,r^{\prime }\,\sin\left(2\,\theta _{2}+2\,\theta _{1}\right)+2\,\ell_{1}\,m_{0}\,r\,\ddot{\theta }_{1}\,\cos\left(\theta _{2}\right)+\ell_{1}\,m_{0}\,r\,\ddot{\theta }_{2}\,\cos\left(\theta _{2}\right)+ \\
\end{align*}
\newpage
\begin{align*}
    & 2\,\ell_{1}\,m_{\mathrm{ACT}}\,r\,\ddot{\theta }_{1}\,\cos\left(\theta _{2}\right)+\ell_{1}\,m_{\mathrm{ACT}}\,r\,\ddot{\theta }_{2}\,\cos\left(\theta _{2}\right)+2\,\ell_{1}\,m_{\mathrm{Obj}}\,r\,\ddot{\theta }_{1}\,\cos\left(\theta _{2}\right)+ \\
    & \ell_{1}\,m_{\mathrm{Obj}}\,r\,\ddot{\theta }_{2}\,\cos\left(\theta _{2}\right)+2\,\ell_{1}\,m_{0}\,\dot{r}\,\dot{\theta }_{1}\,\cos\left(\theta _{2}\right)+2\,\ell_{1}\,m_{0}\,\dot{r}\,\dot{\theta }_{2}\,\cos\left(\theta _{2}\right)+ \\
    & 2\,\ell_{1}\,m_{0}\,r^{\prime }\,\ddot{\theta }_{1}\,\cos\left(\theta _{2}\right)+\ell_{1}\,m_{0}\,r^{\prime }\,\ddot{\theta }_{2}\,\cos\left(\theta _{2}\right)+2\,\ell_{1}\,m_{\mathrm{ACT}}\,\dot{r}\,\dot{\theta }_{1}\,\cos\left(\theta _{2}\right)+ \\
    & 2\,\ell_{1}\,m_{\mathrm{ACT}}\,\dot{r}\,\dot{\theta }_{2}\,\cos\left(\theta _{2}\right)+2\,\ell_{1}\,m_{\mathrm{Obj}}\,\dot{r}\,\dot{\theta }_{1}\,\cos\left(\theta _{2}\right)+2\,\ell_{1}\,m_{\mathrm{Obj}}\,\dot{r}\,\dot{\theta }_{2}\,\cos\left(\theta _{2}\right)+ \\
    & 2\,\ell_{1}\,m_{\mathrm{Obj}}\,r^{\prime }\,\ddot{\theta }_{1}\,\cos\left(\theta _{2}\right)+\ell_{1}\,m_{\mathrm{Obj}}\,r^{\prime }\,\ddot{\theta }_{2}\,\cos\left(\theta _{2}\right)-m_{0}\,{\dot{\phi }}^2\,r\,r^{\prime }\,\sin\left(2\,\theta _{2}+2\,\theta _{1}\right)- \\
    & m_{\mathrm{Obj}}\,{\dot{\phi }}^2\,r\,r^{\prime }\,\sin\left(2\,\theta _{2}+2\,\theta _{1}\right)+0.5000\,\delta _{r}\,\ell_{0}\,m_{0}\,{\dot{\phi }}^2\,\sin\left(-\vartheta+\theta _{2}+\theta _{1}\right)+ \\
    & 0.5000\,\delta _{r}\,\ell_{0}\,m_{\mathrm{Obj}}\,{\dot{\phi }}^2\,\sin\left(-\vartheta+\theta _{2}+\theta _{1}\right)+0.7854\,R^4\,r\,\varrho_{\mathrm{ACT}}\,\dot{\theta }_{1}\,\dot{\theta }_{2}\,\sin\left(2\,\theta _{2}+2\,\theta _{1}\right)- \\
    & 2\,\delta _{r}\,\ell_{1}\,m_{0}\,\dot{\theta }_{1}\,\dot{\theta }_{2}\,\sin\left(\theta _{2}\right)-2\,\delta _{r}\,\ell_{1}\,m_{\mathrm{Obj}}\,\dot{\theta }_{1}\,\dot{\theta }_{2}\,\sin\left(\theta _{2}\right)- \\
    & 2\,\ell_{1}\,\ell_{2}\,m_{0}\,\dot{\theta }_{1}\,\dot{\theta }_{2}\,\sin\left(\theta _{2}\right)-2\,\ell_{1}\,\ell_{2}\,m_{\mathrm{ACT}}\,\dot{\theta }_{1}\,\dot{\theta }_{2}\,\sin\left(\theta _{2}\right)- \\
    & 2\,\ell_{1}\,\ell_{2}\,m_{\mathrm{Obj}}\,\dot{\theta }_{1}\,\dot{\theta }_{2}\,\sin\left(\theta _{2}\right)-2\,\ell_{1}\,\bar{\ell}_{2}\,m_{2}\,\dot{\theta }_{1}\,\dot{\theta }_{2}\,\sin\left(\theta _{2}\right)- \\
    & 2\,\ell_{1}\,m_{0}\,r\,\dot{\theta }_{1}\,\dot{\theta }_{2}\,\sin\left(\theta _{2}\right)-2\,\ell_{1}\,m_{\mathrm{ACT}}\,r\,\dot{\theta }_{1}\,\dot{\theta }_{2}\,\sin\left(\theta _{2}\right)- \\
    & 2\,\ell_{1}\,m_{\mathrm{Obj}}\,r\,\dot{\theta }_{1}\,\dot{\theta }_{2}\,\sin\left(\theta _{2}\right)-2\,\ell_{1}\,m_{0}\,r^{\prime }\,\dot{\theta }_{1}\,\dot{\theta }_{2}\,\sin\left(\theta _{2}\right)- \\
    & 2\,\ell_{1}\,m_{\mathrm{Obj}}\,r^{\prime }\,\dot{\theta }_{1}\,\dot{\theta }_{2}\,\sin\left(\theta _{2}\right)+0.3927\,R^2\,r^2\,\dot{r}\,\varrho_{\mathrm{ACT}}\,\dot{\theta }_{1}\,\cos\left(2\,\theta _{2}+2\,\theta _{1}\right)+ \\
    & 0.3927\,R^2\,r^2\,\dot{r}\,\varrho_{\mathrm{ACT}}\,\dot{\theta }_{2}\,\cos\left(2\,\theta _{2}+2\,\theta _{1}\right)-0.2618\,R^2\,r^3\,\varrho_{\mathrm{ACT}}\,\dot{\theta }_{1}\,\dot{\theta }_{2}\,\sin\left(2\,\theta _{2}+2\,\theta _{1}\right) \\
\end{align*}
\newpage
\subsubsection*{Polar Coordinate \#2}
\begin{align*}
    Q_{\theta_2} = u_{\theta_2}
&= \frac{d}{dt}\!\left(\frac{\partial \mathcal{L}}{\partial \dot{\theta}_2}\right)
 - \frac{\partial \mathcal{L}}{\partial \theta_2}
 + \frac{\partial \mathcal{R}}{\partial \dot{\theta}_2}  \\
& 0.2500\,I_{\mathrm{xx},2}\,\ddot{\theta }_{1}+0.2500\,I_{\mathrm{xx},2}\,\ddot{\theta }_{2}+0.2500\,I_{\mathrm{yy},2}\,\ddot{\theta }_{1}+0.2500\,I_{\mathrm{yy},2}\,\ddot{\theta }_{2}+0.5000\,I_{\mathrm{zz},2}\,\ddot{\theta }_{1}+ \\
    & 0.5000\,I_{\mathrm{zz},2}\,\ddot{\theta }_{2}+\frac{\partial \mathcal{R}}{\partial \dot{\theta}_{2}}+\kappa _{\theta ,2}\,\theta _{2}-\kappa _{\theta ,2}\,\theta _{2,0}+{\delta _{r}}^2\,m_{0}\,\ddot{\theta }_{1}+{\delta _{r}}^2\,m_{0}\,\ddot{\theta }_{2}+ \\
    & {\delta _{r}}^2\,m_{\mathrm{Obj}}\,\ddot{\theta }_{1}+{\delta _{r}}^2\,m_{\mathrm{Obj}}\,\ddot{\theta }_{2}+{\ell_{2}}^2\,m_{0}\,\ddot{\theta }_{1}+{\ell_{2}}^2\,m_{0}\,\ddot{\theta }_{2}+{\ell_{2}}^2\,m_{\mathrm{ACT}}\,\ddot{\theta }_{1}+ \\
    & {\ell_{2}}^2\,m_{\mathrm{ACT}}\,\ddot{\theta }_{2}+{\ell_{2}}^2\,m_{\mathrm{Obj}}\,\ddot{\theta }_{1}+{\ell_{2}}^2\,m_{\mathrm{Obj}}\,\ddot{\theta }_{2}+{\bar{\ell}_{2}}^2\,m_{2}\,\ddot{\theta }_{1}+{\bar{\ell}_{2}}^2\,m_{2}\,\ddot{\theta }_{2}+m_{0}\,r^2\,\ddot{\theta }_{1}+ \\
    & m_{0}\,r^2\,\ddot{\theta }_{2}+m_{\mathrm{ACT}}\,r^2\,\ddot{\theta }_{1}+m_{\mathrm{ACT}}\,r^2\,\ddot{\theta }_{2}+m_{\mathrm{Obj}}\,r^2\,\ddot{\theta }_{1}+m_{\mathrm{Obj}}\,r^2\,\ddot{\theta }_{2}+m_{0}\,{r^{\prime }}^2\,\ddot{\theta }_{1}+ \\
    & m_{0}\,{r^{\prime }}^2\,\ddot{\theta }_{2}+m_{\mathrm{Obj}}\,{r^{\prime }}^2\,\ddot{\theta }_{1}+m_{\mathrm{Obj}}\,{r^{\prime }}^2\,\ddot{\theta }_{2}+0.2500\,I_{\mathrm{xx},2}\,\ddot{\theta }_{1}\,\cos\left(2\,\phi \right)+ \\
    & 0.2500\,I_{\mathrm{xx},2}\,\ddot{\theta }_{2}\,\cos\left(2\,\phi \right)-0.2500\,I_{\mathrm{yy},2}\,\ddot{\theta }_{1}\,\cos\left(2\,\phi \right)-0.2500\,I_{\mathrm{yy},2}\,\ddot{\theta }_{2}\,\cos\left(2\,\phi \right)+ \\
    & 0.1250\,I_{\mathrm{xx},2}\,\ddot{\theta }_{1}\,\cos\left(-2\,\phi +2\,\theta _{2}+2\,\theta _{1}\right)+0.1250\,I_{\mathrm{xx},2}\,\ddot{\theta }_{1}\,\cos\left(2\,\phi +2\,\theta _{2}+2\,\theta _{1}\right)+ \\
    & 0.1250\,I_{\mathrm{xx},2}\,\ddot{\theta }_{2}\,\cos\left(-2\,\phi +2\,\theta _{2}+2\,\theta _{1}\right)+0.1250\,I_{\mathrm{xx},2}\,\ddot{\theta }_{2}\,\cos\left(2\,\phi +2\,\theta _{2}+2\,\theta _{1}\right)- \\
    & 0.1250\,I_{\mathrm{yy},2}\,\ddot{\theta }_{1}\,\cos\left(-2\,\phi +2\,\theta _{2}+2\,\theta _{1}\right)-0.1250\,I_{\mathrm{yy},2}\,\ddot{\theta }_{1}\,\cos\left(2\,\phi +2\,\theta _{2}+2\,\theta _{1}\right)- \\
    & 0.1250\,I_{\mathrm{yy},2}\,\ddot{\theta }_{2}\,\cos\left(-2\,\phi +2\,\theta _{2}+2\,\theta _{1}\right)-0.1250\,I_{\mathrm{yy},2}\,\ddot{\theta }_{2}\,\cos\left(2\,\phi +2\,\theta _{2}+2\,\theta _{1}\right)+ \\
    & 0.2500\,I_{\mathrm{xx},2}\,\ddot{\theta }_{1}\,\cos\left(2\,\theta _{2}+2\,\theta _{1}\right)+0.2500\,I_{\mathrm{xx},2}\,\ddot{\theta }_{2}\,\cos\left(2\,\theta _{2}+2\,\theta _{1}\right)+ \\
    & 0.2500\,I_{\mathrm{yy},2}\,\ddot{\theta }_{1}\,\cos\left(2\,\theta _{2}+2\,\theta _{1}\right)+0.2500\,I_{\mathrm{yy},2}\,\ddot{\theta }_{2}\,\cos\left(2\,\theta _{2}+2\,\theta _{1}\right)- \\
    & 0.5000\,I_{\mathrm{zz},2}\,\ddot{\theta }_{1}\,\cos\left(2\,\theta _{2}+2\,\theta _{1}\right)-0.5000\,I_{\mathrm{zz},2}\,\ddot{\theta }_{2}\,\cos\left(2\,\theta _{2}+2\,\theta _{1}\right)- \\
    & 0.1250\,I_{\mathrm{xx},2}\,{\dot{\theta }_{1}}^2\,\sin\left(-2\,\phi +2\,\theta _{2}+2\,\theta _{1}\right)-0.1250\,I_{\mathrm{xx},2}\,{\dot{\theta }_{1}}^2\,\sin\left(2\,\phi +2\,\theta _{2}+2\,\theta _{1}\right)- \\
    & 0.1250\,I_{\mathrm{xx},2}\,{\dot{\theta }_{2}}^2\,\sin\left(-2\,\phi +2\,\theta _{2}+2\,\theta _{1}\right)-0.1250\,I_{\mathrm{xx},2}\,{\dot{\theta }_{2}}^2\,\sin\left(2\,\phi +2\,\theta _{2}+2\,\theta _{1}\right)+ \\
    & 0.1250\,I_{\mathrm{yy},2}\,{\dot{\theta }_{1}}^2\,\sin\left(-2\,\phi +2\,\theta _{2}+2\,\theta _{1}\right)+0.1250\,I_{\mathrm{yy},2}\,{\dot{\theta }_{1}}^2\,\sin\left(2\,\phi +2\,\theta _{2}+2\,\theta _{1}\right)+ \\
    & 0.1250\,I_{\mathrm{yy},2}\,{\dot{\theta }_{2}}^2\,\sin\left(-2\,\phi +2\,\theta _{2}+2\,\theta _{1}\right)+0.1250\,I_{\mathrm{yy},2}\,{\dot{\theta }_{2}}^2\,\sin\left(2\,\phi +2\,\theta _{2}+2\,\theta _{1}\right)- \\
    & 0.2500\,I_{\mathrm{xx},2}\,{\dot{\theta }_{1}}^2\,\sin\left(2\,\theta _{2}+2\,\theta _{1}\right)-0.2500\,I_{\mathrm{xx},2}\,{\dot{\theta }_{2}}^2\,\sin\left(2\,\theta _{2}+2\,\theta _{1}\right)- \\
\end{align*}
\newpage
\begin{align*}
    & 0.2500\,I_{\mathrm{yy},2}\,{\dot{\theta }_{1}}^2\,\sin\left(2\,\theta _{2}+2\,\theta _{1}\right)-0.2500\,I_{\mathrm{yy},2}\,{\dot{\theta }_{2}}^2\,\sin\left(2\,\theta _{2}+2\,\theta _{1}\right)+ \\
    & 0.5000\,I_{\mathrm{zz},2}\,{\dot{\theta }_{1}}^2\,\sin\left(2\,\theta _{2}+2\,\theta _{1}\right)+0.5000\,I_{\mathrm{zz},2}\,{\dot{\theta }_{2}}^2\,\sin\left(2\,\theta _{2}+2\,\theta _{1}\right)- \\
    & 0.5000\,I_{\mathrm{xx},2}\,\dot{\theta }_{1}\,\dot{\theta }_{2}\,\sin\left(2\,\theta _{2}+2\,\theta _{1}\right)-0.5000\,I_{\mathrm{yy},2}\,\dot{\theta }_{1}\,\dot{\theta }_{2}\,\sin\left(2\,\theta _{2}+2\,\theta _{1}\right)+ \\
    & I_{\mathrm{zz},2}\,\dot{\theta }_{1}\,\dot{\theta }_{2}\,\sin\left(2\,\theta _{2}+2\,\theta _{1}\right)+2\,\delta _{r}\,\ell_{2}\,m_{0}\,\ddot{\theta }_{1}+2\,\delta _{r}\,\ell_{2}\,m_{0}\,\ddot{\theta }_{2}+2\,\delta _{r}\,\ell_{2}\,m_{\mathrm{Obj}}\,\ddot{\theta }_{1}+ \\
    & 2\,\delta _{r}\,\ell_{2}\,m_{\mathrm{Obj}}\,\ddot{\theta }_{2}+2\,\delta _{r}\,m_{0}\,r\,\ddot{\theta }_{1}+2\,\delta _{r}\,m_{0}\,r\,\ddot{\theta }_{2}+2\,\delta _{r}\,m_{\mathrm{Obj}}\,r\,\ddot{\theta }_{1}+2\,\delta _{r}\,m_{\mathrm{Obj}}\,r\,\ddot{\theta }_{2}+ \\
    & 2\,\delta _{r}\,m_{0}\,\dot{r}\,\dot{\theta }_{1}+2\,\delta _{r}\,m_{0}\,\dot{r}\,\dot{\theta }_{2}+2\,\delta _{r}\,m_{0}\,r^{\prime }\,\ddot{\theta }_{1}+2\,\delta _{r}\,m_{0}\,r^{\prime }\,\ddot{\theta }_{2}+2\,\delta _{r}\,m_{\mathrm{Obj}}\,\dot{r}\,\dot{\theta }_{1}+ \\
    & 2\,\delta _{r}\,m_{\mathrm{Obj}}\,\dot{r}\,\dot{\theta }_{2}+2\,\delta _{r}\,m_{\mathrm{Obj}}\,r^{\prime }\,\ddot{\theta }_{1}+2\,\delta _{r}\,m_{\mathrm{Obj}}\,r^{\prime }\,\ddot{\theta }_{2}+2\,\ell_{2}\,m_{0}\,r\,\ddot{\theta }_{1}+2\,\ell_{2}\,m_{0}\,r\,\ddot{\theta }_{2}+ \\
    & 2\,\ell_{2}\,m_{\mathrm{ACT}}\,r\,\ddot{\theta }_{1}+2\,\ell_{2}\,m_{\mathrm{ACT}}\,r\,\ddot{\theta }_{2}+2\,\ell_{2}\,m_{\mathrm{Obj}}\,r\,\ddot{\theta }_{1}+2\,\ell_{2}\,m_{\mathrm{Obj}}\,r\,\ddot{\theta }_{2}+2\,\ell_{2}\,m_{0}\,\dot{r}\,\dot{\theta }_{1}+ \\
    & 2\,\ell_{2}\,m_{0}\,\dot{r}\,\dot{\theta }_{2}+2\,\ell_{2}\,m_{0}\,r^{\prime }\,\ddot{\theta }_{1}+2\,\ell_{2}\,m_{0}\,r^{\prime }\,\ddot{\theta }_{2}+2\,\ell_{2}\,m_{\mathrm{ACT}}\,\dot{r}\,\dot{\theta }_{1}+2\,\ell_{2}\,m_{\mathrm{ACT}}\,\dot{r}\,\dot{\theta }_{2}+ \\
    & 2\,\ell_{2}\,m_{\mathrm{Obj}}\,\dot{r}\,\dot{\theta }_{1}+2\,\ell_{2}\,m_{\mathrm{Obj}}\,\dot{r}\,\dot{\theta }_{2}+2\,\ell_{2}\,m_{\mathrm{Obj}}\,r^{\prime }\,\ddot{\theta }_{1}+2\,\ell_{2}\,m_{\mathrm{Obj}}\,r^{\prime }\,\ddot{\theta }_{2}+2\,m_{0}\,r\,\dot{r}\,\dot{\theta }_{1}+ \\
    & 2\,m_{0}\,r\,\dot{r}\,\dot{\theta }_{2}+2\,m_{0}\,r\,r^{\prime }\,\ddot{\theta }_{1}+2\,m_{0}\,r\,r^{\prime }\,\ddot{\theta }_{2}+2\,m_{\mathrm{ACT}}\,r\,\dot{r}\,\dot{\theta }_{1}+2\,m_{\mathrm{ACT}}\,r\,\dot{r}\,\dot{\theta }_{2}+ \\
    & 2\,m_{\mathrm{Obj}}\,r\,\dot{r}\,\dot{\theta }_{1}+2\,m_{\mathrm{Obj}}\,r\,\dot{r}\,\dot{\theta }_{2}+2\,m_{\mathrm{Obj}}\,r\,r^{\prime }\,\ddot{\theta }_{1}+2\,m_{\mathrm{Obj}}\,r\,r^{\prime }\,\ddot{\theta }_{2}+2\,m_{0}\,\dot{r}\,r^{\prime }\,\dot{\theta }_{1}+ \\
    & 2\,m_{0}\,\dot{r}\,r^{\prime }\,\dot{\theta }_{2}+2\,m_{\mathrm{Obj}}\,\dot{r}\,r^{\prime }\,\dot{\theta }_{1}+2\,m_{\mathrm{Obj}}\,\dot{r}\,r^{\prime }\,\dot{\theta }_{2}-\delta _{r}\,g\,m_{0}\,\sin\left(\theta _{2}+\theta _{1}\right)- \\
    & \delta _{r}\,g\,m_{\mathrm{Obj}}\,\sin\left(\theta _{2}+\theta _{1}\right)-g\,\ell_{2}\,m_{0}\,\sin\left(\theta _{2}+\theta _{1}\right)-g\,\ell_{2}\,m_{\mathrm{ACT}}\,\sin\left(\theta _{2}+\theta _{1}\right)-g\,\ell_{2}\,m_{\mathrm{Obj}}\,\sin\left(\theta _{2}+\theta _{1}\right)- \\
    & g\,\bar{\ell}_{2}\,m_{2}\,\sin\left(\theta _{2}+\theta _{1}\right)-g\,m_{0}\,r\,\sin\left(\theta _{2}+\theta _{1}\right)-g\,m_{\mathrm{ACT}}\,r\,\sin\left(\theta _{2}+\theta _{1}\right)-g\,m_{\mathrm{Obj}}\,r\,\sin\left(\theta _{2}+\theta _{1}\right)- \\
    & g\,m_{0}\,r^{\prime }\,\sin\left(\theta _{2}+\theta _{1}\right)-g\,m_{\mathrm{Obj}}\,r^{\prime }\,\sin\left(\theta _{2}+\theta _{1}\right)+1.1781\,R^4\,r\,\varrho _{\mathrm{ACT}}\,\ddot{\theta }_{1}+1.1781\,R^4\,r\,\varrho _{\mathrm{ACT}}\,\ddot{\theta }_{2}+ \\
    & 1.1781\,R^4\,\dot{r}\,\varrho _{\mathrm{ACT}}\,\dot{\theta }_{1}+1.1781\,R^4\,\dot{r}\,\varrho _{\mathrm{ACT}}\,\dot{\theta }_{2}+0.1309\,R^2\,r^3\,\varrho _{\mathrm{ACT}}\,\ddot{\theta }_{1}+0.1309\,R^2\,r^3\,\varrho _{\mathrm{ACT}}\,\ddot{\theta }_{2}- \\
    & 0.5000\,I_{\mathrm{xx},2}\,\dot{\phi }\,\dot{\theta }_{1}\,\sin\left(2\,\phi \right)-0.5000\,I_{\mathrm{xx},2}\,\dot{\phi }\,\dot{\theta }_{2}\,\sin\left(2\,\phi \right)+0.5000\,I_{\mathrm{yy},2}\,\dot{\phi }\,\dot{\theta }_{1}\,\sin\left(2\,\phi \right)+ \\
    & 0.5000\,I_{\mathrm{yy},2}\,\dot{\phi }\,\dot{\theta }_{2}\,\sin\left(2\,\phi \right)+0.2500\,I_{\mathrm{xx},2}\,\dot{\phi }\,\dot{\theta }_{1}\,\sin\left(-2\,\phi +2\,\theta _{2}+2\,\theta _{1}\right)- \\
    & 0.2500\,I_{\mathrm{xx},2}\,\dot{\phi }\,\dot{\theta }_{1}\,\sin\left(2\,\phi +2\,\theta _{2}+2\,\theta _{1}\right)+0.2500\,I_{\mathrm{xx},2}\,\dot{\phi }\,\dot{\theta }_{2}\,\sin\left(-2\,\phi +2\,\theta _{2}+2\,\theta _{1}\right)- \\
\end{align*}
\newpage
\begin{align*}
    & 0.2500\,I_{\mathrm{xx},2}\,\dot{\phi }\,\dot{\theta }_{2}\,\sin\left(2\,\phi +2\,\theta _{2}+2\,\theta _{1}\right)-0.2500\,I_{\mathrm{yy},2}\,\dot{\phi }\,\dot{\theta }_{1}\,\sin\left(-2\,\phi +2\,\theta _{2}+2\,\theta _{1}\right)+ \\
    & 0.2500\,I_{\mathrm{yy},2}\,\dot{\phi }\,\dot{\theta }_{1}\,\sin\left(2\,\phi +2\,\theta _{2}+2\,\theta _{1}\right)-0.2500\,I_{\mathrm{yy},2}\,\dot{\phi }\,\dot{\theta }_{2}\,\sin\left(-2\,\phi +2\,\theta _{2}+2\,\theta _{1}\right)+ \\
    & 0.2500\,I_{\mathrm{yy},2}\,\dot{\phi }\,\dot{\theta }_{2}\,\sin\left(2\,\phi +2\,\theta _{2}+2\,\theta _{1}\right)-0.5000\,{\delta _{r}}^2\,m_{0}\,{\dot{\phi }}^2\,\sin\left(2\,\theta _{2}+2\,\theta _{1}\right)- \\
    & 0.5000\,{\delta _{r}}^2\,m_{\mathrm{Obj}}\,{\dot{\phi }}^2\,\sin\left(2\,\theta _{2}+2\,\theta _{1}\right)-0.2500\,I_{\mathrm{xx},2}\,\dot{\theta }_{1}\,\dot{\theta }_{2}\,\sin\left(-2\,\phi +2\,\theta _{2}+2\,\theta _{1}\right)- \\
    & 0.2500\,I_{\mathrm{xx},2}\,\dot{\theta }_{1}\,\dot{\theta }_{2}\,\sin\left(2\,\phi +2\,\theta _{2}+2\,\theta _{1}\right)+0.2500\,I_{\mathrm{yy},2}\,\dot{\theta }_{1}\,\dot{\theta }_{2}\,\sin\left(-2\,\phi +2\,\theta _{2}+2\,\theta _{1}\right)+ \\
    & 0.2500\,I_{\mathrm{yy},2}\,\dot{\theta }_{1}\,\dot{\theta }_{2}\,\sin\left(2\,\phi +2\,\theta _{2}+2\,\theta _{1}\right)-0.5000\,{\ell_{2}}^2\,m_{0}\,{\dot{\phi }}^2\,\sin\left(2\,\theta _{2}+2\,\theta _{1}\right)- \\
    & 0.5000\,{\ell_{2}}^2\,m_{\mathrm{ACT}}\,{\dot{\phi }}^2\,\sin\left(2\,\theta _{2}+2\,\theta _{1}\right)-0.5000\,{\ell_{2}}^2\,m_{\mathrm{Obj}}\,{\dot{\phi }}^2\,\sin\left(2\,\theta _{2}+2\,\theta _{1}\right)- \\
    & 0.5000\,{\bar{\ell}_{2}}^2\,m_{2}\,{\dot{\phi }}^2\,\sin\left(2\,\theta _{2}+2\,\theta _{1}\right)-0.5000\,m_{0}\,{\dot{\phi }}^2\,r^2\,\sin\left(2\,\theta _{2}+2\,\theta _{1}\right)-0.5000\,m_{\mathrm{ACT}}\,{\dot{\phi }}^2\,r^2\,\sin\left(2\,\theta _{2}+2\,\theta _{1}\right) \\
    & -0.5000\,m_{\mathrm{Obj}}\,{\dot{\phi }}^2\,r^2\,\sin\left(2\,\theta _{2}+2\,\theta _{1}\right)-0.5000\,m_{0}\,{\dot{\phi }}^2\,{r^{\prime }}^2\,\sin\left(2\,\theta _{2}+2\,\theta _{1}\right)- \\
    & 0.5000\,m_{\mathrm{Obj}}\,{\dot{\phi }}^2\,{r^{\prime }}^2\,\sin\left(2\,\theta _{2}+2\,\theta _{1}\right)+0.5000\,\ell_{0}\,\ell_{2}\,m_{0}\,{\dot{\phi }}^2\,\sin\left(-\vartheta+\theta _{2}+\theta _{1}\right)+ \\
    & 0.5000\,\ell_{0}\,\ell_{2}\,m_{\mathrm{ACT}}\,{\dot{\phi }}^2\,\sin\left(-\vartheta+\theta _{2}+\theta _{1}\right)+0.5000\,\ell_{0}\,\ell_{2}\,m_{\mathrm{Obj}}\,{\dot{\phi }}^2\,\sin\left(-\vartheta+\theta _{2}+\theta _{1}\right)+ \\
    & 0.5000\,\ell_{0}\,\bar{\ell}_{2}\,m_{2}\,{\dot{\phi }}^2\,\sin\left(-\vartheta+\theta _{2}+\theta _{1}\right)+0.5000\,\ell_{0}\,m_{0}\,{\dot{\phi }}^2\,r\,\sin\left(-\vartheta+\theta _{2}+\theta _{1}\right)+ \\
    & 0.5000\,\ell_{0}\,m_{\mathrm{ACT}}\,{\dot{\phi }}^2\,r\,\sin\left(-\vartheta+\theta _{2}+\theta _{1}\right)+0.5000\,\ell_{0}\,m_{\mathrm{Obj}}\,{\dot{\phi }}^2\,r\,\sin\left(-\vartheta+\theta _{2}+\theta _{1}\right)+ \\
    & 0.5000\,\ell_{0}\,m_{0}\,{\dot{\phi }}^2\,r^{\prime }\,\sin\left(-\vartheta+\theta _{2}+\theta _{1}\right)+0.5000\,\ell_{0}\,m_{\mathrm{Obj}}\,{\dot{\phi }}^2\,r^{\prime }\,\sin\left(-\vartheta+\theta _{2}+\theta _{1}\right)+ \\
    & 0.1309\,R^2\,r^3\,\varrho _{\mathrm{ACT}}\,\ddot{\theta }_{1}\,\cos\left(2\,\theta _{2}+2\,\theta _{1}\right)+0.1309\,R^2\,r^3\,\varrho _{\mathrm{ACT}}\,\ddot{\theta }_{2}\,\cos\left(2\,\theta _{2}+2\,\theta _{1}\right)+ \\
    & 0.3927\,R^2\,r^2\,\dot{r}\,\varrho _{\mathrm{ACT}}\,\dot{\theta }_{1}+0.3927\,R^2\,r^2\,\dot{r}\,\varrho _{\mathrm{ACT}}\,\dot{\theta }_{2}+0.3927\,R^4\,r\,\varrho _{\mathrm{ACT}}\,{\dot{\theta }_{1}}^2\,\sin\left(2\,\theta _{2}+2\,\theta _{1}\right)+ \\
    & 0.3927\,R^4\,r\,\varrho _{\mathrm{ACT}}\,{\dot{\theta }_{2}}^2\,\sin\left(2\,\theta _{2}+2\,\theta _{1}\right)+0.5000\,\delta _{r}\,\ell_{1}\,m_{0}\,{\dot{\phi }}^2\,\sin\left(\theta _{2}\right)+ \\
    & 0.5000\,\delta _{r}\,\ell_{1}\,m_{\mathrm{Obj}}\,{\dot{\phi }}^2\,\sin\left(\theta _{2}\right)+\delta _{r}\,\ell_{1}\,m_{0}\,{\dot{\theta }_{1}}^2\,\sin\left(\theta _{2}\right)+\delta _{r}\,\ell_{1}\,m_{\mathrm{Obj}}\,{\dot{\theta }_{1}}^2\,\sin\left(\theta _{2}\right)+ \\
    & 0.5000\,\ell_{1}\,\ell_{2}\,m_{0}\,{\dot{\phi }}^2\,\sin\left(\theta _{2}\right)+0.5000\,\ell_{1}\,\ell_{2}\,m_{\mathrm{ACT}}\,{\dot{\phi }}^2\,\sin\left(\theta _{2}\right)+0.5000\,\ell_{1}\,\ell_{2}\,m_{\mathrm{Obj}}\,{\dot{\phi }}^2\,\sin\left(\theta _{2}\right)+ \\
    & 0.5000\,\ell_{1}\,\bar{\ell}_{2}\,m_{2}\,{\dot{\phi }}^2\,\sin\left(\theta _{2}\right)+\ell_{1}\,\ell_{2}\,m_{0}\,{\dot{\theta }_{1}}^2\,\sin\left(\theta _{2}\right)+\ell_{1}\,\ell_{2}\,m_{\mathrm{ACT}}\,{\dot{\theta }_{1}}^2\,\sin\left(\theta _{2}\right)+ \\
\end{align*}
\newpage
\begin{align*}
    & \ell_{1}\,\ell_{2}\,m_{\mathrm{Obj}}\,{\dot{\theta }_{1}}^2\,\sin\left(\theta _{2}\right)+\ell_{1}\,\bar{\ell}_{2}\,m_{2}\,{\dot{\theta }_{1}}^2\,\sin\left(\theta _{2}\right)+0.5000\,\ell_{1}\,m_{0}\,{\dot{\phi }}^2\,r\,\sin\left(\theta _{2}\right)+ \\
    & 0.5000\,\ell_{1}\,m_{\mathrm{ACT}}\,{\dot{\phi }}^2\,r\,\sin\left(\theta _{2}\right)+0.5000\,\ell_{1}\,m_{\mathrm{Obj}}\,{\dot{\phi }}^2\,r\,\sin\left(\theta _{2}\right)+0.5000\,\ell_{1}\,m_{0}\,{\dot{\phi }}^2\,r^{\prime }\,\sin\left(\theta _{2}\right)+ \\
    & 0.5000\,\ell_{1}\,m_{\mathrm{Obj}}\,{\dot{\phi }}^2\,r^{\prime }\,\sin\left(\theta _{2}\right)+\ell_{1}\,m_{0}\,r\,{\dot{\theta }_{1}}^2\,\sin\left(\theta _{2}\right)+\ell_{1}\,m_{\mathrm{ACT}}\,r\,{\dot{\theta }_{1}}^2\,\sin\left(\theta _{2}\right)+ \\
    & \ell_{1}\,m_{\mathrm{Obj}}\,r\,{\dot{\theta }_{1}}^2\,\sin\left(\theta _{2}\right)+\ell_{1}\,m_{0}\,r^{\prime }\,{\dot{\theta }_{1}}^2\,\sin\left(\theta _{2}\right)+\ell_{1}\,m_{\mathrm{Obj}}\,r^{\prime }\,{\dot{\theta }_{1}}^2\,\sin\left(\theta _{2}\right)- \\
    & 0.5000\,\delta _{r}\,\ell_{1}\,m_{0}\,{\dot{\phi }}^2\,\sin\left(\theta _{2}+2\,\theta _{1}\right)-0.5000\,\delta _{r}\,\ell_{1}\,m_{\mathrm{Obj}}\,{\dot{\phi }}^2\,\sin\left(\theta _{2}+2\,\theta _{1}\right)- \\
    & 0.5000\,\ell_{1}\,\ell_{2}\,m_{0}\,{\dot{\phi }}^2\,\sin\left(\theta _{2}+2\,\theta _{1}\right)-0.5000\,\ell_{1}\,\ell_{2}\,m_{\mathrm{ACT}}\,{\dot{\phi }}^2\,\sin\left(\theta _{2}+2\,\theta _{1}\right)- \\
    & 0.5000\,\ell_{1}\,\ell_{2}\,m_{\mathrm{Obj}}\,{\dot{\phi }}^2\,\sin\left(\theta _{2}+2\,\theta _{1}\right)-0.5000\,\ell_{1}\,\bar{\ell}_{2}\,m_{2}\,{\dot{\phi }}^2\,\sin\left(\theta _{2}+2\,\theta _{1}\right)- \\
    & 0.5000\,\ell_{1}\,m_{0}\,{\dot{\phi }}^2\,r\,\sin\left(\theta _{2}+2\,\theta _{1}\right)-0.5000\,\ell_{1}\,m_{\mathrm{ACT}}\,{\dot{\phi }}^2\,r\,\sin\left(\theta _{2}+2\,\theta _{1}\right)- \\
    & 0.5000\,\ell_{1}\,m_{\mathrm{Obj}}\,{\dot{\phi }}^2\,r\,\sin\left(\theta _{2}+2\,\theta _{1}\right)-0.5000\,\ell_{1}\,m_{0}\,{\dot{\phi }}^2\,r^{\prime }\,\sin\left(\theta _{2}+2\,\theta _{1}\right)- \\
    & 0.5000\,\ell_{1}\,m_{\mathrm{Obj}}\,{\dot{\phi }}^2\,r^{\prime }\,\sin\left(\theta _{2}+2\,\theta _{1}\right)-0.5000\,\delta _{r}\,\ell_{0}\,m_{0}\,{\dot{\phi }}^2\,\sin\left(\vartheta+\theta _{2}+\theta _{1}\right)- \\
    & 0.5000\,\delta _{r}\,\ell_{0}\,m_{\mathrm{Obj}}\,{\dot{\phi }}^2\,\sin\left(\vartheta+\theta _{2}+\theta _{1}\right)-0.1309\,R^2\,r^3\,\varrho _{\mathrm{ACT}}\,{\dot{\theta }_{1}}^2\,\sin\left(2\,\theta _{2}+2\,\theta _{1}\right)- \\
    & 0.1309\,R^2\,r^3\,\varrho _{\mathrm{ACT}}\,{\dot{\theta }_{2}}^2\,\sin\left(2\,\theta _{2}+2\,\theta _{1}\right)-0.5000\,\ell_{0}\,\ell_{2}\,m_{0}\,{\dot{\phi }}^2\,\sin\left(\vartheta+\theta _{2}+\theta _{1}\right)- \\
    & 0.5000\,\ell_{0}\,\ell_{2}\,m_{\mathrm{ACT}}\,{\dot{\phi }}^2\,\sin\left(\vartheta+\theta _{2}+\theta _{1}\right)-0.5000\,\ell_{0}\,\ell_{2}\,m_{\mathrm{Obj}}\,{\dot{\phi }}^2\,\sin\left(\vartheta+\theta _{2}+\theta _{1}\right)- \\
    & 0.5000\,\ell_{0}\,\bar{\ell}_{2}\,m_{2}\,{\dot{\phi }}^2\,\sin\left(\vartheta+\theta _{2}+\theta _{1}\right)-0.5000\,\ell_{0}\,m_{0}\,{\dot{\phi }}^2\,r\,\sin\left(\vartheta+\theta _{2}+\theta _{1}\right)- \\
    & 0.5000\,\ell_{0}\,m_{\mathrm{ACT}}\,{\dot{\phi }}^2\,r\,\sin\left(\vartheta+\theta _{2}+\theta _{1}\right)-0.5000\,\ell_{0}\,m_{\mathrm{Obj}}\,{\dot{\phi }}^2\,r\,\sin\left(\vartheta+\theta _{2}+\theta _{1}\right)- \\
    & 0.5000\,\ell_{0}\,m_{0}\,{\dot{\phi }}^2\,r^{\prime }\,\sin\left(\vartheta+\theta _{2}+\theta _{1}\right)-0.5000\,\ell_{0}\,m_{\mathrm{Obj}}\,{\dot{\phi }}^2\,r^{\prime }\,\sin\left(\vartheta+\theta _{2}+\theta _{1}\right)- \\
    & 0.3927\,R^4\,r\,\varrho _{\mathrm{ACT}}\,\ddot{\theta }_{1}\,\cos\left(2\,\theta _{2}+2\,\theta _{1}\right)-0.3927\,R^4\,r\,\varrho _{\mathrm{ACT}}\,\ddot{\theta }_{2}\,\cos\left(2\,\theta _{2}+2\,\theta _{1}\right)- \\
    & 0.3927\,R^4\,\dot{r}\,\varrho _{\mathrm{ACT}}\,\dot{\theta }_{1}\,\cos\left(2\,\theta _{2}+2\,\theta _{1}\right)-0.3927\,R^4\,\dot{r}\,\varrho _{\mathrm{ACT}}\,\dot{\theta }_{2}\,\cos\left(2\,\theta _{2}+2\,\theta _{1}\right)- \\
    & \delta _{r}\,\ell_{2}\,m_{0}\,{\dot{\phi }}^2\,\sin\left(2\,\theta _{2}+2\,\theta _{1}\right)-\delta _{r}\,\ell_{2}\,m_{\mathrm{Obj}}\,{\dot{\phi }}^2\,\sin\left(2\,\theta _{2}+2\,\theta _{1}\right)+\delta _{r}\,\ell_{1}\,m_{0}\,\ddot{\theta }_{1}\,\cos\left(\theta _{2}\right)+ \\
    & \delta _{r}\,\ell_{1}\,m_{\mathrm{Obj}}\,\ddot{\theta }_{1}\,\cos\left(\theta _{2}\right)-\delta _{r}\,m_{0}\,{\dot{\phi }}^2\,r\,\sin\left(2\,\theta _{2}+2\,\theta _{1}\right)-\delta _{r}\,m_{\mathrm{Obj}}\,{\dot{\phi }}^2\,r\,\sin\left(2\,\theta _{2}+2\,\theta _{1}\right)- \\
\end{align*}
\newpage
\begin{align*}
    & \delta _{r}\,m_{0}\,{\dot{\phi }}^2\,r^{\prime }\,\sin\left(2\,\theta _{2}+2\,\theta _{1}\right)-\delta _{r}\,m_{\mathrm{Obj}}\,{\dot{\phi }}^2\,r^{\prime }\,\sin\left(2\,\theta _{2}+2\,\theta _{1}\right)+\ell_{1}\,\ell_{2}\,m_{0}\,\ddot{\theta }_{1}\,\cos\left(\theta _{2}\right) \\
    & +\ell_{1}\,\ell_{2}\,m_{\mathrm{ACT}}\,\ddot{\theta }_{1}\,\cos\left(\theta _{2}\right)+\ell_{1}\,\ell_{2}\,m_{\mathrm{Obj}}\,\ddot{\theta }_{1}\,\cos\left(\theta _{2}\right)+\ell_{1}\,\bar{\ell}_{2}\,m_{2}\,\ddot{\theta }_{1}\,\cos\left(\theta _{2}\right)- \\
    & \ell_{2}\,m_{0}\,{\dot{\phi }}^2\,r\,\sin\left(2\,\theta _{2}+2\,\theta _{1}\right)-\ell_{2}\,m_{\mathrm{ACT}}\,{\dot{\phi }}^2\,r\,\sin\left(2\,\theta _{2}+2\,\theta _{1}\right)-\ell_{2}\,m_{\mathrm{Obj}}\,{\dot{\phi }}^2\,r\,\sin\left(2\,\theta _{2}+2\,\theta _{1}\right)- \\
    & \ell_{2}\,m_{0}\,{\dot{\phi }}^2\,r^{\prime }\,\sin\left(2\,\theta _{2}+2\,\theta _{1}\right)-\ell_{2}\,m_{\mathrm{Obj}}\,{\dot{\phi }}^2\,r^{\prime }\,\sin\left(2\,\theta _{2}+2\,\theta _{1}\right)+\ell_{1}\,m_{0}\,r\,\ddot{\theta }_{1}\,\cos\left(\theta _{2}\right)+ \\
    & \ell_{1}\,m_{\mathrm{ACT}}\,r\,\ddot{\theta }_{1}\,\cos\left(\theta _{2}\right)+\ell_{1}\,m_{\mathrm{Obj}}\,r\,\ddot{\theta }_{1}\,\cos\left(\theta _{2}\right)+\ell_{1}\,m_{0}\,r^{\prime }\,\ddot{\theta }_{1}\,\cos\left(\theta _{2}\right)+ \\
    & \ell_{1}\,m_{\mathrm{Obj}}\,r^{\prime }\,\ddot{\theta }_{1}\,\cos\left(\theta _{2}\right)-m_{0}\,{\dot{\phi }}^2\,r\,r^{\prime }\,\sin\left(2\,\theta _{2}+2\,\theta _{1}\right)-m_{\mathrm{Obj}}\,{\dot{\phi }}^2\,r\,r^{\prime }\,\sin\left(2\,\theta _{2}+2\,\theta _{1}\right)+ \\
    & 0.5000\,\delta _{r}\,\ell_{0}\,m_{0}\,{\dot{\phi }}^2\,\sin\left(-\vartheta+\theta _{2}+\theta _{1}\right)+0.5000\,\delta _{r}\,\ell_{0}\,m_{\mathrm{Obj}}\,{\dot{\phi }}^2\,\sin\left(-\vartheta+\theta _{2}+\theta _{1}\right)+ \\
    & 0.7854\,R^4\,r\,\varrho _{\mathrm{ACT}}\,\dot{\theta }_{1}\,\dot{\theta }_{2}\,\sin\left(2\,\theta _{2}+2\,\theta _{1}\right)+0.3927\,R^2\,r^2\,\dot{r}\,\varrho _{\mathrm{ACT}}\,\dot{\theta }_{1}\,\cos\left(2\,\theta _{2}+2\,\theta _{1}\right)+ \\
    & 0.3927\,R^2\,r^2\,\dot{r}\,\varrho _{\mathrm{ACT}}\,\dot{\theta }_{2}\,\cos\left(2\,\theta _{2}+2\,\theta _{1}\right)-0.2618\,R^2\,r^3\,\varrho _{\mathrm{ACT}}\,\dot{\theta }_{1}\,\dot{\theta }_{2}\,\sin\left(2\,\theta _{2}+2\,\theta _{1}\right) \\
\end{align*}

 \newpage
\subsubsection{Azimuthal Coordinate}
\begin{align*}
u_\phi =& \frac{d}{dt}\left(\frac{\partial \mathcal{L}}{\partial \dot{\phi}}\right) - \dfrac{\partial \mathcal{L}}{\partial \phi} + \dfrac{\partial \mathcal{R}}{\partial \dot{\phi}}\\
u_{\phi}&= \frac{\partial \mathcal{R}}{\partial \dot{\phi}} 
   + \kappa_{\phi}(\phi-\phi_{0})
\\[4pt]
&\quad
+ 0.0833\,\ddot{\phi}\Big(
12 I_{\mathrm{yy},0}
+12 I_{\mathrm{yy},1}
+12 I_{\mathrm{yy},2}
+12 I_{\mathrm{xx},0}\sin^{2}\phi
+12 I_{\mathrm{xx},1}\sin^{2}\phi
+12 I_{\mathrm{xx},2}\sin^{2}\phi
\\[-2pt]
&\qquad\qquad
-12 I_{\mathrm{yy},0}\sin^{2}\phi
-12 I_{\mathrm{yy},1}\sin^{2}\phi
-12 I_{\mathrm{yy},2}\sin^{2}\phi
+ \pi R^{2} r^{3}\varrho_{\mathrm{ACT}}
+ 3\pi R^{4} r\,\varrho_{\mathrm{ACT}}
\Big)
\\[6pt]
&\quad
+ 0.2500\,\dot\phi\Big(
\pi R^{4}\dot r\,\varrho_{\mathrm{ACT}}
+ \pi R^{2}r^{2}\dot r\,\varrho_{\mathrm{ACT}}
\\[-2pt]
&\qquad\qquad
-4\sin(2\phi)
 (I_{\mathrm{yy},2}+I_{\mathrm{yy},1}+I_{\mathrm{yy},0})\dot\phi
+4\sin(2\phi)
 (I_{\mathrm{xx},2}+I_{\mathrm{xx},1}+I_{\mathrm{xx},0})\dot\phi
\Big)
\\[6pt]
&\quad
- \dot{\phi}^{2}\sin(2\phi)
\Big(
-0.5(I_{\mathrm{yy},2}+I_{\mathrm{yy},1}+I_{\mathrm{yy},0})
+0.5(I_{\mathrm{xx},2}+I_{\mathrm{xx},1}+I_{\mathrm{xx},0})
\Big)
\\[8pt]
&\quad
+ (m_{\mathrm{Obj}}+m_{0})
\Big(
\sin(\theta_{1}+\theta_{2})\sin\phi\,(r'+r+\ell_{2}+\delta_{r})
+ \ell_{1}\sin\phi\sin\theta_{1}
+ \ell_{0}\sin\phi\sin\vartheta
\Big)
\\[2pt]
&\qquad\times
\Big(
\ell_{1}\ddot{\phi}\sin\phi\sin\theta_{1}
 -\ddot r \sin(\theta_{1}+\theta_{2})\cos\phi
 +\ell_{0}\ddot{\phi}\sin\phi\sin\vartheta
\\[-2pt]
&\qquad\quad
 -\cos(\theta_{1}+\theta_{2})\cos\phi(\ddot\theta_{1}+\ddot\theta_{2})
   (r'+r+\ell_{2}+\delta_{r})
\\[-2pt]
&\qquad\quad
 +\ell_{1}\dot\phi^{2}\cos\phi\sin\theta_{1}
 +\ell_{0}\dot\phi^{2}\cos\phi\sin\vartheta
 +\ell_{1}\dot\theta_{1}^{2}\cos\phi\sin\theta_{1}
\\[-2pt]
&\qquad\quad
 +\ddot{\phi}\sin(\theta_{1}+\theta_{2})\sin\phi\,(r'+r+\ell_{2}+\delta_{r})
\\[-2pt]
&\qquad\quad
 -2\dot r\cos(\theta_{1}+\theta_{2})\cos\phi(\dot\theta_{1}+\dot\theta_{2})
\\[-2pt]
&\qquad\quad
 +\sin(\theta_{1}+\theta_{2})\cos\phi(\dot\theta_{1}+\dot\theta_{2})^{2}
   (r'+r+\ell_{2}+\delta_{r})
\\[-2pt]
&\qquad\quad
 +\dot\phi^{2}\sin(\theta_{1}+\theta_{2})\cos\phi(r'+r+\ell_{2}+\delta_{r})
\\[-2pt]
&\qquad\quad
 +2\dot\phi\dot r\sin(\theta_{1}+\theta_{2})\sin\phi
 -\ell_{1}\ddot\theta_{1}\cos\phi\cos\theta_{1}
\\[-2pt]
&\qquad\quad
 + 2\ell_{1}\dot\phi\dot\theta_{1}\cos\theta_{1}\sin\phi
 + 2\dot\phi\cos(\theta_{1}+\theta_{2})\sin\phi(\dot\theta_{1}+\dot\theta_{2})
   (r'+r+\ell_{2}+\delta_{r})
\Big)
\end{align*}

\begin{align*}
&\quad
+ m_{2}\cos\phi\,
\Big(
   \bar\ell_{2}\sin(\theta_{1}+\theta_{2})
 +\ell_{1}\sin\theta_{1}
 +\ell_{0}\sin\vartheta
\Big)
\\[2pt]
&\qquad\times
\Big(
   \bar\ell_{2}\cos(\theta_{1}+\theta_{2})\sin\phi(\ddot\theta_{1}+\ddot\theta_{2})
 - \bar\ell_{2}\dot\phi^{2}\sin(\theta_{1}+\theta_{2})\sin\phi
 -\ell_{1}\dot\phi^{2}\sin\phi\sin\theta_{1}
\\[-2pt]
&\qquad\quad
 -\ell_{0}\dot\phi^{2}\sin\phi\sin\vartheta
 -\ell_{1}\dot\theta_{1}^{2}\sin\phi\sin\theta_{1}
 - \bar\ell_{2}\sin(\theta_{1}+\theta_{2})\sin\phi(\dot\theta_{1}+\dot\theta_{2})^{2}
\\[-2pt]
&\qquad\quad
 + \bar\ell_{2}\ddot\phi\,\sin(\theta_{1}+\theta_{2})\cos\phi
 +\ell_{1}\ddot\phi\cos\phi\sin\theta_{1}
 +\ell_{0}\ddot\phi\cos\phi\sin\vartheta
\\[-2pt]
&\qquad\quad
 +\ell_{1}\ddot\theta_{1}\cos\theta_{1}\sin\phi
 + 2\bar\ell_{2}\dot\phi\cos(\theta_{1}+\theta_{2})\cos\phi(\dot\theta_{1}+\dot\theta_{2})
\\[-2pt]
&\qquad\quad
 + 2l_{1}\dot\phi\dot\theta_{1}\cos\phi\cos\theta_{1}
\Big)
\\[10pt]
&\quad
+ \cos^{2}(\theta_{1}+\theta_{2})\,\cos\phi\,\sin\phi\,
   (\dot\theta_{1}+\dot\theta_{2})^{2}\,
   \big(-I_{\mathrm{yy},2}+I_{\mathrm{xx},2}\big)
\\[6pt]
&\quad
+ \dot\theta_{1}^{2}\,\cos\phi\,\cos^{2}\theta_{1}\,\sin\phi\,
   \big(-I_{\mathrm{yy},1}+I_{\mathrm{xx},1}\big)
\\[6pt]
&\quad
+\ell_{0}^{2}m_{\mathrm{BF}}\ddot\phi\,
   \cos^{2}\phi\,\sin^{2}\vartheta
\;+\;
 \ell_{0}^{2}m_{\mathrm{BF}}\ddot\phi\,
   \sin^{2}\phi\,\sin^{2}\vartheta
\end{align*}
\newpage
\vspace{2em}
\section{Structural Mechanics and Reaction Analysis}
    \subsection{Reaction-Analysis}
Now that the dynamics are analyzed, let us determine the reactions at the Gearshaft for the azimuthal coordinate, as seen in the section with the Euler-Lagrange equations of motion, $\phi$ does \textbf{NOT} depend on gravity; and this makes sense conceptually as phi is confined to the longitudinal/ lateral motion of the robotic apparatus,  and thus we can now define the structural mechanics of the Robotic Apparatus: 

Let us consider the Free-Body Diagram of the Robotic Apparatus Shown below at the base-frame: 

\begin{figure}
    \centering
    \includegraphics[width=\linewidth]{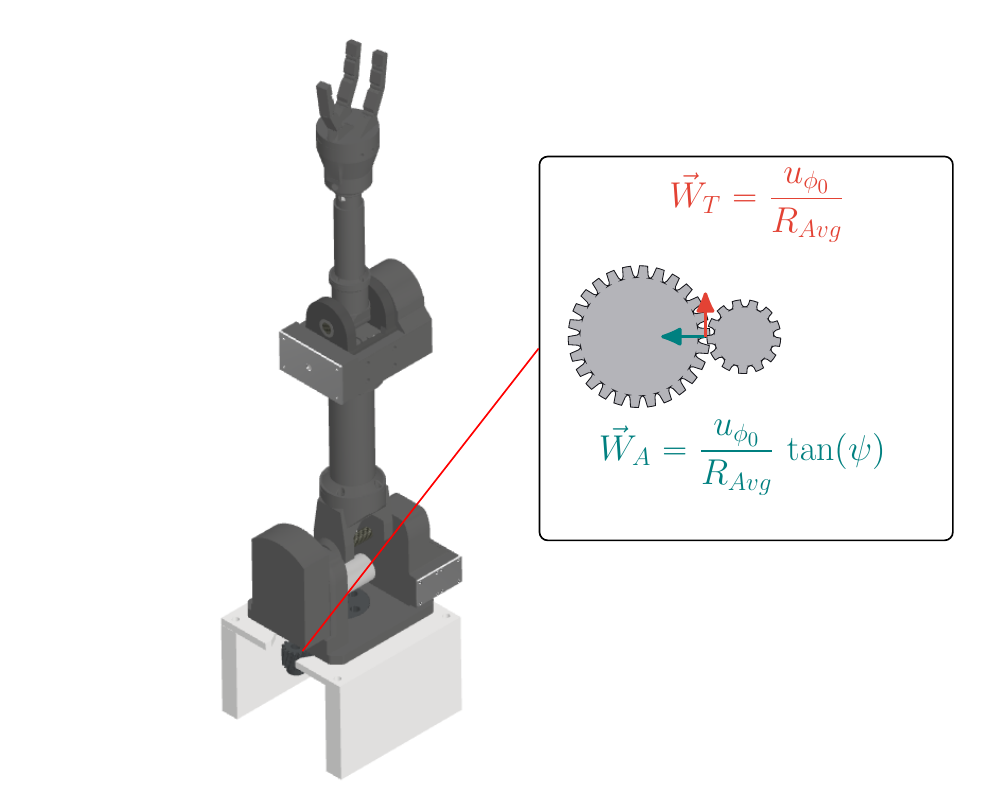}
    \caption{Gear Diagram in the white internal gearbox as shown above. The axial force is in the negative x-direction, and the tangential vector is in the positive y-direction, and the radial vector in the negative z-direction, which uses the same dynamics frame of reference.}
    \label{fig:placeholder}
\end{figure}

The weight vector, is given as follows and is denoted in terms of one of the unknown coordinates. 
\begin{align*}
       \vec{W}=  \left(\begin{array}{c} -\frac{G_{p}\,u_{\phi ,0}\,\mathrm{tan}\left(\psi \right)}{R_{\mathrm{avg}}}\\ \frac{G_{p}\,u_{\phi ,0}}{R_{\mathrm{avg}}}\\ -\frac{G_{p}\,u_{\phi ,0}\,\mathrm{tan}\left(\varphi \right)}{R_{\mathrm{avg}}} \end{array}\right)
\end{align*}
\newpage
\vspace{6em}
The forces in this analysis are strictly done in terms of the self-weight plus the mass of the object; which then, the summation of forces (denoted in cartesian for brevity), are given as follows: 
\begin{align*}
    \sum \vec{F}_{x}  = A_{x}+B_{x}-\frac{G_{p}\,u_{\phi ,0}\,\mathrm{tan}\left(\psi \right)}{R_{\mathrm{avg}}}=0
\end{align*}
\begin{align*}
    \sum \vec{F}_{y}  = A_{y}+B_{y}+\frac{G_{p}\,u_{\phi ,0}}{R_{\mathrm{avg}}}=0
\end{align*}
\begin{align*}
    \sum \vec{F}_{z} = B_{z}-g\,m_{1}-g\,m_{2}-g\,m_{\mathrm{ACT}}-g\,m_{\mathrm{BF}}-g\,m_{\mathrm{Obj}}-\frac{G_{p}\,u_{\phi ,0}\,\mathrm{tan}\left(\varphi \right)}{R_{\mathrm{avg}}}=0
\end{align*}
The corresponding moment based reaction vectors, are then given as follows: 
\begin{align*}
\sum \vec{M}_{x} = B_{z}\,l_{G,y}+B_{y}\,H_{w}-g\,m_{1}\,y_{1}-g\,m_{2}\,y_{2}-g\,m_{\mathrm{BF}}\,y_{0}-g\,m_{\mathrm{ACT}}\,y_{A}-g\,m_{\mathrm{Obj}}\,y_{\mathrm{Obj}}+\frac{G_{p}\,h_{w}\,u_{\phi ,0}}{R_{\mathrm{avg}}}=0
\end{align*}
\begin{align*}
\sum \vec{M}_{y} = G_{p}\,u_{\phi ,0}\,\mathrm{tan}\left(\varphi \right)-B_{x}\,H_{w}-B_{z}\,l_{G,x}+g\,m_{1}\,x_{1}+g\,m_{2}\,x_{2}+ \\g\,m_{\mathrm{BF}}\,x_{0}+g\,m_{\mathrm{ACT}}\,x_{A}+g\,m_{\mathrm{Obj}}\,x_{\mathrm{Obj}}+\frac{G_{p}\,h_{w}\,u_{\phi ,0}\,\mathrm{tan}\left(\psi \right)}{R_{\mathrm{avg}}}=0
\end{align*}
\begin{align*}
   \sum \vec{M}_{z} =  -B_{x}\,l_{G,y}+B_{y}\,l_{G,x}+G_{p}\,u_{\phi ,0}=0
\end{align*}

The corresponding stiffness matrix, is then given by: 
\begin{align*}
\mathbf{K} =
\begin{bmatrix}
1 & 0 & -\dfrac{G_{p}\tan(\psi)}{R_{\mathrm{avg}}} & 1 & 0 & 0 \\
0 & 0 & \dfrac{G_{p}h_{w}}{R_{\mathrm{avg}}} & 0 & H_{w} & l_{G,y} \\
0 & 1 & \dfrac{G_{p}}{R_{\mathrm{avg}}} & 0 & 1 & 0 \\
0 & 0 & G_{p}\tan(\varphi)+\dfrac{G_{p}h_{w}\tan(\psi)}{R_{\mathrm{avg}}} & -H_{w} & 0 & -l_{G,x} \\
0 & 0 & -\dfrac{G_{p}\tan(\varphi)}{R_{\mathrm{avg}}} & 0 & 0 & 1 \\
0 & 0 & G_{p} & -l_{G,y} & l_{G,x} & 0
\end{bmatrix}\\
\vec{q} = \begin{bmatrix}
0 \\[4pt]
g\,m_{\mathrm{Obj}}\,y_{\mathrm{Obj}}
+ g\,m_{\mathrm{ACT}}\,y_{A}
+ g\,m_{2}\,y_{2}
+ g\,m_{1}\,y_{1}
+ g\,m_{\mathrm{BF}}\,y_{0} \\[4pt]
0 \\[4pt]
-g\,m_{\mathrm{Obj}}\,x_{\mathrm{Obj}}
- g\,m_{\mathrm{ACT}}\,x_{A}
- g\,m_{2}\,x_{2}
- g\,m_{1}\,x_{1}
- g\,m_{\mathrm{BF}}\,x_{0} \\[4pt]
g\,m_{\mathrm{Obj}}
+ g\,m_{\mathrm{BF}}
+ g\,m_{\mathrm{ACT}}
+ g\,m_{2}
+ g\,m_{1} \\[4pt]
0
\end{bmatrix}
\end{align*}
The reaction vector, is given as follows: 
\begin{align*}
    \vec{R} = \begin{bmatrix}
A_{x} \\
A_{y} \\
u_{\phi,0} \\
B_{x} \\
B_{y} \\
B_{z}
\end{bmatrix}
\end{align*}
Inverting the stiffness matrix and multiplying by the vector $\vec{q}$ yields the reaction forces and the required torque for the azimuthal coordinate; and the reaction torque is given as follows.
\begin{align*}
u_{\phi_{0}}
&=
\frac{
R_{\mathrm{avg}}\,g\,
\Big(
l_{G,x}\big(
m_{\mathrm{Obj}}\,y_{\mathrm{Obj}}
+ m_{\mathrm{ACT}}\,y_{A}
+ m_{2}\,y_{2}
+ m_{1}\,y_{1}
+ m_{\mathrm{BF}}\,y_{0}
\big)
}{
G_{p}\,\Big(
h_{w}\,l_{G,x}
- H_{w}\,R_{\mathrm{avg}}
+ R_{\mathrm{avg}}\,l_{G,y}\,\tan(\varphi)
+ h_{w}\,l_{G,y}\,\tan(\psi)
\Big)
}\\ &
- \frac{l_{G,y}\big(
m_{\mathrm{Obj}}\,x_{\mathrm{Obj}}
+ m_{\mathrm{ACT}}\,x_{A}
+ m_{2}\,x_{2}
+ m_{1}\,x_{1}
+ m_{\mathrm{BF}}\,x_{0}
\big)
\Big)}{G_{p}\,\Big(
h_{w}\,l_{G,x}
- H_{w}\,R_{\mathrm{avg}}
+ R_{\mathrm{avg}}\,l_{G,y}\,\tan(\varphi)
+ h_{w}\,l_{G,y}\,\tan(\psi)
\Big)}
\end{align*}
For brevity, since the other reactions are not used in the control-law; they will not be considered in the calculations - and note that the results of the Graviational Potential Gradinet will yield the other initial torque responses, and from the Euler-Lagrange Derivation; the gravity gradient is given by: 
\begin{align*}
G &= \frac{\partial V_{G.P.E}}{\partial q_i} \\
  &= 
  \begin{bmatrix}
  %
  % ===== FIRST COMPONENT =====
  &g\,\cos(\theta_1+\theta_2)\,(m_{\mathrm{Obj}}+m_0+m_{\mathrm{ACT}})
  \\[4pt]
  %
  % ===== SECOND COMPONENT =====
  &-g\left[
      (m_{\mathrm{Obj}}+m_0)
      \left(
          \sin(\theta_1+\theta_2)(r' + r + \ell_2 + \delta_r)
          + \ell_1 \sin\theta_1
      \right)
      \right. \\[2pt]
      &\qquad\qquad
      +\;m_{\mathrm{ACT}}
      \left(
          \sin(\theta_1+\theta_2)(r+\ell_2)
          + \ell_1 \sin\theta_1
      \right)
      \\[6pt]
      &\qquad\qquad
      +\;m_2\left(
          \bar{\ell}_2 \sin(\theta_1+\theta_2)
          + \ell_1 \sin\theta_1
      \right)
      \\[2pt]
      &\qquad\qquad
      \left.
      +\;\bar{\ell}_1 m_1 \sin\theta_1
  \right]
  \\[6pt]
  %
  % ===== THIRD COMPONENT =====
  &-g\left[
        m_{\mathrm{ACT}} \sin(\theta_1+\theta_2)(r+\ell_2)
        + \bar{\ell}_2 m_2 \sin(\theta_1+\theta_2)
      \right. \\[2pt]
      &\qquad\qquad
      \left.
      +\;(m_{\mathrm{Obj}}+m_0)
         \sin(\theta_1+\theta_2)(r' + r + \ell_2 + \delta_r)
  \right]
  \\[6pt]
  %
  % ===== FOURTH COMPONENT =====
 &0
  \end{bmatrix}
\end{align*}
Thus, the initial torque/ force vector will yield the required velocities for the Gradient based calculations; and $\tau_{0}$ is given through: 

\begin{align*}
   \therefore  \tau_{0}
    &=
    \frac{\partial}{\partial q_i}
    \!\left( g\sum_{i=1}^{4} m_i z_i \right)
    \begin{bmatrix}
        \hat{r} \\[3pt]
        \hat{\theta}_{1} \\[3pt]
        \hat{\theta}_{2} \\[3pt]
        0
    \end{bmatrix}
    + u_{\phi_{0}}\,\hat{\phi}.
\end{align*}
To evaluate the velocities for the Gradinet-Descent based calculations, it officially becomes: 
\begin{align*}
v_{0} = v_{N.L}\mathrm{sgn}\left( q_{i,f} -q_{i,0}\right)\left( 1 - \frac{\|\tau_{0}\|}{\tau_{Stall}}\right)
\end{align*}
\newpage
\section{Inertial Verification}
In this section of the appendix; we investigate how the values are derived from the CAD Model. This section describes the Inertia at the respected centers of mass for each component; in addition to its diagonalized inertia tensor. Here, the measurements were explicitly derived from \textit{Autodesk Inventor}. 
\begin{figure}[H]
    \centering
    \includegraphics[width=0.75\linewidth]{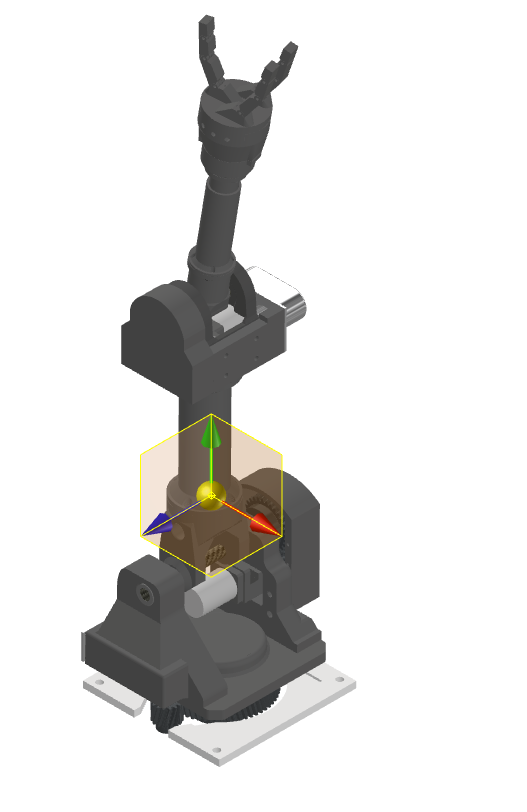}
\caption{Robotic arm CAD model. Claw design courtesy of Michael Kneoller.}
    \label{fig:placeholder}
\end{figure}
From the Diagram above, the Robotic Apparatus has a total center of mass distribution at the center of its structure; which is a good sign for the Mechanical Design. Let us discuss the centers of mass, and the inertial calculations for each of the components.
Since Autodesk Inventor does not use the conventional coordinate system used in Physics derivations; it must be converted properly. 
\subsection{Base Frame}
For the base frame, it is convenient to denote the Parallel-Axis Theory about the shaft mounting of Link 1; and this is done by first using coordinate transformations to weave out any possibility of accidentally including a rotated component. For the angle $\vartheta$, and the length $\ell_{0}$, they are obtained through trivial trigonometry. 
\begin{figure}[H]
    \centering
    \includegraphics[width=0.65\linewidth]{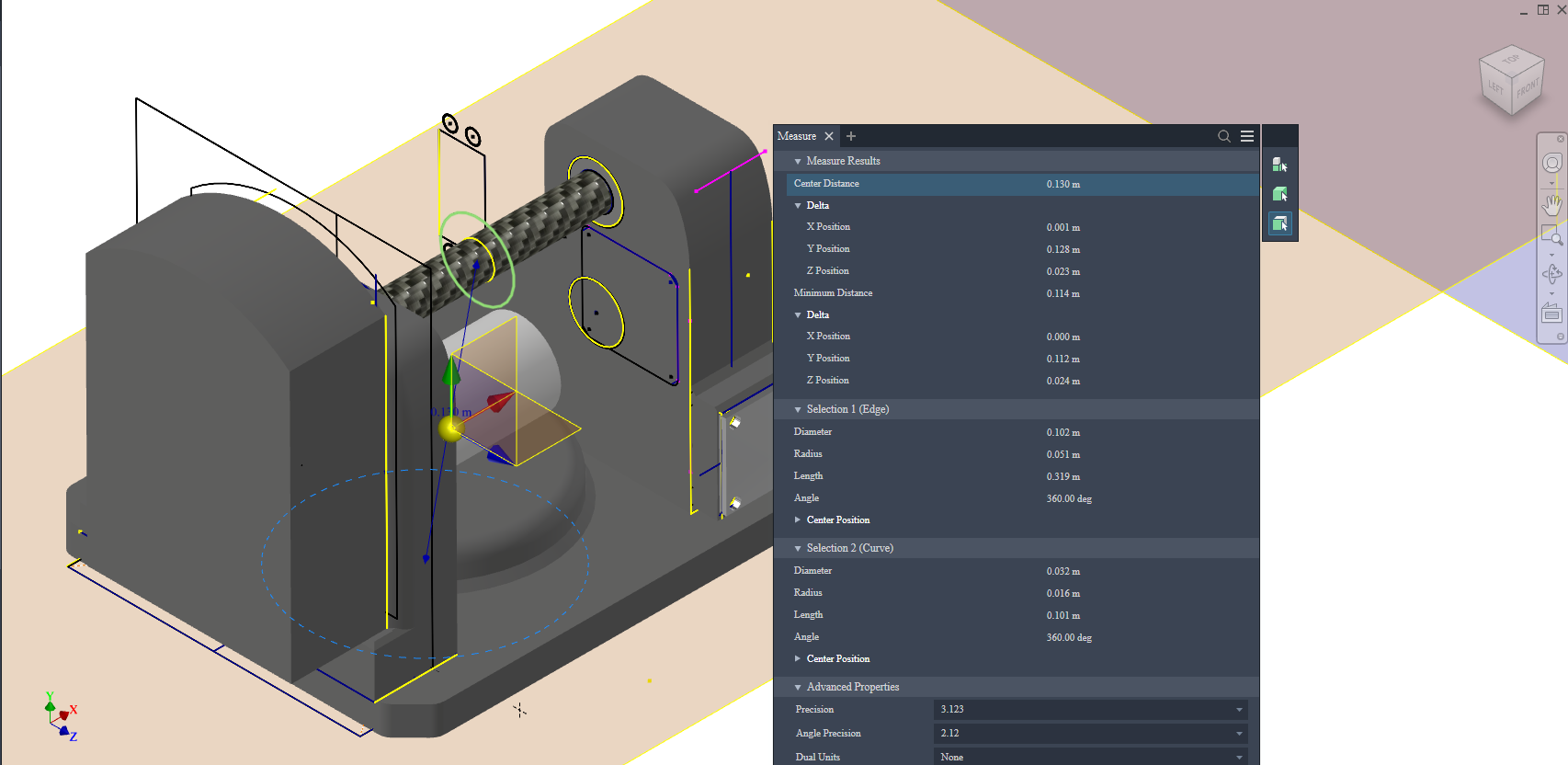}
    \caption{Base Frame CAD Assembly}
    \label{fig:placeholder}
\end{figure}

\begin{align*}
    \ell_{0} = 0.130 \quad \vartheta = \cos^{-1}\left(\frac{z_{0}}{\ell_{0}}\right) = 0.1756\, \textrm{rad}
\end{align*}
Additionally, since these distances do not change unless failure in Mechanical Function occurs, the parallel axis distances are thus taken as follows from the Autodesk Inventor measurements as follows, and for each realtive distance; consider the diagram as shown above. 
\begin{figure}[t]
    \centering
    \subfloat[Base Frame measurement for Parallel Axis Shifting (Longitudinal) \label{fig:link1a}]{
        \includegraphics[width=0.46\linewidth]{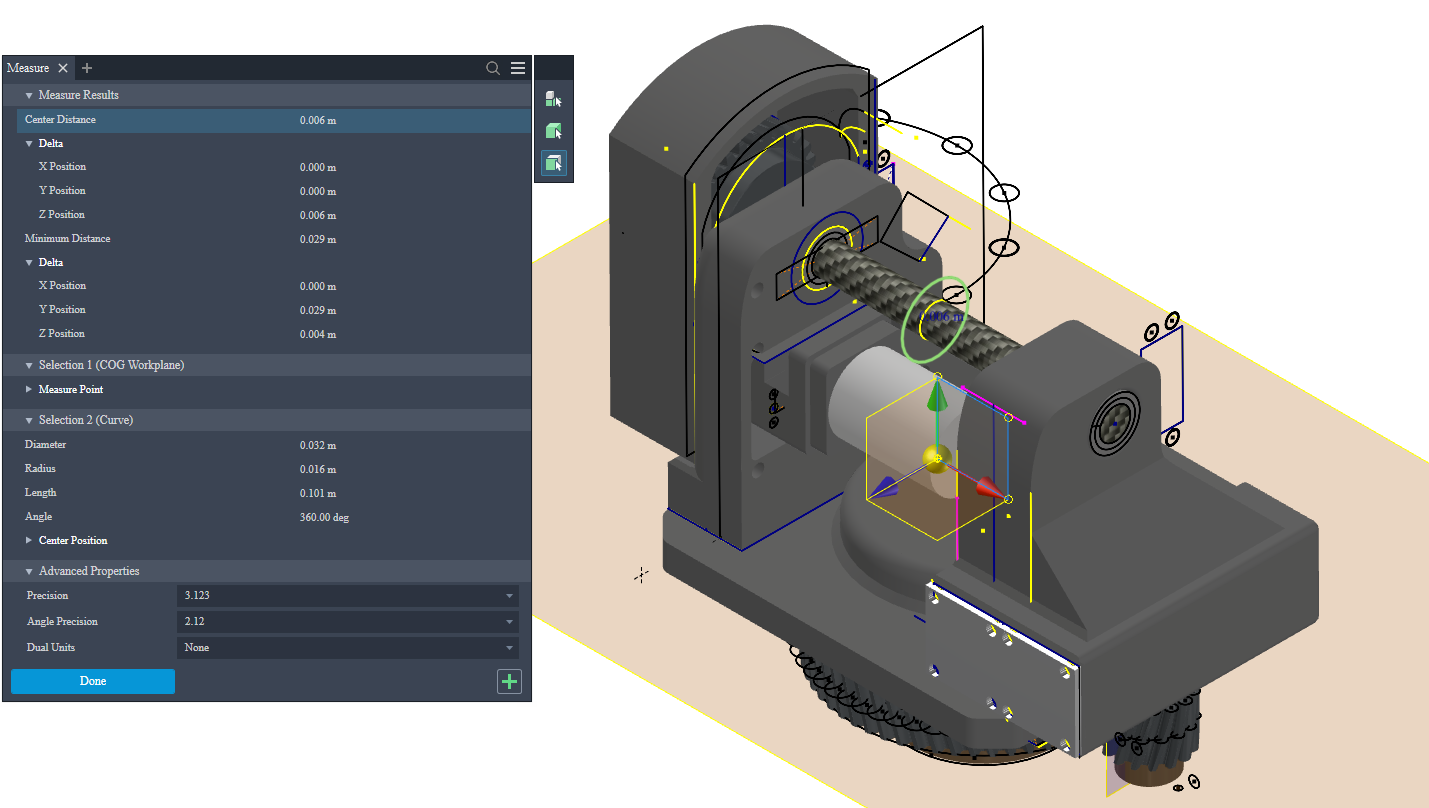}
    }\hfill
    \subfloat[Base Frame measurement (Lateral)\label{fig:link1b}]{
        \includegraphics[width=0.46\linewidth]{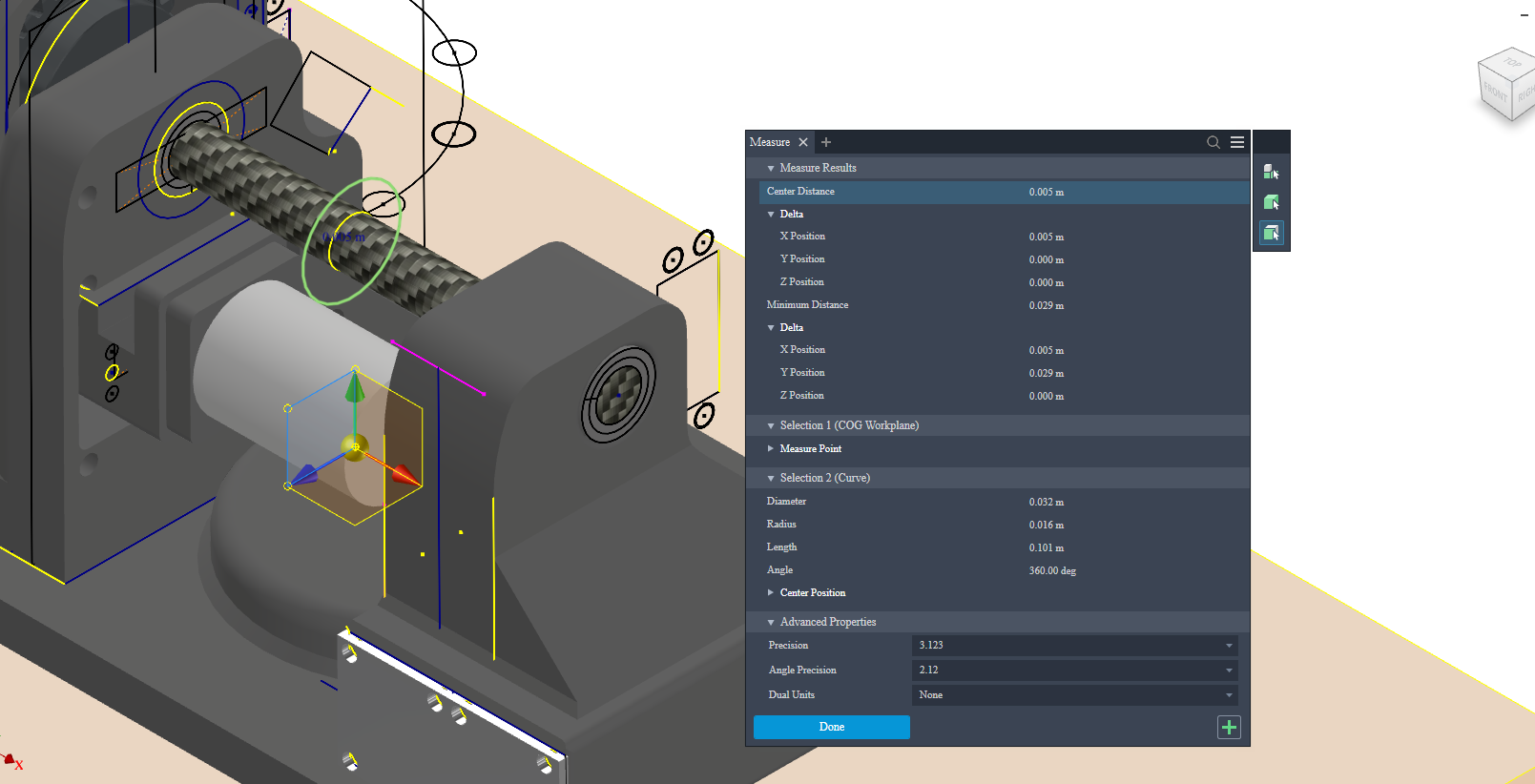}
    }
    \caption{Baseframe inertial measurements and Inventor principal--axis data.}
    \label{}
\end{figure}

\begin{figure}[H]
    \centering
    \includegraphics[width=0.5\linewidth]{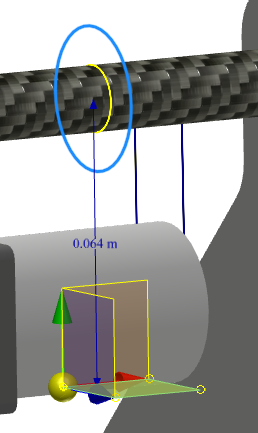}
    \caption{Base Frame measurement for Parallel Axis Shifting (Vertical)}
    \label{fig:placeholder}
\end{figure}

\subsection*{Base--Frame Inertia Transformation}

\noindent\textbf{Note:} The center--of--mass (CM) measurements used in this 
section are obtained from the \emph{measured link masses} rather than Autodesk 
Inventor’s reported CM. Inventor reports the center of mass relative to its own 
model reference frame, which does not coincide with the dynamic reference frame 
used in the manipulator formulation. For this reason, independent CM 
measurements were taken for all base--frame components.

\begin{figure}[H]
    \centering
    \includegraphics[width=0.75\linewidth]{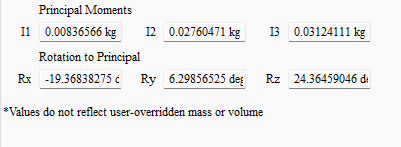}
    \caption{Base--Frame Rotation Matrices}
    \label{fig:baseframe_rotation}
\end{figure}

The CAD model reports the base--frame inertia tensor about its local 
center--of--gravity (CG) as
\[
I_0 =
\begin{bmatrix}
0.01187588 & 0          & 0 \\
0.00748855 & 0.02456819 & 0 \\
0.00036682 & -0.00146963 & 0.03076740
\end{bmatrix}.
\]

To rotate this tensor into the physical coordinate system used in the 
manipulator dynamics, the Inventor--reported principal--axis orientation angles 
are first converted to radians:
\[
(R_{0x}, R_{0y}, R_{0z}) =
(-19.3684^\circ,\; 6.2986^\circ,\; 24.3646^\circ).
\]

The three successive rotations are
\[
R_x =
\begin{bmatrix}
1 & 0 & 0 \\
0 & \cos R_{0x} & -\sin R_{0x} \\
0 & \sin R_{0x} &  \cos R_{0x}
\end{bmatrix},\qquad
R_y =
\begin{bmatrix}
\cos R_{0y} & 0 & \sin R_{0y} \\
0 & 1 & 0 \\
-\sin R_{0y} & 0 & \cos R_{0y}
\end{bmatrix},
\]
\[
R_z =
\begin{bmatrix}
\cos R_{0z} & -\sin R_{0z} & 0 \\
\sin R_{0z} &  \cos R_{0z} & 0 \\
0 & 0 & 1
\end{bmatrix}.
\]

The composite rotation from the CAD CG frame to the global CAD inertial frame is
\[
R_0 = R_z R_y R_x =
\begin{bmatrix}
1.0000 & -0.0074 & 0.0019 \\
0.0074 & 1.0000 & 0.0059 \\
-0.0019 & -0.0059 & 1.0000
\end{bmatrix}.
\]

Because the inertia must be expressed about the shaft axis in the manipulator 
frame, the measured center--of--mass offset
\[
(x_{\mathrm{Mount/CM}}, y_{\mathrm{Mount/CM}}, z_{\mathrm{Mount/CM}})
= (0.005,\; 0.006,\; 0.065)\;\text{m}
\]
and the measured base--frame mass
\[
m_{\mathrm{BF}} = 3.53790071\;\text{kg}
\]
are inserted into the parallel--axis theorem:
\[
I_{xx} = m_{\mathrm{BF}} (y_{\mathrm{Mount/CM}}^2 + z_{\mathrm{Mount/CM}}^2)
         + I_{\mathrm{body},x},
\]
\[
I_{yy} = m_{\mathrm{BF}} (x_{\mathrm{Mount/CM}}^2 + z_{\mathrm{Mount/CM}}^2)
         + I_{\mathrm{body},y},
\]
\[
I_{zz} = m_{\mathrm{BF}} (x_{\mathrm{Mount/CM}}^2 + y_{\mathrm{Mount/CM}}^2)
         + I_{\mathrm{body},z}.
\]

This yields the final inertia tensor of the base frame expressed about the 
shaft axis in the physical coordinate frame:
\[
I_{\mathrm{BF}} =
\operatorname{diag}(0.0270,\; 0.0396,\; 0.0310)\;\text{kg}\cdot\text{m}^2.
\]

It is consistent that $I_{yy}$ exceeds $I_{zz}$, because the base frame contains 
counterweights used to approximate the mass distribution of the internal gears 
and the drive motor. This places approximately $0.8$--$1$\,kg of additional mass 
near the lateral extremes of the structure, increasing the rotational inertia 
about the $y$--axis.
\subsection{Link 1}

Unlike the base frame, Link~1 is always treated about its own center of mass 
in the dynamic model. Therefore, no explicit parallel--axis correction is 
required: the shift from joint frame to CM is already handled by the 
position-vector definitions $\bar{\ell}_i$ within the Euler--Lagrange 
derivation.

\begin{figure}[H]
    \centering
    \subfloat[]{\includegraphics[width=0.45\linewidth]{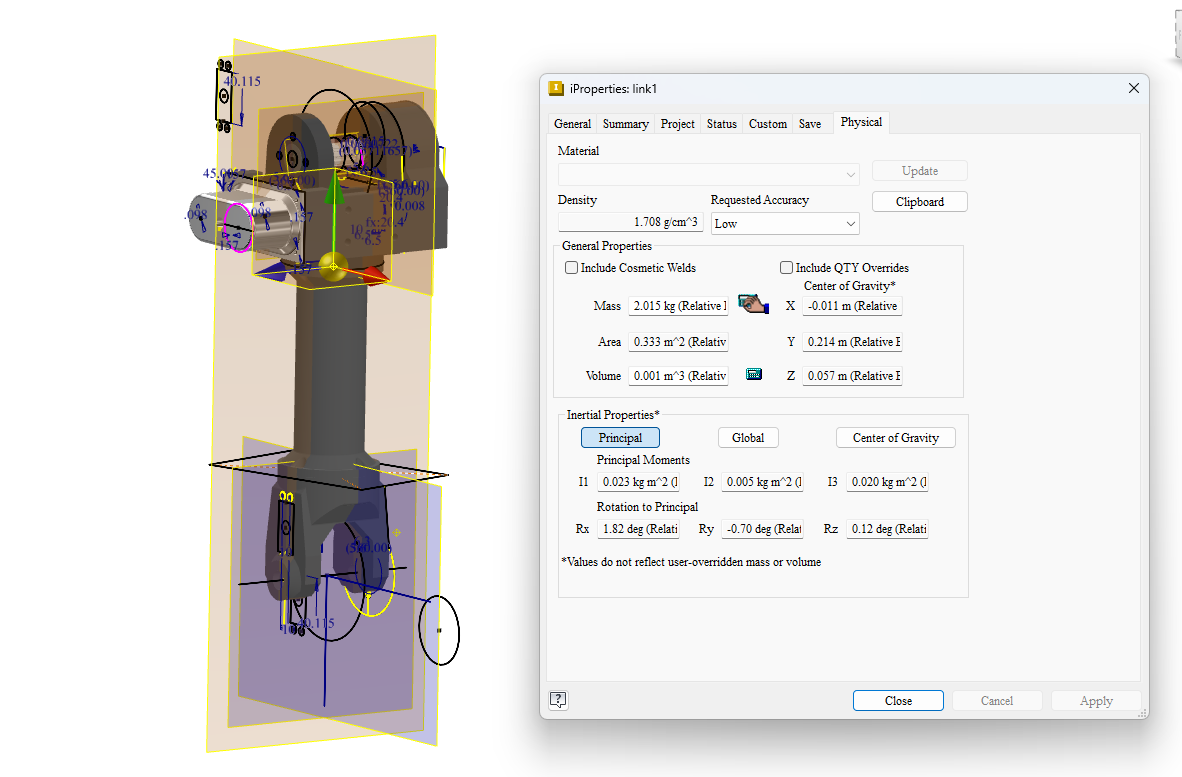}}
    \hfill
    \subfloat[]{\includegraphics[width=0.45\linewidth]{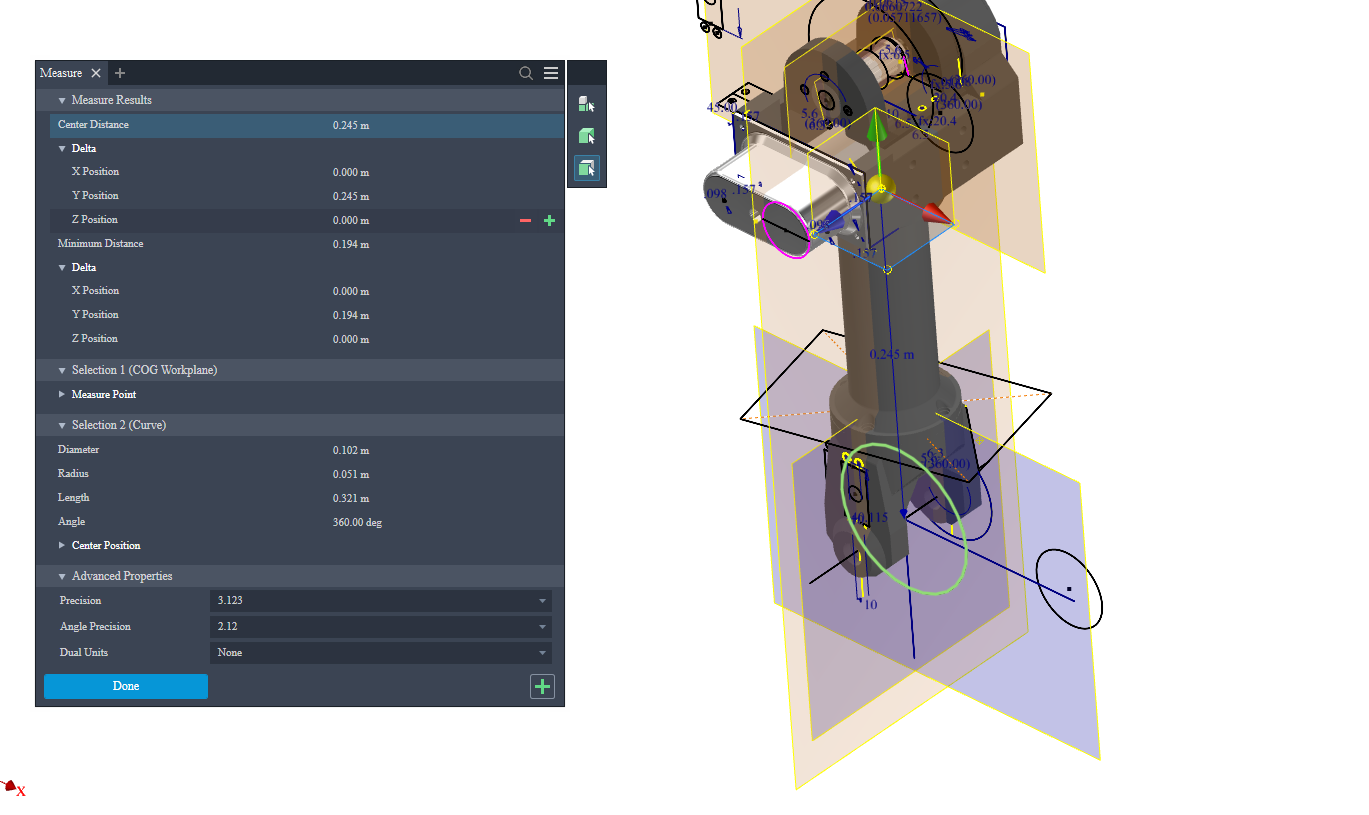}}
    \caption{Link~1 inertial measurements and Inventor principal-axis data.}
    \label{fig:link1_measurements}
\end{figure}

\vspace{0.5em}
The Inventor screenshot of Link~1, together with the measured CM-to-joint 
distance, is shown in Fig.~\ref{fig:link1_measurements}.
From Inventor’s physical properties window, Link~1 has:
\[
m_1 = 2.015\;\text{kg},
\qquad
I_1 = \operatorname{diag}(0.023,\;0.005,\;0.020)\;\text{kg}\cdot\text{m}^2,
\]
with principal-axis orientation
\[
(R_{x},R_{y},R_{z})
=
(1.82^\circ,\;-0.70^\circ,\;0.12^\circ).
\]

The corresponding rotation matrices are
\[
R_x =
\begin{bmatrix}
1 & 0 & 0 \\
0 & \cos R_x & -\sin R_x \\
0 & \sin R_x & \cos R_x
\end{bmatrix},\quad
R_y =
\begin{bmatrix}
\cos R_y & 0 & \sin R_y \\
0 & 1 & 0 \\
-\sin R_y & 0 & \cos R_y
\end{bmatrix},
\]
\[
R_z =
\begin{bmatrix}
\cos R_z & -\sin R_z & 0 \\
\sin R_z & \cos R_z & 0 \\
0 & 0 & 1
\end{bmatrix}.
\]

The composite rotation is
\[
R_0 = R_z R_y R_x,
\]
and the transformed inertia tensor is
\[
I_{1,\mathrm{body}} = R_0^\top I_1 R_0.
\]

Numerically, using the MATLAB code,
\[
\lambda(I_{1,\mathrm{body}})
=
\{0.021,\;0.0240,\;0.0050\}\;\text{kg}\cdot\text{m}^2,
\]
which matches Inventor’s principal values to machine precision.

Because Link~1 is modeled about its CM, this tensor is used directly in the 
manipulator mass matrix $M(q)$ without further translation.
\newpage
\subsection{Link 2}
As with Link~1, the dynamics for Link~2 are expressed directly about its center 
of mass. Therefore, no parallel--axis theorem correction is necessary, since 
all translational effects are already handled through the CM position vector 
$\bar{\ell}_2$ in the kinematic formulation. In this model, Link~2 includes the 
entire actuator assembly (outer housing), and the actuator rod was handled analytically in the derivation. 
In the earlier symbolic derivation, the actuator rod was modeled separately for 
clarity. For simplicity of computation, the rod’s interior is assumed to be solid, which slightly overestimates inertia but 
does not affect the qualitative dynamics or the control formulation.
\begin{figure}[t!]
    \centering
    \subfloat[Link 2 Inertial View\label{fig:link2a}]{
        \includegraphics[width=0.46\linewidth]{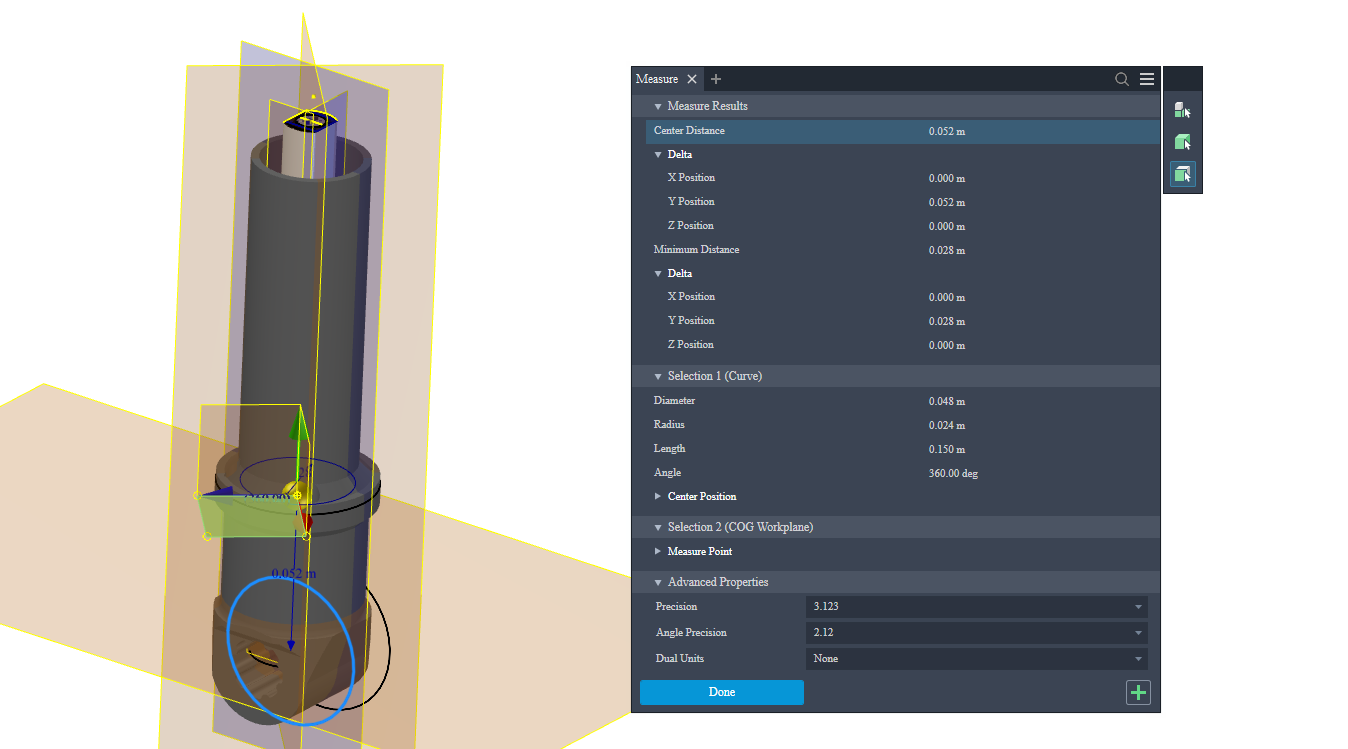}
    }\hfill
    \subfloat[Link 2 CM Measurement\label{fig:link2b}]{
        \includegraphics[width=0.46\linewidth]{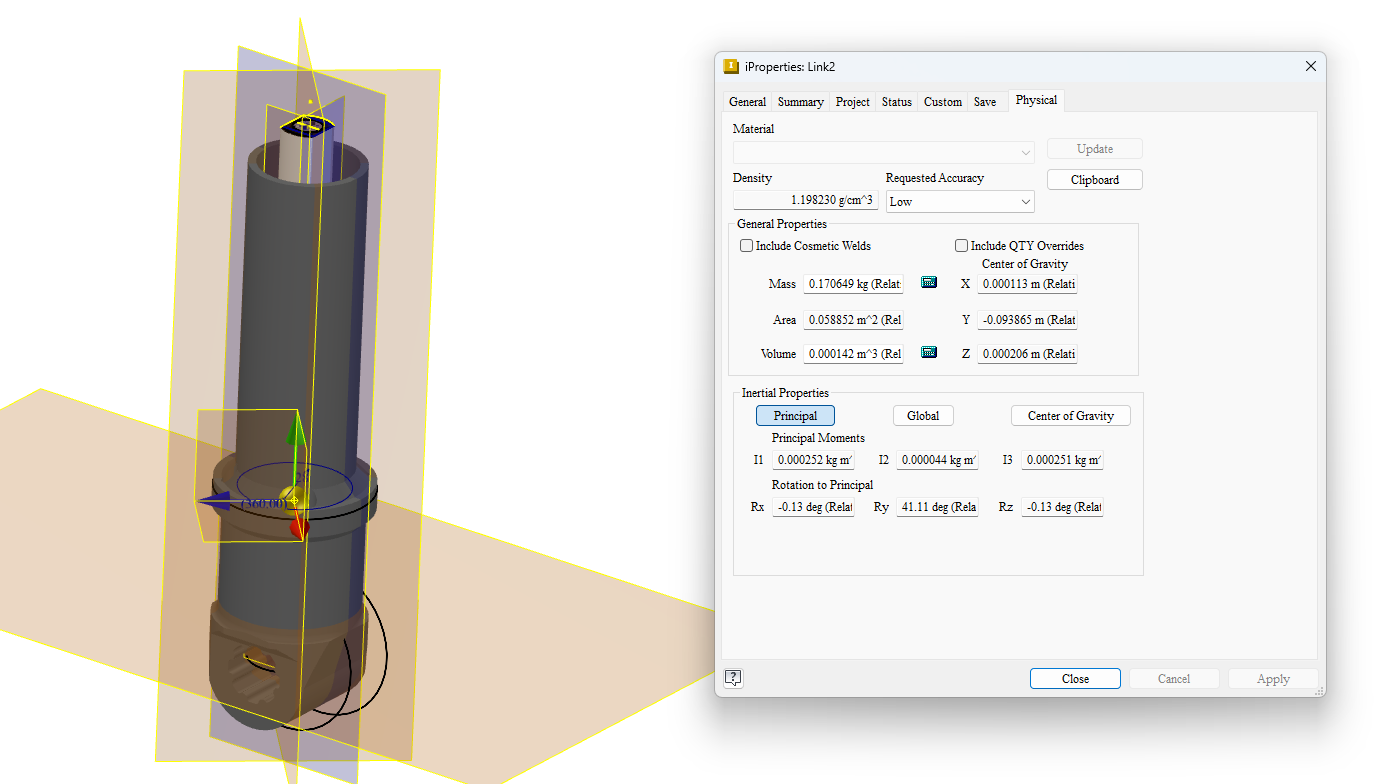}
    }
    \caption{Link 2 inertial measurements and principal--axis orientation.}
    \label{fig:link2_measurements}
\end{figure}
Inventor reports:
\[
m_2 = 0.170649\;\text{kg},
\qquad
I_2 = \operatorname{diag}(0.000252,\;0.000044,\;0.000251)\;\text{kg}\cdot\text{m}^2,
\]
with the principal-axis orientation
\[
(R_x,R_y,R_z)
= (-0.13^\circ,\;41.11^\circ,\;-0.13^\circ).
\]

Constructing the same rotation matrices,
\[
R_0 = R_z R_y R_x,
\qquad
I_{2,\mathrm{body}} = R_0^\top I_2 R_0,
\]
gives the eigenvalues
\[
\lambda(I_{2,\mathrm{body}})
=
\{2.53\!\times\!10^{-4},\;2.52\!\times\!10^{-4}\\;4.40\!\times\!10^{-5} \},
\]
Again, because Link~2 is always modeled about its center of mass, this inertia 
tensor is used directly in the full Euler--Lagrange formulation.

Finally, for the claw and end--effector assembly, the moment of inertia is not 
explicitly included. Its inertial distribution varies with the mass and shape 
of the object being grasped, and therefore no fixed inertia tensor would be 
representative. Instead, its effect enters implicitly through the kinematic 
position vectors used in the formulation; any payload contributes through its 
mass and CM location, which is sufficient for the torque and dynamic balance 
equations without requiring a dedicated parallel--axis correction.
\subsection{Claw}
\begin{figure}[H]
    \centering
    \subfloat[Claw Mass\label{fig:claw_mass}]{
        \includegraphics[width=0.46\linewidth]{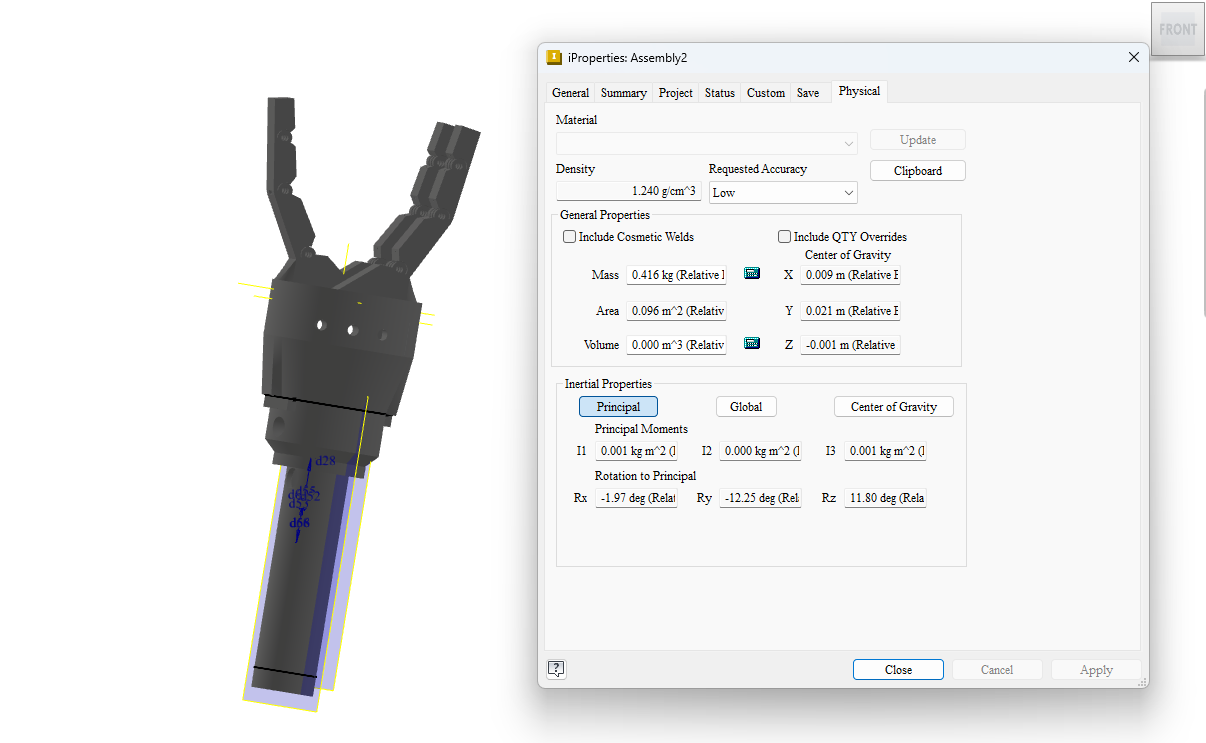}
    }\hfill
    \subfloat[Measurement of $\delta_{r}$\label{fig:delta_r}]{
        \includegraphics[width=0.46\linewidth]{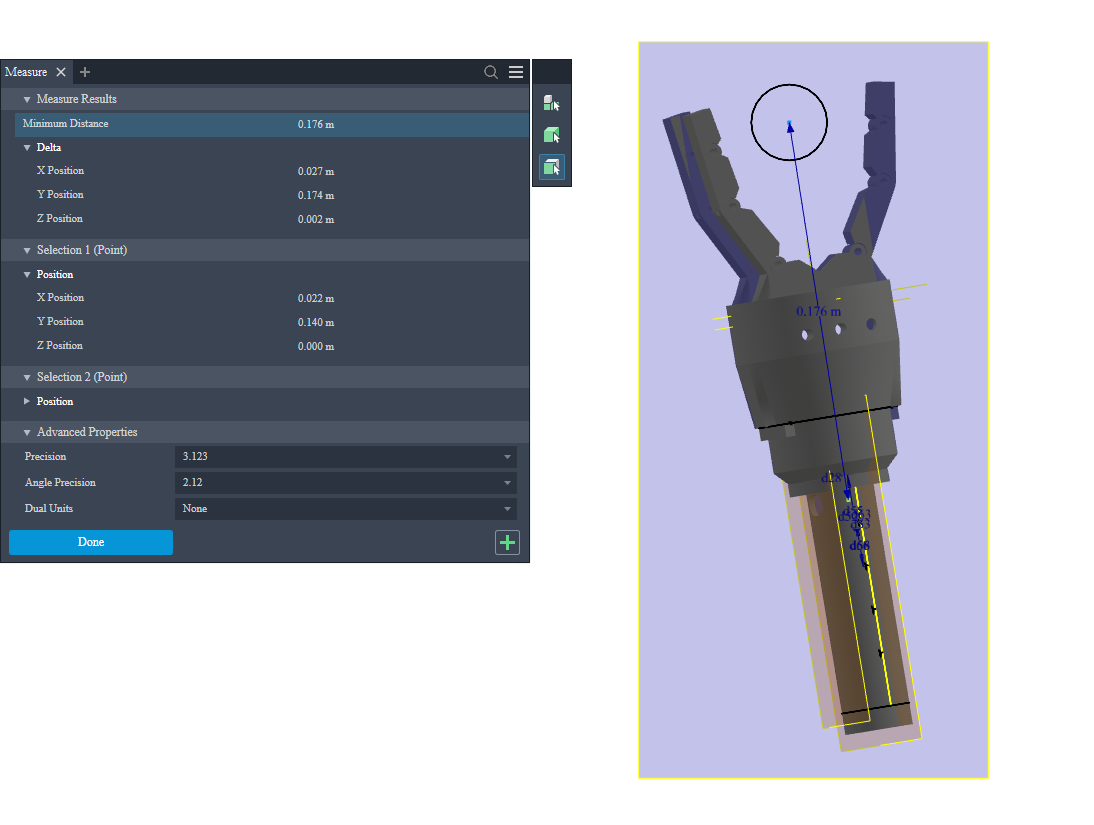}
    }
    \caption{Claw and $\delta_{r}$ measurements used for the effective end--effector model.}
    \label{fig:claw_and_delta_r}
\end{figure}

The Claw on the other hand; is not considered in terms of its inertia; since its a soft claw and thus its dynamics are nontrivial. So the Claw is approximated as a point mass, where: 
\begin{align*}
    m_{Obj} = m_{Claw} + m_{Obj}
\end{align*}
If the external object mass is zero, its merely the claw; and note here that the parameter $\delta_{r}$ is taken by measuring roughly where the claw pinchers are, and where the robot needs to reach; effectively acting as the end-effector. 
To summarize everything, please consult table 2 for the numerical results. 
\begin{table}[H]
\centering
\renewcommand{\arraystretch}{1.25}
\begin{tabular}{|l|c|c|c|l|}
\hline
\textbf{Component} 
& \textbf{Mass (kg)} 
& \textbf{CM Offset (m)} 
& \textbf{Inertia (kg$\cdot$m$^{2}$)} 
& \textbf{Notes} \\ 
\hline

Base Frame 
& 3.538 
& $(0.005,\;0.006,\;0.065)$ 
& $\mathrm{diag}(0.0270,\;0.0396,\;0.0310)$ 
& Parallel–axis corrected \\

Link 1 
& 2.015 
& $\bar{\ell}_{1}=0.152$ 
& $\mathrm{diag}(0.021,\;0.024,\;0.005)$ 
& About CM \\

Link 2 
& 0.171 
& $\bar{\ell}_{2}=0.052$ 
& $\mathrm{diag}(2.53\!\times\!10^{-4},\;2.52\!\times\!10^{-4},\;4.40\!\times\!10^{-5})$ 
& About CM \\

Actuator Rod 
& 0.228
& $r(t)$ 
& Analytical (cylindrical model) 
& Closed-form expression \\

Claw + Object 
& Variable 
& $\delta_{r}=0.103$ 
& Point mass 
& No inertia tensor used \\

\hline
\end{tabular}
\caption{Summary of inertial properties used in the dynamic model.}
\label{tab:inertial_summary}
\end{table}

\newpage

\end{document}